%% file: main.tex
\documentclass[]{template}

\usepackage[utf8]{inputenc}             
\usepackage[T1]{fontenc}                
\usepackage{url}                        
\usepackage{booktabs}                   
\usepackage{multirow}
\usepackage{colortbl}
\usepackage{multicol}
\usepackage{amsfonts}                   
\usepackage{nicefrac}                   
\usepackage{microtype}                  
\usepackage[dvipsnames]{xcolor}         
\usepackage{tocloft}
\usepackage{setspace}

\usepackage{latexsym}

\usepackage{graphicx}
\usepackage{float}
\usepackage{subcaption}
\usepackage{wrapfig}
\usepackage{lipsum}

\usepackage{adjustbox}
\usepackage{makecell}

\usepackage{bm}

\usepackage{tabularx} 
\usepackage{ragged2e} 
\newcolumntype{L}{>{\RaggedRight\hangafter=1\hangindent=0em}X}

\usepackage{enumitem}

\usepackage{amsmath}
\usepackage{amssymb}
\usepackage{mathtools}
\usepackage{amsthm}
\usepackage{arydshln}

\setboolean{logo}{true}    

\usepackage[linesnumbered,ruled,vlined]{algorithm2e}

\hypersetup{
    colorlinks=true,
    linkcolor=red,
    citecolor=Cerulean,
    filecolor=magenta,      
    urlcolor=magenta,
}

\usepackage[capitalize,noabbrev]{cleveref}
\crefname{section}{§}{§§}
\Crefname{section}{§}{§§}

\usepackage{calligra}
\DeclareMathAlphabet{\mathcalligra}{T1}{calligra}{m}{n}

\usepackage{pifont}

\theoremstyle{plain}

\theoremstyle{definition}

\theoremstyle{remark}

\renewcommand{\paragraph}[1]{\vspace{1mm}\noindent\textbf{#1}}

\DeclareCaptionLabelFormat{cont}{#1~#2\alph{ContinuedFloat}}
\usepackage[most]{tcolorbox}
\tcbset{
  promptbox/.style={
    top=10pt,
    colback=lightgray!20,
    colframe=Black,
    colbacktitle=NavyBlue,
    enhanced,
    center,
    attach boxed title to top center={yshift=-0.1in,xshift=0.0in},
    boxed title style={boxrule=0pt,colframe=white,},
  }
}
\newtcolorbox{promptbox}[2][]{promptbox, title=#2,#1}
\tcbset{
  takeawaybox/.style={
    top=10pt,
    colback=lightgray!20,
    colframe=Black,
    colbacktitle=BurntOrange,
    enhanced,
    center,
    attach boxed title to top center={yshift=-0.1in,xshift=0.0in},
    boxed title style={boxrule=0pt,colframe=white,},
  }
}
\newtcolorbox{takeawaybox}[2][]{takeawaybox, title=#2,#1}
\tcbset{
  observationbox/.style={
    top=10pt,
    colback=lightgray!20,
    colframe=Black,
    colbacktitle=YellowGreen,
    enhanced,
    center,
    attach boxed title to top center={yshift=-0.1in,xshift=0.0in},
    boxed title style={boxrule=0pt,colframe=white,},
  }
}
\newtcolorbox{observationbox}[2][]{observationbox, title=#2,#1}

\usepackage{xspace}

\newcommand\blfootnote[1]{%
  \begingroup
  \renewcommand\thefootnote{}\footnote{#1}%
  \addtocounter{footnote}{-1}%
  \endgroup
}

\usepackage{twemojis}
\usepackage{fontawesome}
\newcommand{\huggingface}{\raisebox{-1.5pt}{\includegraphics[height=1.05em]{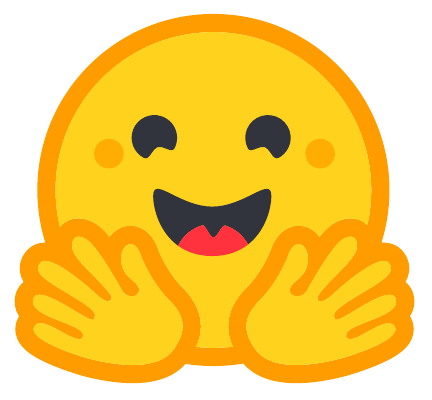}}\xspace}

\newcommand{\github}{\raisebox{-1.5pt}{\includegraphics[height=1.05em]{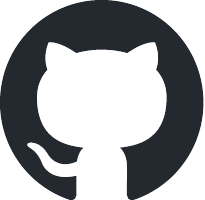}}\xspace}

\usepackage{CJK}

\usepackage{duckuments} 
\usepackage{marvosym}
\usepackage{authblk}

\input{preamble}

\title{\modelname: Democratizing Unified Multimodal Models for Understanding, Reasoning, Generation and Editing}

\author[1,2*]{Changyao Tian}
\author[1,3*]{Danni Yang}
\author[1,4*]{Guanzhou Chen}
\author[1,4*]{Erfei Cui}
\author[1,4*]{Zhaokai Wang} 
\author[1,2*]{\authorcr Yuchen Duan}
\author[1,5*]{Penghao Yin}

\author[1,3]{Sitao Chen}
\author[1,6]{Ganlin Yang}

\author[4]{Mingxin Liu}
\author[4]{Zirun Zhu}
\author[7]{\authorcr Ziqian Fan}
\author[4]{Leyao Gu}
\author[1,4]{Haomin Wang}
\author[1,8]{Qi Wei}
\author[1,8]{Jinhui Yin}
\author[4]{Xue Yang}
\author[4]{Zhihang Zhong}

\author[1]{\authorcr Qi Qin}
\author[1]{Yi Xin}
\author[1]{Bin Fu}
\author[1]{Yihao Liu}
\author[1]{Jiaye Ge}
\author[1]{Qipeng Guo}

\author[9]{Gen Luo}
\author[2]{Hongsheng Li}
\author[1]{Yu Qiao}

\author[1\textsuperscript{\Letter}]{\authorcr Kai Chen}
\author[1\textsuperscript{\Letter}]{Hongjie Zhang}

\affil[1]{Shanghai AI Laboratory}
\affil[2]{CUHK MMLab}
\affil[3]{Fudan University}
\affil[4]{Shanghai Jiao Tong University}

\affil[5]{Tsinghua University}
\affil[6]{University of Science and Technology of China}
\affil[7]{South China University of Technology}
\affil[8]{Nanjing University}
\affil[9]{Xiamen University}
\input{sections/0.abstract}

\begin{document}

\blfootnote{* Equal Contribution. \textsuperscript{\Letter} Corresponding author. This work was done when Changyao Tian was an intern at Shanghai AI Laboratory. }

\maketitle

\begin{center}
    \renewcommand{\arraystretch}{1.5}
    \vspace{1em}
    \begin{tabular}{rll}
        \github{} & \textbf{GitHub Repo} & \url{https://github.com/OpenGVLab/InternVL-U} \\
        \huggingface{} & \textbf{HuggingFace Model} & \url{https://huggingface.co/InternVL-U/InternVL-U} \\
        \github{} & \textbf{GenEditEvalKit} & \url{https://github.com/open-compass/GenEditEvalKit} \\
        \github{} & \textbf{TextEdit Benchmark} & \url{https://github.com/open-compass/TextEdit} \\
    \end{tabular}
\end{center}

\clearpage

\input{sections/1.introduction}
\input{sections/2.related}
\input{sections/3.methodology}
\input{sections/4.data}

\input{sections/5.experiment}
\input{sections/6.conclusion}

\clearpage
\bibliographystyle{plain}
\bibliography{refs}


\clearpage
\appendix
\input{sections/appendix}


\end{document}

%% file: sections/0.abstract.tex
\begin{abstract}
Unified multimodal models (UMMs) that integrate understanding, reasoning, generation, and editing face inherent trade-offs between maintaining strong semantic comprehension and acquiring powerful generation capabilities.
In this report, we present \modelname, a lightweight 4B-parameter UMM that democratizes these capabilities within a unified framework.
Guided by the principles of unified contextual modeling and modality-specific modular design with decoupled visual representations, \modelname integrates a state-of-the-art Multimodal Large Language Model (MLLM) with a specialized MMDiT-based visual generation head.
To further bridge the gap between aesthetic generation and high-level intelligence, we construct a comprehensive data synthesis pipeline targeting high-semantic-density tasks, such as text rendering and scientific reasoning, under a reasoning-centric paradigm that leverages Chain-of-Thought (CoT) to better align abstract user intent with fine-grained visual generation details.
Extensive experiments demonstrate that \modelname achieves a superior performance - efficiency balance. Despite using only 4B parameters, it consistently outperforms unified baseline models with over 3× larger scales such as BAGEL (14B) on various generation and editing tasks, while retaining strong multimodal understanding and reasoning capabilities.
\end{abstract}

%% file: sections/1.introduction.tex
\begin{figure}[!htbp]
    \centering
    \includegraphics[width=0.95\linewidth]{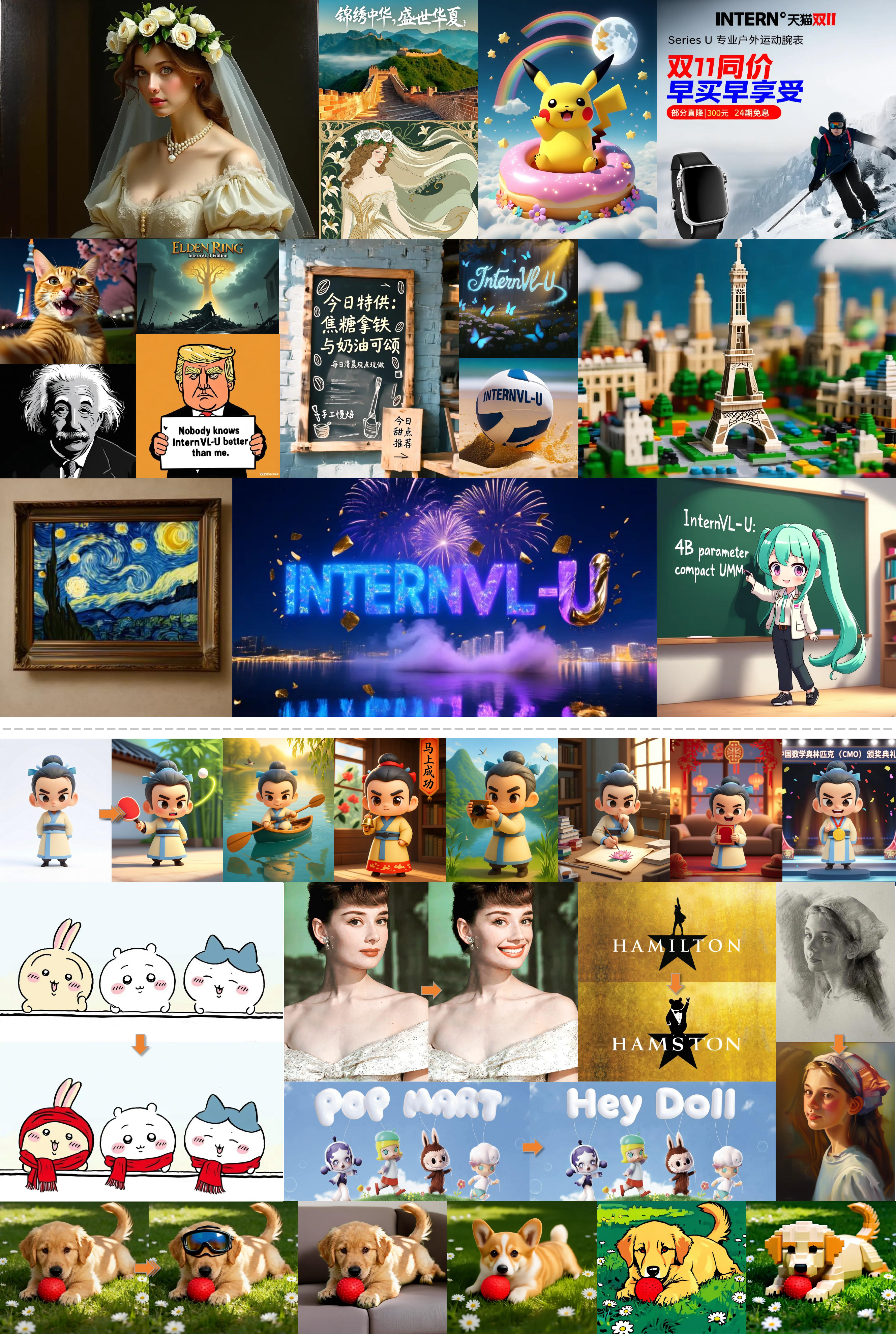} 
    \caption{
    \textbf{Showcases of \modelname for general text-to-image generation (top) and image editing (bottom).} \modelname supports high-fidelity image generation and editing at any resolution. 
    }
    \label{fig:teaser}
\end{figure}

\begin{figure}[!htbp]
    \centering
    \includegraphics[width=0.95\linewidth]{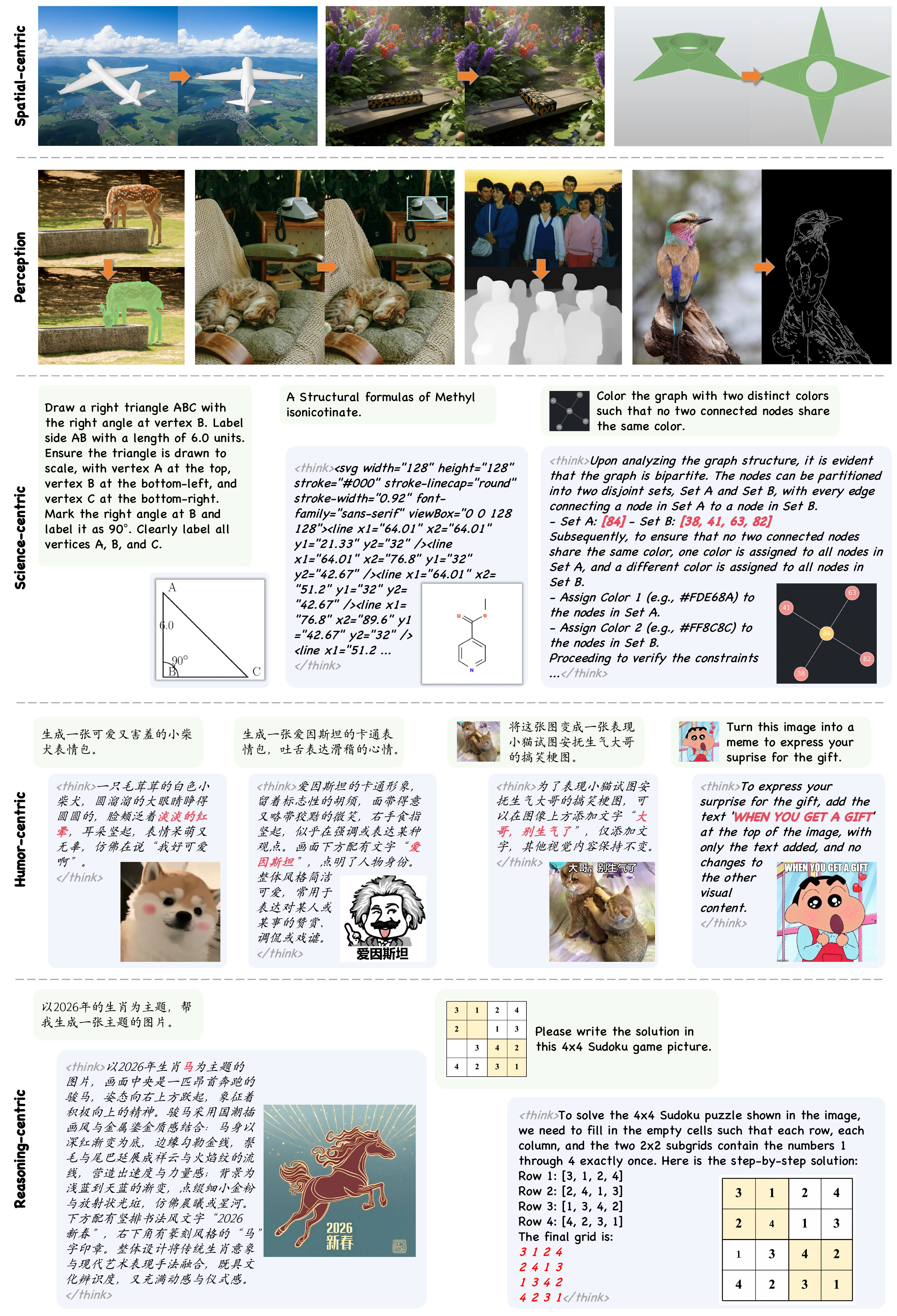} 
    \caption{
    \textbf{Showcases of \modelname for spatial-centric, perception, science-centric, humor-centric, and reasoning-centric text-to-image generation or editing tasks.} \modelname demonstrates such core multimodal capabilities across various visual domains.
    }
    \label{fig:teaser2}
\end{figure}

\clearpage
\newpage
\renewcommand{\contentsname}{\centering \Large \bfseries Contents}

\definecolor{softblue}{RGB}{55,65,204}
\hypersetup{
    linkcolor=softblue,
}

\begingroup
\setstretch{1.3}
\tableofcontents
\endgroup
\hypersetup{
    colorlinks=true,
    linkcolor=red,
    citecolor=Cerulean,
    filecolor=magenta,      
    urlcolor=magenta,
}

\clearpage

\section{Introduction}

Unified multimodal models (UMMs) have witnessed rapid advancement in recent years~\cite{team2024chameleon,chen2025blip3,li2025onecat,tian2025unigen,tang2025unilip}. The emergence of models like GPT-4o~\cite{hurst2024gpt} demonstrates that integrating native image generation with advanced linguistic capabilities not only enables users to execute complex visual tasks via natural language but also paves the way for exploring Artificial General Intelligence (AGI) and World Models~\cite{deng2025bagel,cui2025emu3}. While closed-source models have exhibited remarkable general-purpose performance, the research community has actively explored various architectural and representational strategies to construct such unified models. These efforts can generally be categorized into two paradigms: (1) \textit{Fully-native UMMs}~\cite{team2024chameleon,cui2025emu3,deng2025bagel,wang2025ovis,xin2025lumina,xie2025show}, which is trained from scratch or initialized from unimodal components (\eg, ViT, LLM) and jointly trained on multimodal understanding and generation tasks from scratch; and (2) \textit{Fully-ensemble UMMs}~\cite{pan2025transfer,lin2025uniworld,song2025query,wu2025openuni}, which construct a unified system by post-hoc aligning pre-trained multimodal understanding models with pre-trained image generation models. However, both paradigms face significant limitations.

For fully-native UMMs, the community has yet to reach a consensus on the optimal design across modeling, representation, and architecture~\cite{he2025emmaefficientmultimodalunderstanding}. Not only do theoretical divergences exist, but no single approach has demonstrated a decisive advantage in performance or efficiency~\cite{li2025onecat}. Furthermore, jointly training multimodal understanding and generation capabilities from scratch presents substantial engineering challenges, particularly in balancing the conflicting data distributions of different modalities. Crucially, this paradigm often necessitates foregoing the benefits of state-of-the-art (SOTA) multimodal understanding models~\cite{Qwen-VL,chen2024internvl} already available in the community, thereby incurring prohibitive training costs and risks.
Conversely, fully-ensemble UMMs typically attach an external and separately pre-trained image generator as a visual generation head~\cite{xie2024sana,flux2024,rombach2022high}. In practice, they face a recurring trade-off. They can scale the head to very large parameter counts to reach top-tier visual quality, as in Qwen-Image~\cite{wu2025qwen} and Hunyuan Image 3.0~\cite{cao2025hunyuanimage}, but this substantially increases training and deployment cost. Alternatively, they can retain a smaller head while introducing elaborate and often fragmented conditioning pipelines, such as requiring multi-encoder text conditioning in Stable Diffusion 3~\cite{mmdit} or designs that decouple text and image conditions in Z-Image~\cite{cai2025z}. Either way, the resulting interfaces are difficult to align cleanly with the hidden-state space of a single MLLM, which limits how much can be gained from post-alignment training under constrained resources.

To address these challenges, we first systematically analyze the design principles of unified models from three dimensions: modeling, architecture, and representation. We posit that within a unified semantic reasoning space, a model should employ hybrid modeling objectives to accommodate the statistical properties of different modalities, adhere to modality-specific modularity to enhance overall architectural efficiency, and utilize decoupled visual representations to balance high-level semantic understanding with low-level pixel reconstruction. Guided by these principles, we propose \modelname, a streamlined and efficient unified multimodal model. Built upon InternVL 3.5~\cite{wang2025internvl3_5}, an open-source MLLM with SoTA performance, we integrate a custom MMDiT-based visual generation head with a unified semantic conditioning interface aligned to the MLLM hidden states. Through a three-stage progressive training strategy, \modelname not only inherits the robust understanding and reasoning capabilities of its predecessor but also acquires powerful multimodal generation and editing skills. Furthermore, \modelname leverages self-reflection reasoning to utilize world knowledge inherited from the MLLM, further enhancing these capabilities.

Design unification alone, however, does not guarantee a truly AGI-oriented UMM, because the capabilities a model ultimately acquires are strongly shaped by the objectives and data regimes it is trained on~\cite{rombach2022high,wang2025internvl3_5}. A unified multimodal model is expected to be both visually competent and semantically reliable, yet today's visual generation and multimodal understanding models are optimized for fundamentally different goals and use cases. Traditional generation models primarily target \textit{low-level} perceptual quality, such as aesthetics and visual fidelity, whereas understanding models emphasize \textit{high-level} intelligence, including knowledge injection and reasoning emergence. This objective mismatch poses a major obstacle to developing AGI-oriented UMMs. We argue that a key driver is the domain gap in training data distributions. Generation models are predominantly trained on natural-image corpora (\eg, portraits and landscapes) rich in texture and high-frequency details but relatively low in semantic density. In contrast, understanding models rely heavily on text-rich and structurally organized data, including synthetic images such as GUIs, infographics, and OCR-centric documents, which may exhibit simpler textures but contain dense semantics, abundant textual cues, and structured knowledge.

Consistent with this diagnosis, next-generation commercial models (\eg, Nano-Banana Pro~\cite{deepmind_gemini3proimage_2025}) have begun to actively narrow the gap by emphasizing typographic precision and knowledge-faithful content creation beyond aesthetics alone.
Inspired by this trend, and to unlock the potential of \modelname as an AGI-oriented UMM, we construct a comprehensive multimodal data synthesis pipeline targeting various capabilities, including text rendering, scientific reasoning, spatial and humor generation. Specifically, for ``high semantic density'' text scenarios, we design a fully automated text rendering and editing pipeline covering bilingual typography and local consistency editing to address the lack of symbolic precision in generative models. For ``knowledge-intensive'' scientific scenarios, we leverage programmatic tools (\eg, GeoGebra, SVG) and academic corpora to construct structured visual-text data across disciplines like mathematics, physics, and computer science. Furthermore, to better capture the abstract and underspecified nature of user intent, we propose a ``Reasoning-centric'' data synthesis paradigm. By introducing explicit Chain-of-Thought (CoT), we transform vague instructions into executable steps containing planning and constraints, achieving a leap from simple instruction following to deep intent alignment in tasks such as meme generation, geometric transformation, and logically constrained editing. 
By integrating data from these pipelines, \modelname retains its powerful general-purpose generation capabilities while significantly enhancing its ability to generate accurate text rendering and editing, spatial reasoning, humor generation, and multidisciplinary scientific knowledge.

Extensive empirical evaluations demonstrate that \modelname achieves a superior balance between performance and efficiency. As discussed in Section~\ref{sec:experiments}, in text-to-image generation, it consistently outperforms existing unified models across general, text-centric, and knowledge-intensive benchmarks, approaching the capabilities of significantly larger specialized generation models. Specifically, it exhibits exceptional instruction following and effectively addresses the deficiency of previous unified architectures in legible text rendering. Crucially, integrating the CoT strategy serves as a vital catalyst for both generation and editing, enabling the model to excel in knowledge-rich generation and complex logic-dependent editing tasks while delivering remarkable performance gains. Furthermore, regarding multimodal understanding, \modelname retains the robust capabilities of its predecessor, surpassing comparable unified baselines without compromising its native visual-language comprehension. To facilitate efficient benchmarking for the community, we further introduce \textit{GenEditEvalKit}~\cite{umm_evalkit_github} to streamline UMM evaluation and \textit{TextEdit Benchmark}~\cite{textedit_github} to provide a more comprehensive text-editing benchmark.

To summarize, our contributions are threefold:
\begin{itemize}
    \item We propose \modelname, an efficient UMM built on Unified Contextual Modeling, Decoupled Visual Representations, and Modality-Specific Modularity. By integrating a customized MMDiT-based generation head with decoupled ViT and VAE representations, our architecture resolves the conflict between semantic understanding and pixel reconstruction, enabling powerful generative skills without compromising native understanding capabilities.

    \item We construct a comprehensive data pipeline targeting high-semantic-density tasks, including text rendering, scientific reasoning, spatial manipulation, and humor generation. Furthermore, we introduce the ``Reasoning-centric'' paradigm that leverages Chain-of-Thought to transform abstract user instructions into executable plans, effectively bridging the gap between vague intent and precise visual execution.

    \item Extensive evaluations demonstrate that \modelname consistently outperforms unified baselines in generation and editing, particularly in text-rich and knowledge-intensive scenarios. Crucially, it retains robust multimodal understanding capabilities, surpassing comparable unified models without compromising visual-language comprehension.
    
\end{itemize}

%% file: sections/2.related.tex
\section{Related Work}

\subsection{Multimodal Large Language Models}
Recent advancements in Multimodal Large Language Models (MLLMs) have revolutionized vision-language tasks. Representative open-source families, such as LLaVA~\cite{liu2023llava,liu2023improvedllava,liu2024llavanext}, Qwen-VL~\cite{Qwen-VL,Qwen2-VL,Qwen2.5-VL,Qwen3-VL}, and InternVL~\cite{chen2024internvl,chen2024expanding,chen2024far,zhu2025internvl3,luo2024mono_internvl,mono_internvl_v1.5,wang2025internvl3_5}, alongside proprietary models like GPT~\cite{achiam2023gpt,GPT-5} and Gemini~\cite{team2024gemini,comanici2025gemini,gemini25,Gemini-3-Flash}, have demonstrated exceptional capabilities in visual understanding.
Standard MLLMs typically adopt a unified architecture connecting a vision encoder~\cite{dosovitskiy2020image} with an LLM~\cite{touvron2023llama,yang2025qwen3,anil2023palm} via adapters~\cite{li2022blip,liu2023improvedllava}.
Furthermore, recent trends have expanded towards processing interleaved image-text sequences~\cite{cui2025emu35nativemultimodalmodels,deng2025bagel,tian2024mminterleaved,yang2024vision} and video understanding~\cite{2023videochat,Maaz2023VideoChatGPT,lin2023video,wang2025internvideo2,yang2025cambrian,yang2025kwai}, pushing the boundaries of long-context multimodal interaction.

\subsection{Visual Generative Models}
Visual generation has evolved from early GANs~\cite{isola2017image,karras2019style,goodfellow2020generative} to dominant diffusion-based frameworks~\cite{ho2020denoising,rombach2022high,flux2024} and flow-matching paradigms~\cite{lipman2024flowmatchingguidecode,liu2022flow}, which offer superior scalability and sample quality. Parallelly, discrete token-based approaches~\cite{tian2024visual,esser2021taming,ramesh2021zero,chang2022maskgit} generate images autoregressively via VQ-style codecs, enabling a unified token space with LLMs.
State-of-the-art text-to-image models, including Stable Diffusion 3.5~\cite{rombach2022high}, FLUX.2~\cite{flux-2-2025}, Hunyuan Image~3.0~\cite{cao2025hunyuanimage}, and Qwen-Image~\cite{wu2025qwen}, emphasize instruction following and complex scene generation.
Meanwhile, the community has explored data-centric and architecture-centric approaches to improve rendering of complex structures and text, as well as generalization to long prompts and multi-concept descriptions~\cite{wu2025qwen,wang2025ovis_image,team2025longcat,cai2025z}. In addition, several closed-source state-of-the-art systems, such as Nano Banana Pro~\cite{deepmind_gemini3proimage_2025}, GPT-Image-1.5~\cite{GPT-Image-1.5}, and Seedream 4.0~\cite{seedream2025seedream}, have demonstrated strong performance on instruction-following and complex multi-concept image generation tasks.
Additionally, instruction-driven editing~\cite{brooks2023instructpix2pix,labs2025flux1kontextflowmatching,liu2025step1x-edit} has gained traction, requiring models to manipulate specific regions while preserving semantic consistency.

\subsection{Unified Multimodal Models}
Unified Multimodal Models (UMMs) aim to integrate understanding, generation, and editing within a single foundation model. By coupling a powerful LLM with visual tokenizers or latent representations, UMMs can understand and generate visual content in a unified manner.
Existing approaches generally fall into two categories:
(1) \textbf{Auto-Regressive Discrete-token methods}, such as Chameleon~\cite{team2024chameleon}, Emu3~\cite{cui2025emu3}, and SynerGen-VL~\cite{li2025synergen}, treat image generation as next-token prediction, naturally unifying modalities but often facing challenges in visual fidelity.
(2) \textbf{Diffusion/Hybrid methods}, such as BLIP-3o~\cite{chen2025blip3o}, BAGEL~\cite{deng2025bagel}, Ovis-U1~\cite{wang2025ovis}, and others~\cite{li2025onecat,tian2025unigen,shen2025mammothmoda2,liu2025tuna,wang2025skywork,li2025uniworld,he2025emmaefficientmultimodalunderstanding}, combine the reasoning power of LLMs with the high-fidelity generation of diffusion (or flow matching) models.
Recent works also explore other distinct unified  paradigms~\cite{modelmanzano,tang2025unilip,li2025lavida,yang2025mmada,xin2025lumina}.
Following this line of research, our work \modelname is built upon the open-source MLLM (\ie InternVL3.5~\cite{wang2025internvl3_5}) to unify general-purpose understanding, generation, and editing, as well as capabilities for domain-specific scenarios (\eg text rendering, science, memes), within a single framework.

%% file: sections/3.methodology.tex
\section{Method: \modelname}

\begin{figure}[!t]
    \centering
    \includegraphics[width=0.95\linewidth]{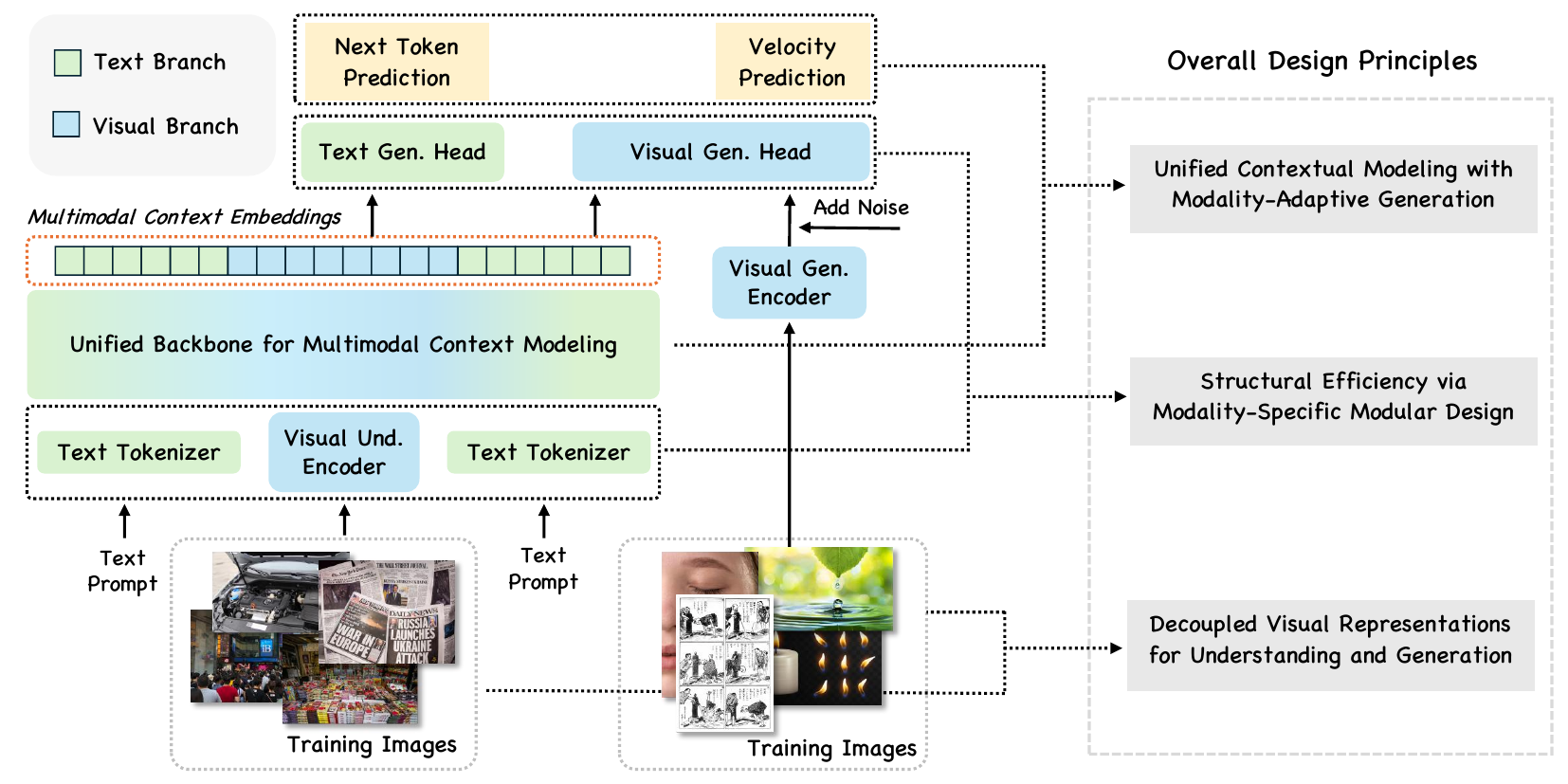} 
    \caption{\textbf{The architectural design of \modelname}. The framework highlights three design principles: (1) unified contextual modeling supporting modality-adaptive generation targets, (2) structural efficiency via a unified backbone with modality-specific modular design, and (3) decoupled visual representations for understanding and generation tasks. 
    Und. and Gen. denote Understanding and Generation, respectively.
    }
    \label{fig:overall_arch}
\end{figure}

\subsection{Model Architecture}

In this section, we elaborate on the overall architectural design principle of \modelname, along with the detailed architecture of our visual generation head.

\subsubsection{Overall Design Principles}
\label{sec:model_design}

As shown in~\cref{fig:overall_arch}, unlike recent approaches that enforce a homogenized processing pipeline for all modalities~\cite{liang2024mixture}, our architecture is driven by the philosophy that distinct modalities require tailored handling to maximize efficiency and performance.
We articulate our design principles through three key dimensions: modeling paradigm, structural efficiency, and data representation.  

\noindent\textbf{Unified Contextual Modeling with Modality-Adaptive Generation.}
Our first principle addresses the dichotomy between multimodal understanding (context) and generation (prediction). We argue that while contextualization benefits from a unified representation to facilitate deep semantic fusion, generation shall respect the inherent statistical properties of each modality.

\begin{itemize}
    \item \textbf{Unified Context, Adaptive Targets}: In the context phase, we project both visual and linguistic tokens into a shared latent space, employing a unified autoregressive (AR) paradigm with causal masking. This ensures that the model captures the complex high-level semantic dependencies between modalities during the reasoning process.
    \item \textbf{Hybrid Generative Objectives}: However, for the prediction targets, we diverge from the ``tokenization-for-all'' approach~\cite{cui2025emu3}. Text, being inherently discrete and sequential, is best modeled via a categorical distribution over a finite vocabulary using cross-entropy loss. Conversely, visual signals are continuous and spatially correlated. 
    While discrete visual tokenization is a viable alternative (as in VQ-VAE-based AR models), it may introduce quantization bottlenecks and make fine-grained spatial modeling less direct.
    Therefore, we adopt a hybrid AR + Diffusion modeling paradigm. We model image generation in a continuous multivariate probability space using Flow Matching (a generalized formulation of diffusion), while retaining the AR objective for text. 
    This design allows the model to preserve the strengths of autoregressive language modeling for text, while leveraging the high-fidelity generation capabilities of diffusion-based methods for images.
\end{itemize}

\noindent\textbf{Structural Efficiency via Modality-Specific Modular Design}. 
Our second principle addresses the computational inefficiency of fully modality-agnostic architectures (\eg, Mixture-of-Transformer~(MoT)~\cite{liang2024mixture}), which treat all modalities as uniform token sequences. We argue that different modalities possess varying "semantic densities": text is semantically dense, whereas raw visual patches are sparse and redundant.

\begin{itemize}
    \item \textbf{Encoder-Based MLLM Initialization}: To mitigate the parameter and FLOPs wastage inherent in processing raw modalities with a generic transformer, we incorporate modality-specific encoding stems. We initialize our multimodal context modeling backbone with an encoder-based architecture (leveraging a pre-trained ViT~\cite{chen2024internvl}) rather than a more monolithic or native multimodal design~\cite{luo2025mono, tian2025navil}. 
    This design introduces a necessary inductive bias that efficiently aggregates visual information before it enters the unified latent space.
    \item \textbf{Modality-Specific Generation Head}: Furthermore, recognizing that the decoding requirements for text and images differ, we extend the pre-trained MLLM with a dedicated generation head based on the Multimodal Diffusion Transformer (MMDiT)~\cite{mmdit} architecture for image generation. Instead of burdening the context modeling backbone with pixel-level synthesis, 
    the MMDiT serves as a dedicated generative module that takes the unified hidden states as conditioning signals and synthesizes images in a continuous visual latent space.
    This hierarchical design ensures that the backbone focuses on semantic reasoning, while the specialized stems and heads handle modality-specific translation, resulting in a more unified yet computationally efficient UMM.
\end{itemize}

\noindent\textbf{Decoupled Visual Representations for Understanding and Generation}. Our third principle challenges the assumption that the visual representation used for comprehending an image must be identical to the one used for generating it. We propose an asymmetric representation strategy 
motivated by the observation that image understanding primarily relies on semantically informative features, whereas image generation additionally requires representations that preserve reconstructable low-level visual details,
much as humans can perceive complex scenes they cannot necessarily draw.

\begin{itemize}
    \item \textbf{Semantic Input for Context Understanding}: For the understanding tasks (context), we only utilize high-level semantic features extracted directly from raw pixels via a pre-trained ViT. This helps preserve the semantic fidelity required for complex reasoning.
    \item \textbf{Compressed Output for Generation Target}: For the generation tasks (target), we employ a separate Variational Autoencoder (VAE) trained specifically for image reconstruction. This VAE compresses images into a latent space suitable for synthesis. 
\end{itemize}

By decoupling these representations, we not only avoid the ``optimization trade-off'', where a single encoder struggles to balance the high-level abstraction needed for understanding with the low-level pixel details needed for generation, but also avoid the increased computation cost and infrastructure complexity incurred by inputting the generation target into the context backbone. This allows us to leverage the strongest available pre-trained encoders for understanding without compromising the generative quality.

\begin{figure}[!t]
    \centering
    \includegraphics[width=0.95\linewidth]{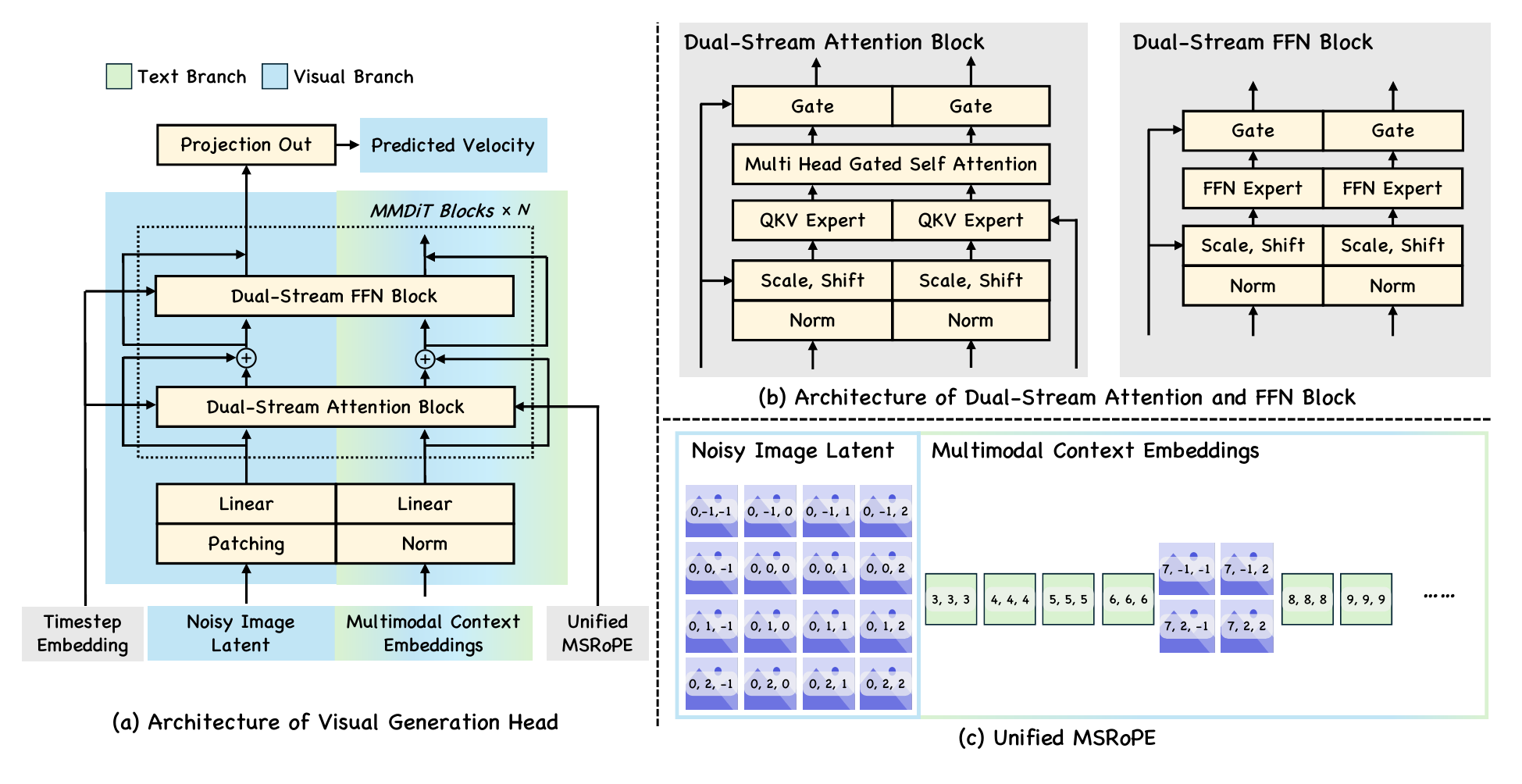} 
    \caption{\textbf{Architecture of the Visual Generation Head}. (a) Overview of the head with dual-stream MMDiT blocks. (b) Detailed structure of the Dual-Stream Attention Block and Dual-Stream FFN Block. (c) Illustration of the Unified MSRoPE (Multi-Scale Rotary Positional Embeddings) applied to VAE image latents and multimodal context embeddings.
    }
    \label{fig:overall_arch_head}
\end{figure}

\subsubsection{Visual Generation Head}

Based on the proposed principles, in this section, we further detail the implementation of our custom-developed visual generation head, as shown in~\cref{fig:overall_arch_head}. 

\noindent\textbf{Dual Projectors for Context and Target Input}. The feature distributions of the multimodal hidden states (\textit{context}) and the VAE's image latents (\textit{target}) exhibit substantially different feature distributions. To bridge this heterogeneity, we employ independent linear projectors to map them into the conditioning space of the visual generation module. Crucially, we observe that the multimodal context embeddings tend to exhibit larger magnitudes and more pronounced outliers than the VAE latents. To reduce this scale mismatch and improve training stability, we introduce an additional normalization layer on the VLM branch before projection, explicitly normalizing the variance of the context features to unity.

\noindent\textbf{Dual-Stream MMDiT Block with Gated Attention}. We adopt a fully Dual-Stream architecture to account for the distinct statistical properties of multimodal context and generative targets. While both streams interact via 
joint self-attention
to capture token-level dependencies, they utilize disentangled parameters for the QKVO projections and Feed-Forward Networks (FFNs). Furthermore, to enhance non-linearity and mitigate ``attention-sink'' phenomena in high-resolution, long-context scenarios, we integrate an element-wise Gating Mechanism~\cite{qiu2025gated} into the attention block. Formally, the modulated output \(\mathbf{O'}\) of the attention layer is:
\begin{equation}
    \mathbf{O'} = \mathbf{O} \odot \sigma(\mathbf{X}\mathbf{W}_g)
\end{equation}
where \(\sigma\) denotes the sigmoid function, \(\mathbf{X}\), \(\mathbf{O}\) denote the input and output of the attention layer, and \(\mathbf{W}_g\) denotes the learnable gating projection matrix, which is also disentangled for each stream. To the best of our knowledge, this is the first integration of a gating mechanism within the MMDiT architecture, offering improved expressivity with minimal parameter overhead.

\noindent\textbf{Unified MSRoPE with Resolution Interpolation}. We employ Multimodal Scalable RoPE (MSRoPE)~\cite{wu2025qwenimagetechnicalreport} to encode positional information, ensuring rigorous preservation of spatial structures.

\begin{itemize}
    \item \textbf{Unified 3D Encoding}: Unlike previous works~\cite{wu2025qwenimagetechnicalreport} that often treat visual tokens in the multimodal context as flattened 1D sequences, we apply unified 3D positional embeddings (temporal, height, width) to both the generative targets and the visual tokens within the context. This alignment significantly benefits tasks requiring precise spatial reasoning, such as image editing.
    \item \textbf{Positional Interpolation}: To facilitate resolution scaling, we address the ``tiling artifact'' observed when directly extrapolating position indices during high-resolution fine-tuning. Instead, we adopt a Resolution Interpolation strategy. We define the position embedding range based on the maximum target resolution (\eg, 1024px). During the initial low-resolution pre-training (\eg, 512px), rather than using a smaller index range, we utilize the full range but increase the stride between adjacent tokens. This ensures that the model learns a consistent global spatial representation from the outset, minimizing the domain gap when scaling to higher resolutions.
\end{itemize}

\subsection{Training Strategy}

\subsubsection{Training Objective}

To endow the UMM with the capability to process and generate multimodal content, we formulate a joint optimization objective. Given a multimodal context sequence $\mathbf{c}$, the model is trained to simultaneously predict discrete text tokens $\mathbf{x}$ and continuous image latent representations $\mathbf{z}$.

\noindent\textbf{Autoregressive Text Generation}. For the textual component, we treat text generation as a sequence modeling problem over a discrete vocabulary. We employ the standard Next-Token Prediction (NTP) objective, minimizing the negative log-likelihood of the target tokens conditioned on the context and preceding tokens:
\begin{equation}
    \mathcal{L}_{\text{NTP}} = - \frac{1}{T} \sum_{t=1}^{T} \log p_\theta(x_t \mid x_{<t}, \mathbf{c})
\end{equation}

where $x_t$ denotes the $t$-th token in the text sequence of length $T$, $x_{<t}$ represents the preceding tokens, and $\theta$ parameterizes the unified model. This objective ensures the model retains the reasoning and instruction-following capabilities inherent in the MLLM backbone.

\noindent\textbf{Flow Matching for Image Generation}. For the visual component, we adopt the Flow Matching framework with velocity parameterization to model the continuous distribution of image latents. Unlike diffusion models that predict noise $\epsilon$, we regress the velocity vector field $v_\theta$ that transports the probability density from a Gaussian noise distribution to the data distribution. 
We assume the standard linear interpolation path between the noise $\mathbf{z}_0 \sim \mathcal{N}(\mathbf{0}, \mathbf{I})$ and the ground-truth image latent $\mathbf{z}_1$, following the common formulation used in Flow Matching and transport paths inspired by Optimal Transport. 
The intermediate state at time $t \in [0, 1]$ is defined as $\mathbf{z}_t = t\mathbf{z}_1 + (1-t)\mathbf{z}_0$. The objective is to minimize the mean squared error between the predicted velocity and the target drift:

\begin{equation}
    \mathcal{L}_{\text{FM}} = \mathbb{E}_{t \sim \mathcal{U}[0,1], \mathbf{z}_0 \sim \mathcal{N}(\mathbf{0}, \mathbf{I}), \mathbf{z}_1 \sim p_{\text{data}}} \left[ \| v_\theta(\mathbf{z}_t, t, \mathbf{c}) - (\mathbf{z}_1 - \mathbf{z}_0) \|^2 \right]
\end{equation}

where $v_\theta(\mathbf{z}_t, t, \mathbf{c})$ is the model output predicting the velocity vector at time $t$ conditioned on context $\mathbf{c}$, and $(\mathbf{z}_1 - \mathbf{z}_0)$ represents the ground-truth instantaneous velocity along the linear trajectory.

\noindent\textbf{Unified Training Objective}. The final training objective is a weighted sum of the discrete and continuous losses:

\begin{equation}
    \mathcal{L}_{\text{Total}} = \alpha \cdot \mathcal{L}_{\text{NTP}} + \beta \cdot \mathcal{L}_{\text{FM}}
\end{equation}

where $\alpha$ and $\beta$ are scalar hyperparameters balancing the two modalities. In practice, we dynamically adjust these coefficients across different training stages (\eg, pre-training vs. supervised fine-tuning) to prioritize specific capabilities, such as visual fidelity or reasoning capabilities.

\subsubsection{Training Pipeline}

To maximize training efficiency while adhering to the architectural principles outlined in Section~\ref{sec:model_design}, we initialize our UMM from a pre-trained MLLM solely optimized for understanding tasks. Since the base MLLM lacks visual generative capabilities, we design a three-stage curriculum that progressively unlocks visual synthesis skills before unifying them with semantic reasoning.

\noindent\textbf{Stage1: Generation Head Pre-training}. In the initial phase, we focus on grounding the newly initialized visual generation head to the MLLM's latent space. We freeze the MLLM to preserve its semantic representations and train only the generation head and projectors. 
Following previous work~\cite{xie2024sana}, we skip the 256px pre-training and utilize a fixed resolution of 512px to accelerate early convergence.
Unlike prior approaches~\cite{wang2025ovis, wu2025qwenimagetechnicalreport} that rely solely on text-to-image data for initialization, we incorporate a mixture of text-to-image generation and image editing datasets from the outset. This multi-task strategy forces the generation head to attend to both textual instructions and visual context tokens simultaneously, establishing a robust foundation for multimodal condition alignment.

\noindent\textbf{Stage2: Any-resolution Continued Pre-training.}
Building upon the stable initialization, we advance to variable-resolution training to handle diverse aspect ratios and enhance visual fidelity. The MLLM backbone remains frozen.
We perform a secondary filtration of the training corpus, retaining only high-aesthetic samples and discarding those with extreme aspect ratios that might induce training instability.
The resolution of the generated images is controlled within the range of 512 to 1024 pixels, while the aspect ratio is maintained within the range of 0.5 to 2.0.
For image editing tasks, maintaining pixel-level alignment between the input condition and the output is critical. To this end, we further explicitly inject the VAE latent of the condition image into the visual generation head to achieve better pixel-level consistency. 

\noindent\textbf{Stage3: Unified Supervised Finetuning.}
The final stage aims to further synergize the visual generative capabilities acquired in previous stages with the reasoning capabilities of pre-trained MLLM.
Therefore, the entire model is unfrozen, including the MLLM backbone, to enable end-to-end optimization.
The training corpus is further filtered based on stricter criteria, along with additional CoT reasoning data, detailed in Section~\ref{sec:data-reasoning}. By mixing these CoT data with image generation and editing data, the model is allowed to plan its generation via textual reasoning before executing it in the visual domain.

%% file: sections/4.data.tex
\section{Data Construction}

\input{tables/data-opensource}

To equip \modelname with multimodal generation and editing capabilities on top of InternVL's strong multimodal understanding foundation, we construct a large-scale training corpus by combining publicly available datasets with synthetic data pipelines tailored to diverse generation and editing tasks.

\subsection{Open-source Data Collection}

As shown in \cref{tab:tasks_datasets}, we have collected a large number of high-quality image generation and image editing datasets as our initial data pool.
To better address long-tail cases, particularly in the domains of human portraits and text-rich imagery, we further augment this open-source corpus with specialized datasets.

\subsection{General Data Preprocessing and Synthesis}

\begin{figure}[htbp]
    \centering
    \includegraphics[width=0.95\linewidth]{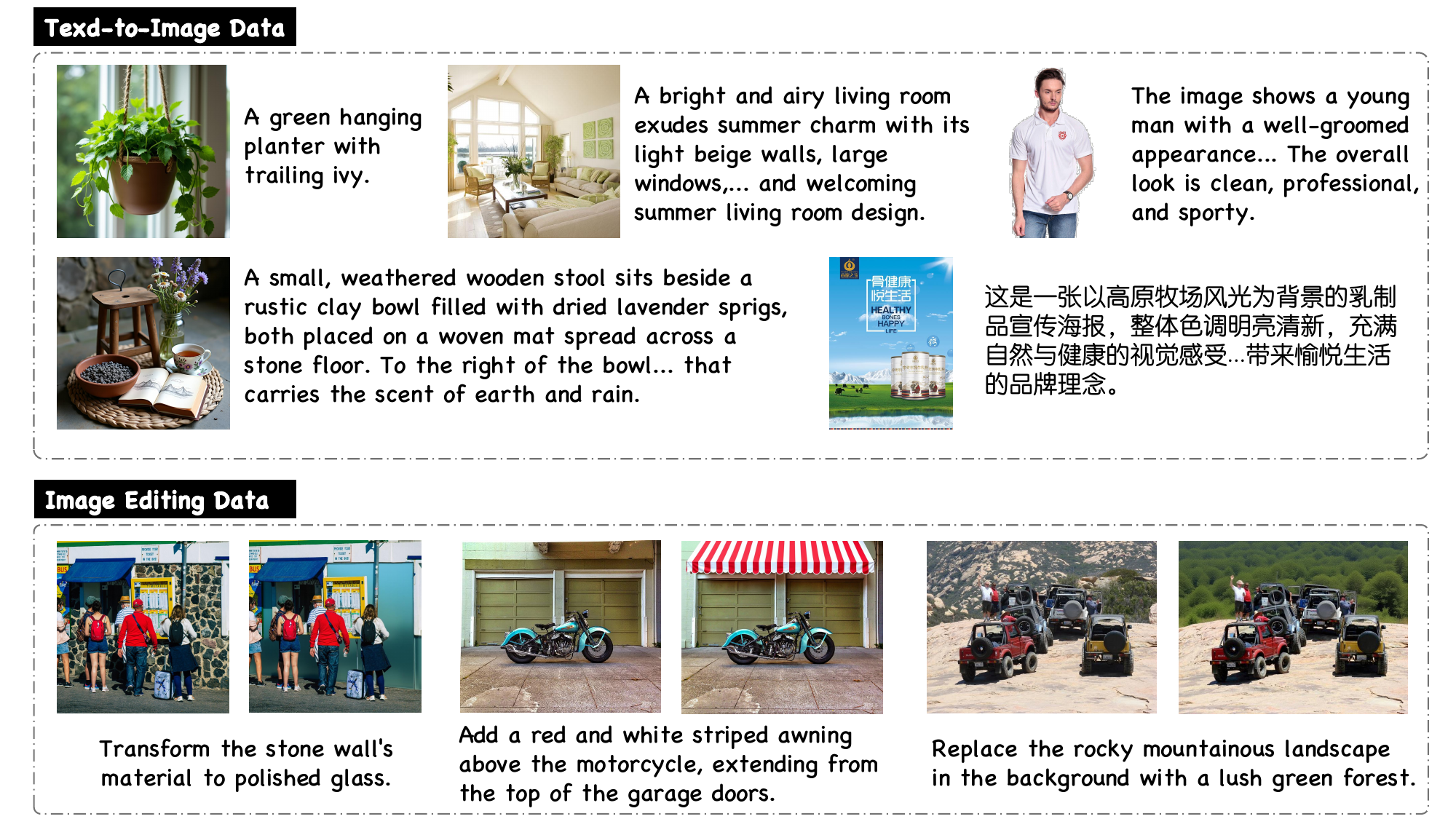} 
    \caption{
    \textbf{Examples of general data synthesized by our pipeline.}
    The synthesized data features varied textual annotations and covers diverse visual domains, including portraits, posters, natural scenes, \etc.
    }
    \label{fig:general_data_example}
\end{figure}

We first design the general data preprocessing and synthesis pipelines for image generation and image editing. Representative examples are shown in \cref{fig:general_data_example}.

\begin{figure}[htbp]
    \centering
    \includegraphics[width=0.85\linewidth]{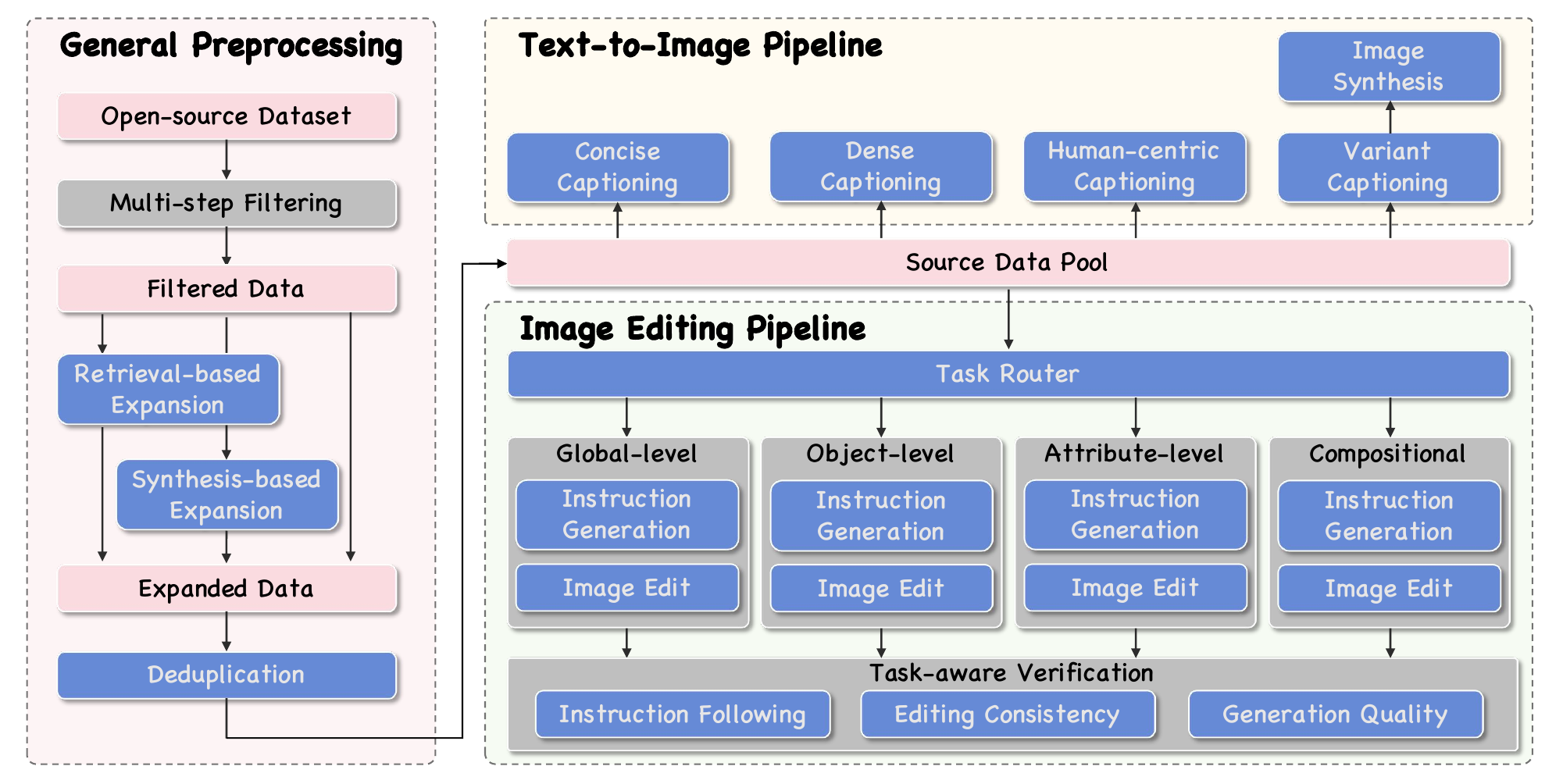} 
    \caption{\textbf{Overview of our general data synthesis pipeline.} 
    First, the preprocessing stage applies filtering, expansion, and deduplication to construct a high-quality source pool. Building upon this, two parallel branches are deployed to generate text-to-image pairs and instruction-guided editing data, respectively.
    }
    \label{fig:general_data_pipe}
\end{figure}

\subsubsection{General Preprocessing} 

Given the collected open-source datasets, we first apply a general preprocessing pipeline for filtering, expansion, and deduplication, as illustrated in \cref{fig:general_data_pipe}.
Specifically, we begin with a rigorous multi-dimensional filtering protocol to exclude low-quality samples. This process includes filtering based on aesthetic scores, resolution thresholds, safety standards (\eg, NSFW detection), and watermark identification, yielding a pristine subset of high-quality samples.

To enrich domain coverage and intra-domain diversity, a dual-branch expansion workflow is then implemented: retrieval-based and synthesis-based expansion. In the retrieval-based branch, we capture long-tail concepts and real-world variations absent in standard datasets by utilizing both image and text queries in large-scale search engines. Complementing this, the synthesis-based branch densifies the image manifold by creating realistic variants of existing samples. Finally, to ensure a non-redundant source pool, we compute perceptual hashes (p-hash)~\cite{zauner2010implementation} for all collected samples, removing near-duplicates to maximize data efficiency.

\subsubsection{Text-to-Image Data}

As shown in \cref{fig:general_data_pipe}, different captioning strategies are adopted to enhance the diversity and quality of the captions of the collected data.
Specifically, a pre-trained MLLM (\ie, Qwen2.5-VL~\cite{Qwen2.5-VL}) is adopted as the image captioning agent and is prompted to generate captions at varying levels of granularity, including: 
\begin{itemize}
    \item \textbf{Concise Captioning}: Short, lucid descriptions of core visual elements, facilitating strong concept binding and prompt adherence.
    \item \textbf{Dense Captioning}: Hierarchical descriptions to ensure the model attends to holistic scene structure and details, covering foreground subjects, background environments, ‌and stylistic features.
    \item \textbf{Human-Centric Captioning}: Designed specifically for portraiture, focusing on fine-grained attributes such as facial features, expressions, poses, and clothing details.
\end{itemize}
To address the long-tail scarcity of images containing legible text, we further introduce a targeted data expansion strategy tailored for text-rich data. 
Given specific textual content and background environments, variant captions are first generated by the image captioner. Then an image generation expert (\eg, Qwen-Image~\cite{wu2025qwenimagetechnicalreport}) is employed to synthesize corresponding images. This process significantly densifies the distribution of high-quality text-image pairs.
To equip the model with robust bilingual capabilities, a comprehensive English-to-Chinese translation pipeline is also conducted across the dataset, ensuring the model can interpret and generate content with equal proficiency in both languages.

\subsubsection{Image Editing Data}

For image editing data synthesis, a more sophisticated pipeline is further introduced, as illustrated in \cref{fig:general_data_pipe}.
We first categorize the editing tasks into four primary classes: 
\textbf{(1) Global-level}: Modifying overall style, tone, or background while preserving the structural layout.
\textbf{(2) Object-level}: precise addition, removal, or replacement of objects with boundary handling.
\textbf{(3) Attribute-level}: Adjusting fine-grained properties such as color, material, size, and count.
\textbf{(4) Compositional}: Executing compound instructions that require coherent, multi-step operations.

Given the source data pool, we then adopt a multi-agent framework to generate instruction-edit pairs. 
An MLLM-based router is responsible for determining the specific editing tasks for each source image. Based on the routing results, images are dispatched to a modular pool of specialized agents. 
To ensure high-quality synthesis, we stratify these agents into two functional categories: Instruction Generation and Image Editing.

\begin{itemize}
    \item \textbf{Instruction Generation}: Acknowledging that distinct editing tasks require capturing different levels of semantic granularity, we implement these agents using Qwen2.5-VL-72B~\cite{Qwen2.5-VL}, conditioned with task-specific prompts to produce precise and context-aware instructions.
    \item \textbf{Image Editing}: Since existing open-source models~\cite{wu2025qwenimagetechnicalreport, labs2025flux1kontextflowmatching, liu2025step1x-edit} exhibit varying strengths due in part to differences in their training data distributions, 
    we integrate a heterogeneous ensemble of editing models. By assigning different tasks to the model best suited for that specific granularity, we ensure optimal visual fidelity across the entire dataset.
\end{itemize}

To ensure the reliability of the constructed corpus, an automated, task-aware verification module is also introduced, as shown in \cref{fig:general_data_pipe}. We define a tripartite evaluation protocol assessed by MLLMs:
\textbf{(1) Instruction Following}: Verifying if the edited image faithfully executes the prompt.
\textbf{(2) Editing Consistency}: Evaluating semantic and structural coherence with the source image.
\textbf{(3) Generation Quality}: Assessing visual fidelity, realism, and artifact suppression.
Each image-editing pair is scored against task-specific prompts. Only samples exceeding predefined thresholds across all three dimensions are retained. This adaptive filtering effectively removes misaligned or low-quality examples, resulting in a robust dataset characterized by strong instruction alignment and visual coherence.

\subsection{Text-centric Data Synthesis}

\begin{figure}[htbp]
    \centering
    \includegraphics[width=0.95\linewidth]{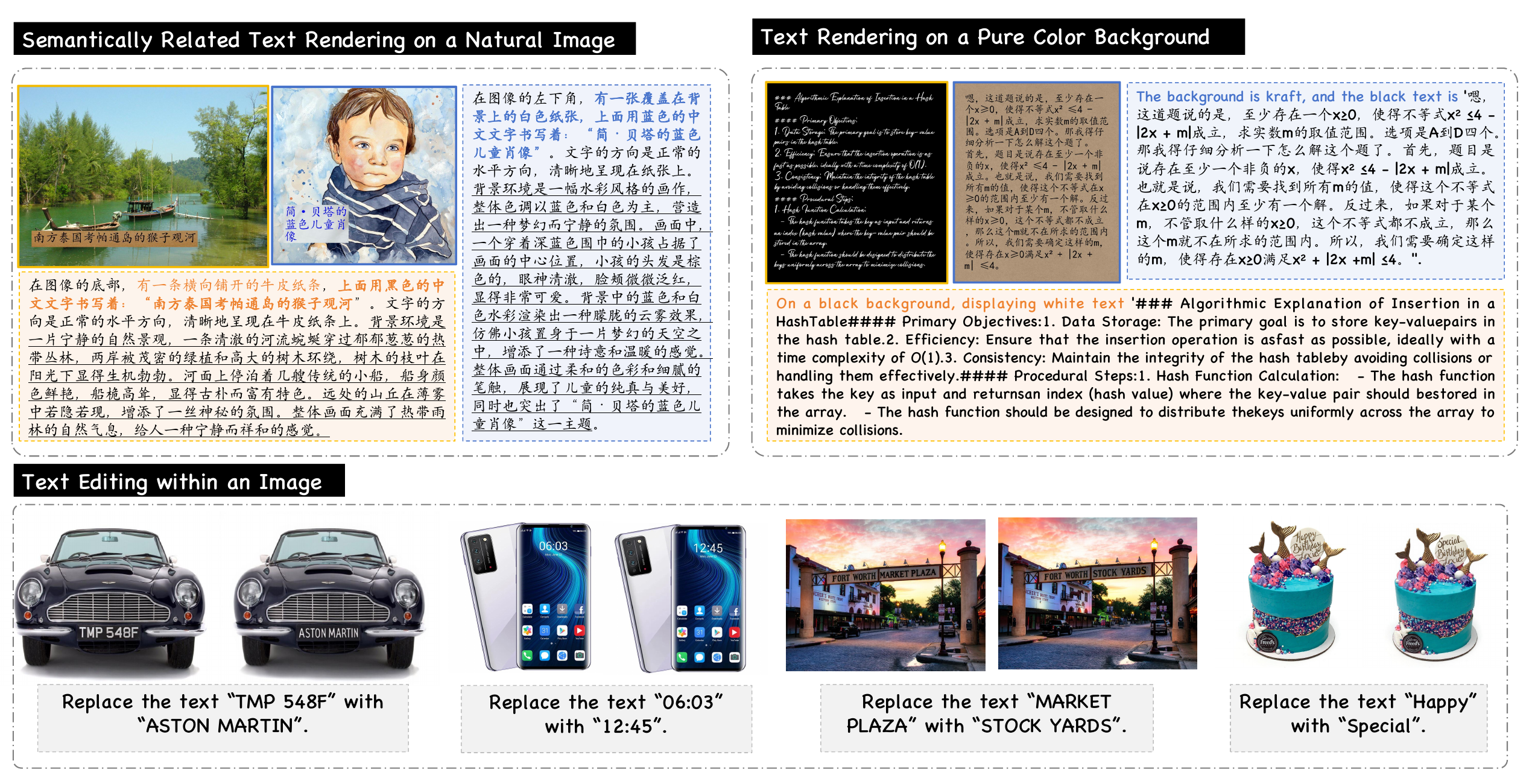} 
    \caption{\textbf{Three types of text-centric data synthesized by our pipeline.} The first type overlays semantically relevant text onto natural images using a masked background image.
The second type renders text on solid-color backgrounds, focusing on clean and aesthetically pleasing layouts.
The third type performs text editing within existing images, such as modifying text on license plates, mobile interfaces, signboards, and similar surfaces.}
    \label{fig:text-data-show}
\end{figure}

Textual elements in visual media are highly semantically dense and crucial for communication, making accurate text rendering and fine-grained editing essential for real-world applications.  Despite recent advances in text-related tasks~\cite{cui2025paddleocr30technicalreport,lan2025flux,tuo2023anytext,wu2025qwenimagetechnicalreport,qian2024anytrans}, general multimodal models still struggle with text-centric generation and editing, often exhibiting spelling errors, poor support for non-alphabetic languages, layout misalignment, and unintended visual artifacts.
Considering its significance,
we further introduce a text-centric data synthesis pipeline aimed at further improving text rendering and editing capabilities. As shown in \cref{fig:text-data-show}, we cover three representative data types: (1) semantically related text rendering on natural images, (2) text rendering on pure-color backgrounds, and (3) text editing within images, which together strengthen a model's integrated ability to understand, generate, and modify textual content in visual contexts. 

\begin{figure}[htbp]
    \centering
    \includegraphics[width=0.9\linewidth]{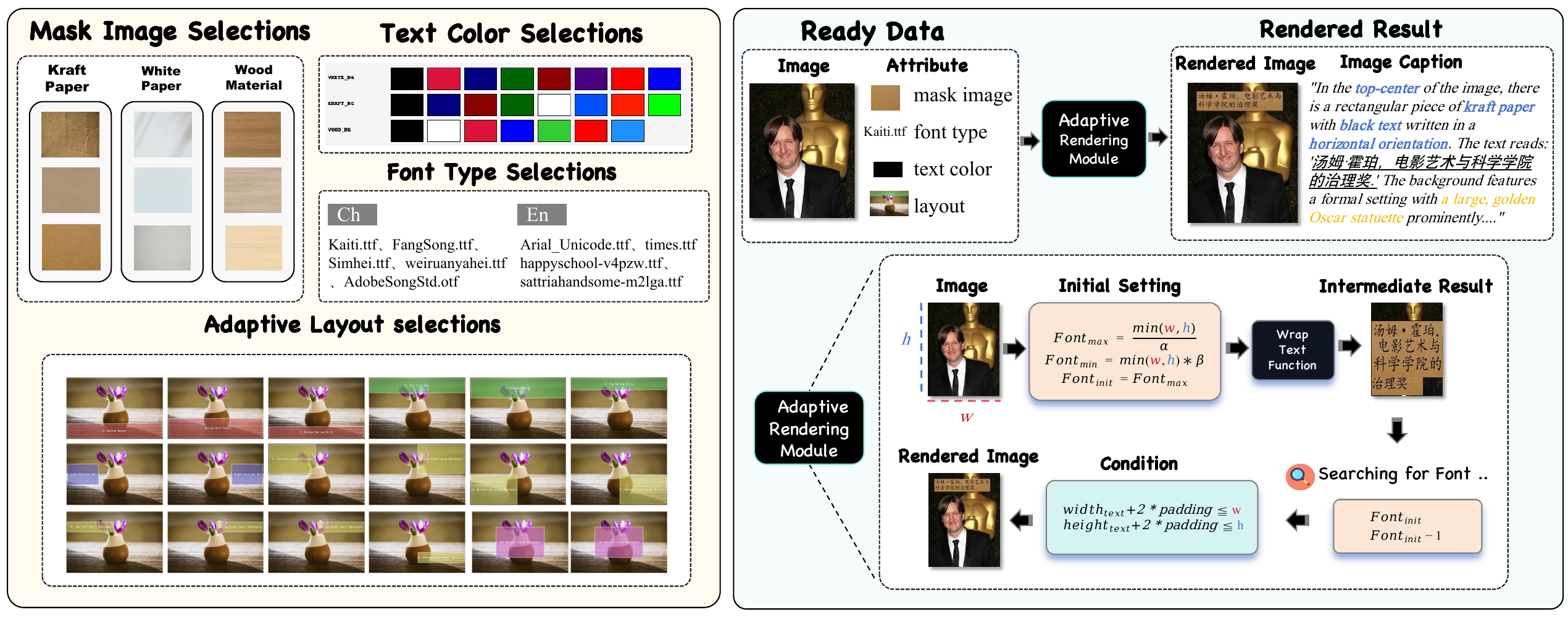} 
    \caption{\textbf{Text rendering data construction pipeline.} For synthetic text rendering, we prepare mask images, font colors, font styles, and adaptive layout options. During the rendering process, these attributes are randomly sampled, and the text is rendered with typography that adapts to its length.
    }  
    \label{fig:text_render}
\end{figure}

\subsubsection{Text-to-Image Data}
To equip the model with a stronger visual text rendering capability, we design a comprehensive automatic text-rendering data synthesis pipeline that can produce high-quality and diverse textual data. This pipeline supports both semantically related text rendering on natural images and text rendering on pure-color backgrounds, covering different languages (\ie, both Chinese and English).

Specifically, as shown in \cref{fig:text_render}, for semantically related text rendering on natural images, we directly render the original paired caption annotation onto the source image. To increase the text diversity, we also render pure text data on a randomly picked pure-color background.
During rendering, the mask of the rendering region, the text color and font type, as well as an adaptive layout design scheme, are also taken into account. The text size can be further adaptively adjusted with line breaks inserted automatically to ensure that the text is arranged aesthetically on the image. Finally, an image captioner (\ie, Qwen2.5-VL-72B~\cite{Qwen2.5-VL}) is used to re-caption the rendered images.

\subsubsection{Image Editing Data}

\begin{figure}[htbp]
    \centering
    \includegraphics[width=0.85\linewidth]{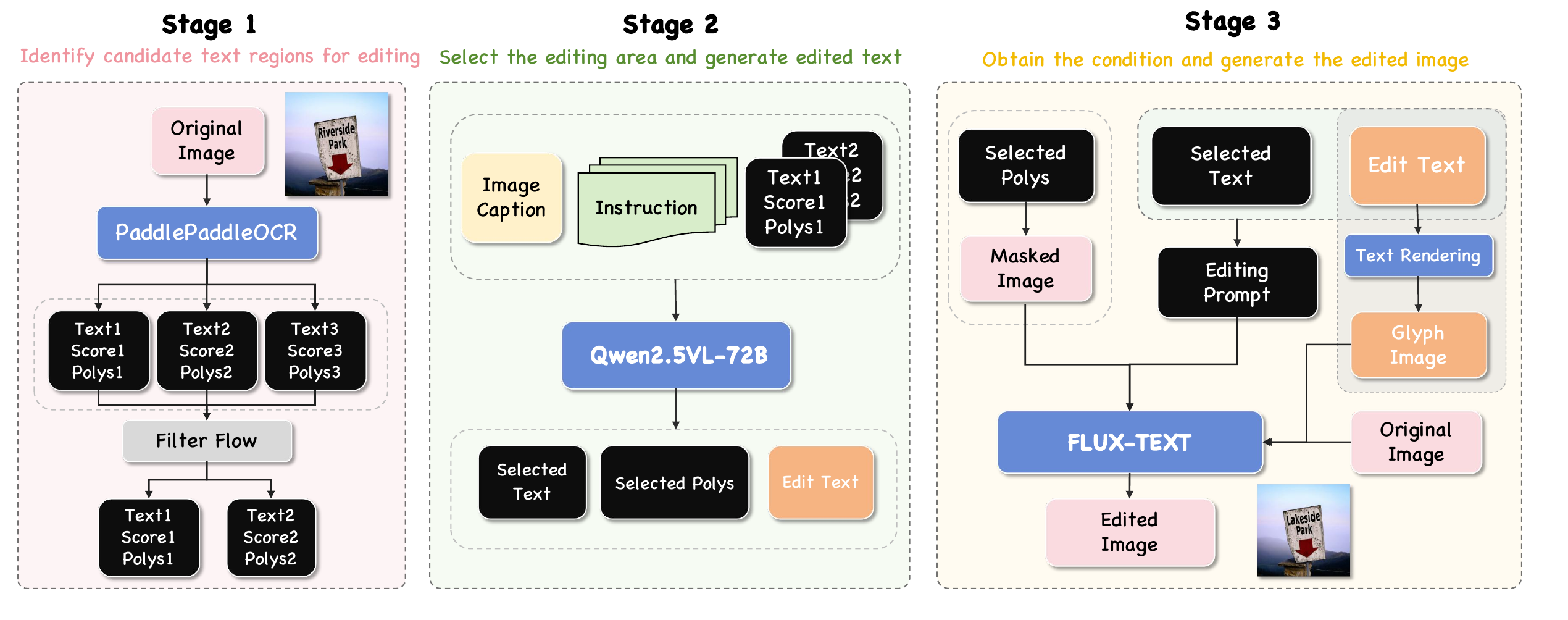} 
    \caption{\textbf{Text editing data construction pipeline.} First, we use an OCR tool to extract candidate text regions for editing. Second, we generate the editing instructions. Third, we use a generative model to produce the edited ground truth. Through these three steps, we synthesize high-quality text-editing triplets.
    }
    \label{fig:text-edit-data}
\end{figure}

For synthesizing text-aware image editing, we design a three-stage pipeline, as shown in \cref{fig:text-edit-data}. This pipeline can generate high-quality paired samples covering text edits in both natural and virtual scenes. 
First, we employ OCR tools (\ie, PaddleOCR~\cite{cui2025paddleocr30technicalreport}) to detect text regions and extract the recognized text, confidence scores, and bounding polygons. Second, an MLLM-based instruction agent (\ie, Qwen2.5-VL-72B~\cite{Qwen2.5-VL}) is prompted to filter candidate regions, verify textual relevance and visual consistency, and produce semantically explicit editing instructions. 
Finally, these selected texts, polygons, and editing instructions are then taken as inputs for a text-editing agent~(\eg, Flux-Text~\cite{lan2025flux}) to perform precise, context-aware text editing. This workflow yields high-quality, semantically aligned image-text editing pairs suitable for text-aware applications.

\subsection{Science-centric Data Synthesis}

\begin{figure*}[t]
    \centering
    \includegraphics[width=0.85\linewidth]{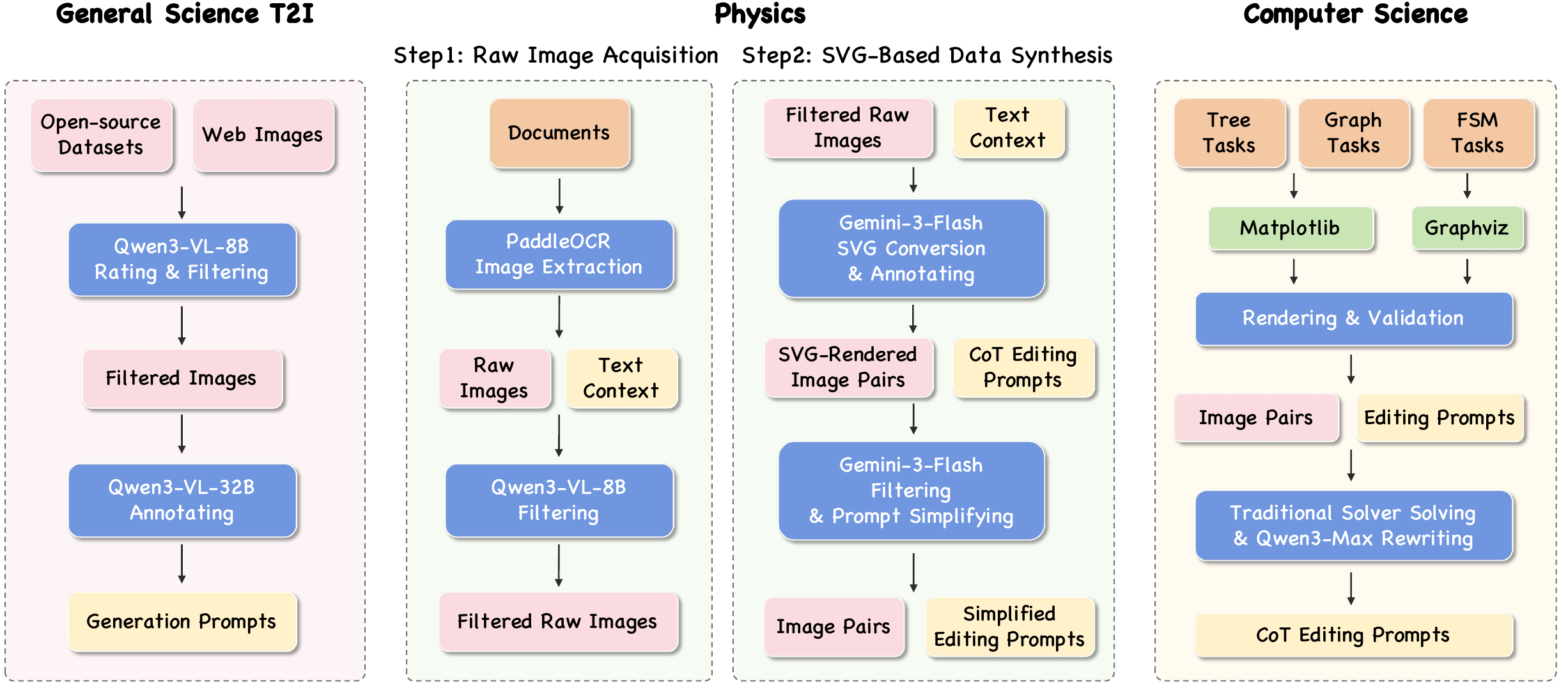}
    \caption{
\textbf{Science data generation pipeline.}
For general science T2I, we collect web images and open-source datasets and design automatic filtering and annotating with open-source models. For physics, we obtain images from documents by PaddleOCR, and propose an SVG-based pipeline for high-quality and affordable image-pair generation. For computer science, we define tasks and render images with Python libraries.
    }
    \label{fig:science-data-pipeline}
\end{figure*}

\begin{figure*}[t]
    \centering
    \includegraphics[width=0.9\linewidth]{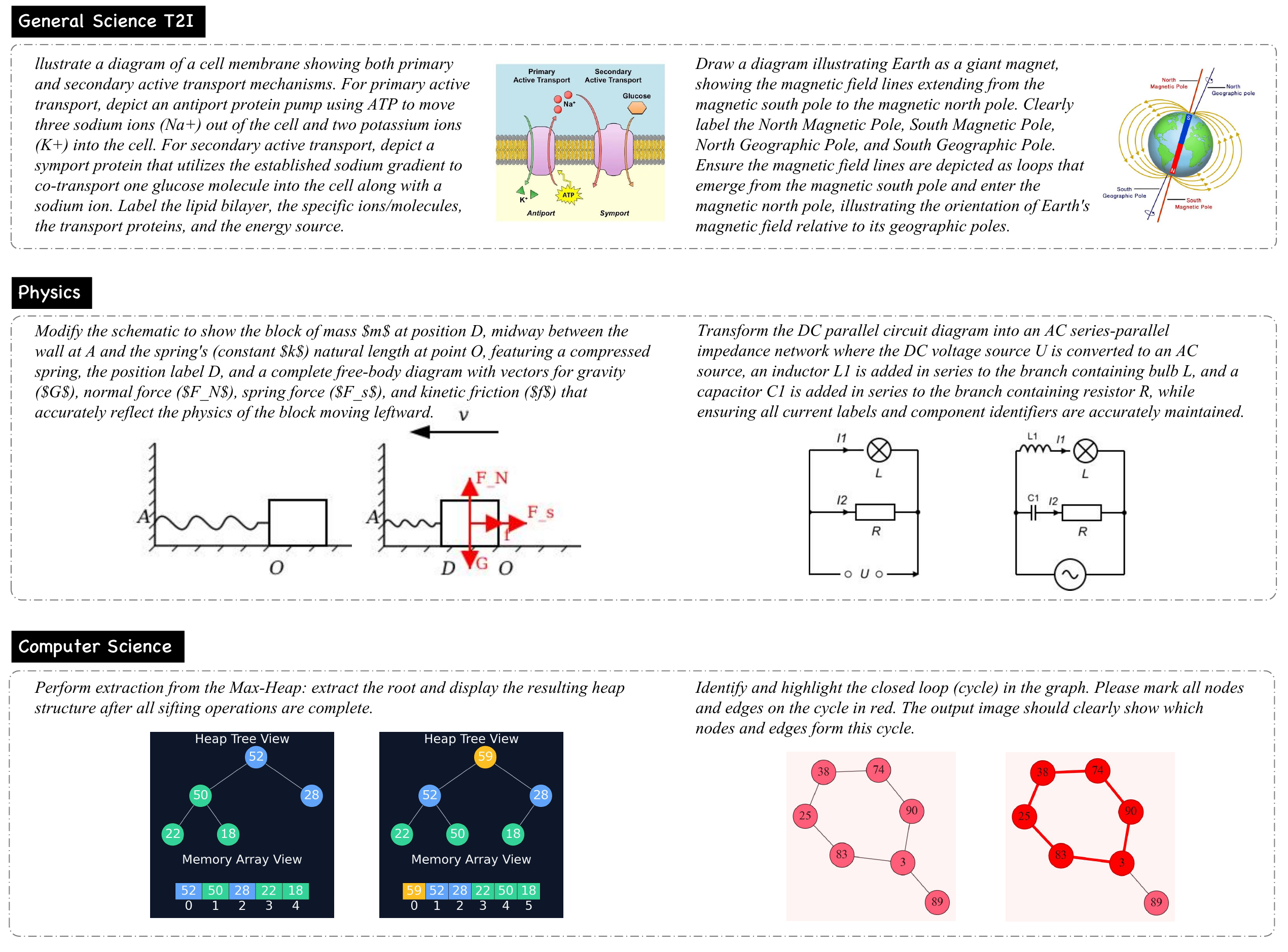}
    \caption{
    \textbf{Examples of science-centric data.}
    General science T2I features dense and detailed instructions and requires depicting multidisciplinary concepts.
    Physics and computer science focus on image editing with disciplinary reasoning.
    }
    \label{fig:science-data}
\end{figure*}

Scientific images are of great importance to both the science academia and the AI industry. 
While the understanding of science-centric images has attracted wide attention~\cite{MMMU, hu2025surveyscientific, scientist_first_exam, xu2025probing}, their generation is still in a relatively preliminary stage~\cite{GenExam, sridbench}. 
To enhance the model's capability for generating images with strict structuredness, high semantic coherence, rigid knowledge dependence, and deep reasoning,
we design various rigorous pipelines to curate science-centric data for text-to-image and image editing. These data cover various disciplines such as physics, chemistry, biology, and computer science.
Specifically, text-to-image data are mainly curated from existing understanding datasets and web images.
For editing data, which are more difficult to collect as they require input-output image pairs with diverse and meaningful editing prompts, we design multiple data engines to synthesize editing data for 
physics and computer science. The pipeline of science data generation is shown in \cref{fig:science-data-pipeline}, and
examples of science data are given in \cref{fig:science-data}.

\subsubsection{General Science Generation Data}

For text-to-image generation, the images are collected from various sources, including open-source multimodal scientific understanding datasets (\cite{MMMU,AI2D,kembhavi2017tqa,mv-math,zhang2024mavis,liu2024cmm,huang2024chemeval,chebi,chemQA2024}), textbooks, and competitions (\eg, IPHO, Gaokao, Kaoyan).
After that, a multi-stage filtering strategy is adopted to obtain high-quality data. In the first stage, images with resolutions below 256p are removed, and the resulting images are deduplicated based on p-hash. 
Images identical to the benchmarks in Section~\ref{sec:experiments} are also removed to avoid data contamination. 
In the second stage, we employ an open-source MLLM (\ie, Qwen3-VL-8B~\cite{Qwen3-VL}) to rate and filter the images from multiple dimensions, including image types, subjects, text length, image complexity, and subject knowledge density.
Please refer to Section~\ref{sec:appendix-general-science} for more filtering details. 
An image captioner (\ie, Qwen3-VL-32B~\cite{Qwen3-VL}) is then prompted to synthesize the corresponding caption.

\subsubsection{SVG-based Physics Editing Data}

Although abundant physics images are abundant on the Internet, it is difficult to obtain \textit{paired} physics images for editing from open-source datasets or synthesizing with existing libraries and software. Using an existing proprietary image editing model (\eg, Nano Banana Pro~\cite{deepmind_gemini3proimage_2025}) to generate a paired image from a given image can lead to extremely high costs and inconsistent quality.
Inspired by recent advances in Scalable Vector Graphics (SVG) understanding and generation~\cite{wang2025internsvg, sgpbench}, 
as well as the strong capability of state-of-the-art multimodal models~\cite{Gemini-3-Flash, GPT-5}, 
we design an SVG-based pipeline to synthesize physics image pairs, as shown in \cref{fig:science-data-pipeline}. SVG enables high-quality and resolution-independent generation of target images in a cost-effective manner, by manipulating structured SVG code rather than editing raster images directly.

In the first stage, we collect heterogeneous-source documents of physics textbooks and exams, and use PaddleOCR~\cite{paddleocr3} to extract images and their textual context. We then use Qwen3-VL-8B~\cite{Qwen3-VL} to filter out images of undesirable domains, \eg, illustrative figures in textbooks, real-world photographs in exam questions, mathematical formulas, and complex images that are difficult to convert into SVG format. 
These images cover sub-disciplines such as electromagnetism, mechanics, optics, circuits, thermology, and atomic physics.

Subsequently, an image and its context are fed into Gemini-3-Flash~\cite{Gemini-3-Flash} to generate the corresponding SVG code of the original image. The model also decides whether the original image should serve as the input or output image, and generates an editing prompt and the SVG code of the other image. The two SVG codes are rendered into a pair of images. 
The original prompts generated by Gemini-3-Flash~\cite{Gemini-3-Flash} are detailed and explicit and can serve as a step-by-step reasoning prompt. To evaluate subject-specific knowledge and competence, we perform key information extraction and summarization on these prompts to obtain a simplified, implicit prompt as the final data.

Finally, another filtering is performed on each triplet of <input\_image, prompt, output\_image> to examine the correctness and quality of the data. 
We use Gemini-3-Flash~\cite{Gemini-3-Flash} to filter and remove images containing rich text formatting and images with physical structural errors or annotation errors. 
Compared to directly using Nano Banana Pro~\cite{deepmind_gemini3proimage_2025} to generate the output image, this SVG-based approach greatly reduces the cost from \$0.16 to \$0.03 per sample.

\subsubsection{Computer Science Editing Data}
We construct a data engine for computer science editing data based on several Python libraries. These data focus on operations and algorithms on data structures like trees, graphs and finite state machines (FSM). Specifically, the tasks include:

\begin{itemize}
    \item \textbf{Tree}: Topology editing and node manipulation, traversal visualization, binary search tree (BST) operations, heap operations with dual views, Huffman coding trees, and lowest common ancestor (LCA) and path highlighting.
    \item \textbf{Graph}: K-hop neighborhood identification, degree identification, cycle detection, bipartite graph coloring, shortest path reasoning, and directed graph reachability.
    \item \textbf{FSM}: String tracing, state role identification, and transition logic completion.
\end{itemize}
Definitions of these tasks are in Section~\ref{sec:appendix-computer-science}.

To ensure high-quality data generation, we first select task-specific rendering engines based on task complexity to balance efficiency and visual fidelity. For structurally simpler tasks such as trees and graphs, we employ matplotlib to expedite the generation process. In contrast, for state machines requiring dense information display, we utilize Graphviz (with circo layouts) to ensure optimal topological clarity.
To maintain spatial consistency across image pairs (\eg, the position of a node in the images should not change), we define fixed anchor points for nodes. This constraint ensures that invariant components remain visually consistent.

We then perform validation to detect and eliminate samples with node or edge overlaps by calculating the distance between two nodes or between a node and an edge. Any generated instance failing this occlusion check is discarded, ensuring the overall structural integrity of the dataset. Finally, we generate CoT editing prompts by solving with a traditional solver and rewriting, detailed in Section~\ref{sec:data-reasoning}.

\begin{figure*}[t]
    \centering
    \includegraphics[width=0.9\linewidth]{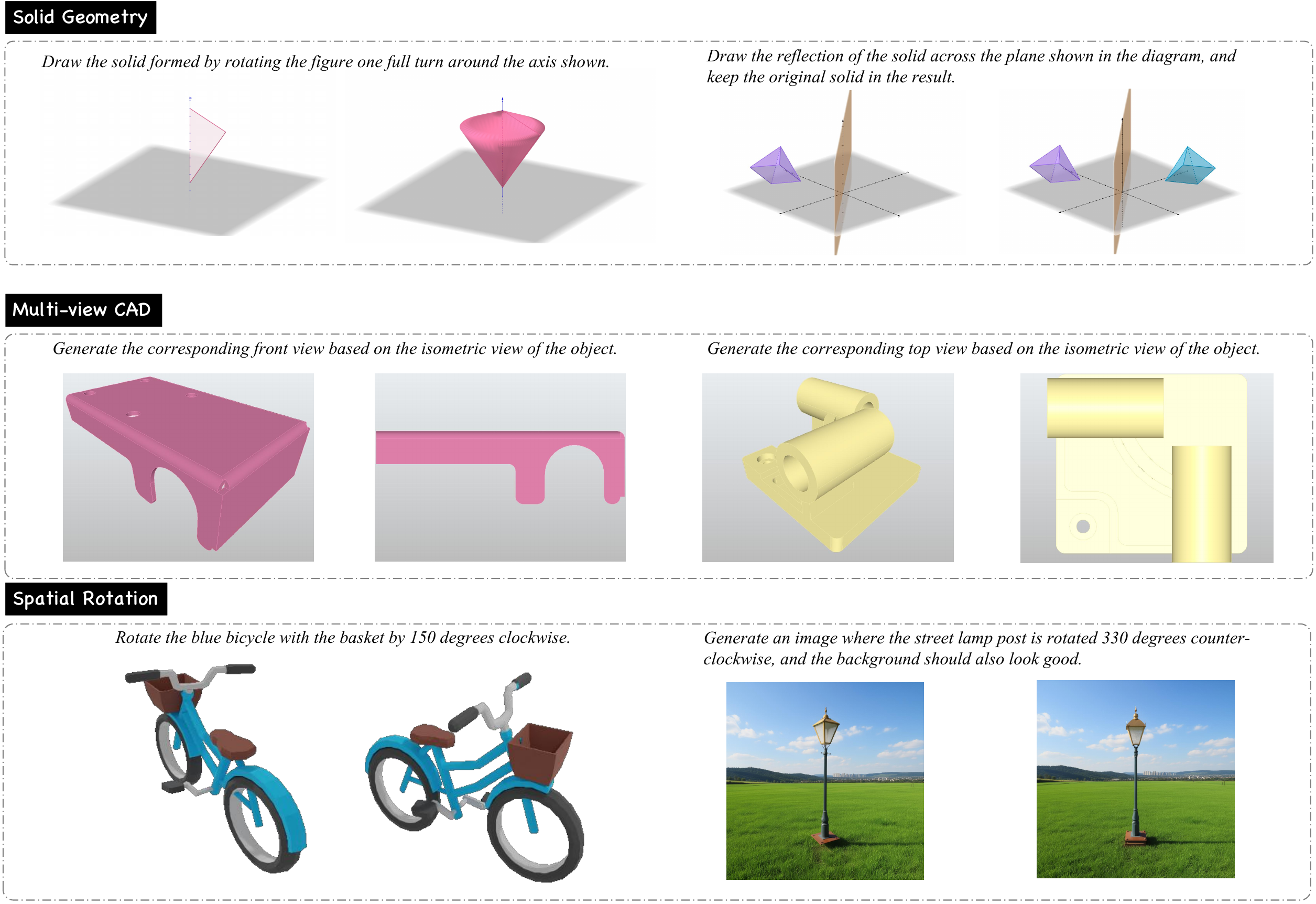}
    \caption{
    \textbf{Examples of spatial-centric data. }
    We consider three spatial-centric scenarios: solid geometry (\eg, revolution, symmetry), multi-view CAD (three-view drawing), and spatial rotation of 3D objects.
    }
    \label{fig:spatial-data}
\end{figure*}

\subsection{Spatial-centric Data Synthesis}

To enhance the spatial understanding capability of our model, we synthesize spatial-centric data from three distinct domains: solid geometry, multi-view CAD, and general spatial rotation. Example images are shown in \cref{fig:spatial-data}.

\subsubsection{Solid Geometry Editing Data}
We use GeoGebra~\cite{geogebra} and matplotlib to render solid geometry editing data.
The tasks include:

\begin{itemize}
    \item \textbf{Solid of Revolution}: Draw a solid formed by rotating a given shape around a given axis.
    \item \textbf{Plane Symmetry}: Draw the solid that is centrally symmetric to a given solid with respect to a given plane.
    \item \textbf{Point Symmetry}: Draw the solid that is centrally symmetric to a given solid with respect to a given point.
    \item \textbf{Solid Translation}: Translate a given solid by a given translation vector.
    \item \textbf{Solid Projection}: Draw the orthogonal projection of a given solid onto the \(xoy\)-plane.
\end{itemize}

Implementations of these tasks are in Section~\ref{sec:appendix-computer-science}.

\subsubsection{Multi-view CAD Editing Data} 

To further enhance the model's spatial understanding capabilities, we constructed corresponding three-view editing data based on an open-source CAD dataset. The model was required to predict other corresponding views based on the input isometric view and editing instructions. Specifically, we used the open-source ABC dataset~\cite{koch2018abc} and the OCC Python library to render the CAD files in the dataset into images of corresponding perspectives, including isometric views, front views, side views, and top views. We also increased the diversity of data distribution by randomly setting the colors and materials of objects during the rendering process. 
Examples of the final constructed editing dataset are shown in \cref{fig:science-data}.

\subsubsection{Spatial Rotation Editing Data}

\begin{figure}[htbp]
    \centering
    \includegraphics[width=0.9\linewidth]{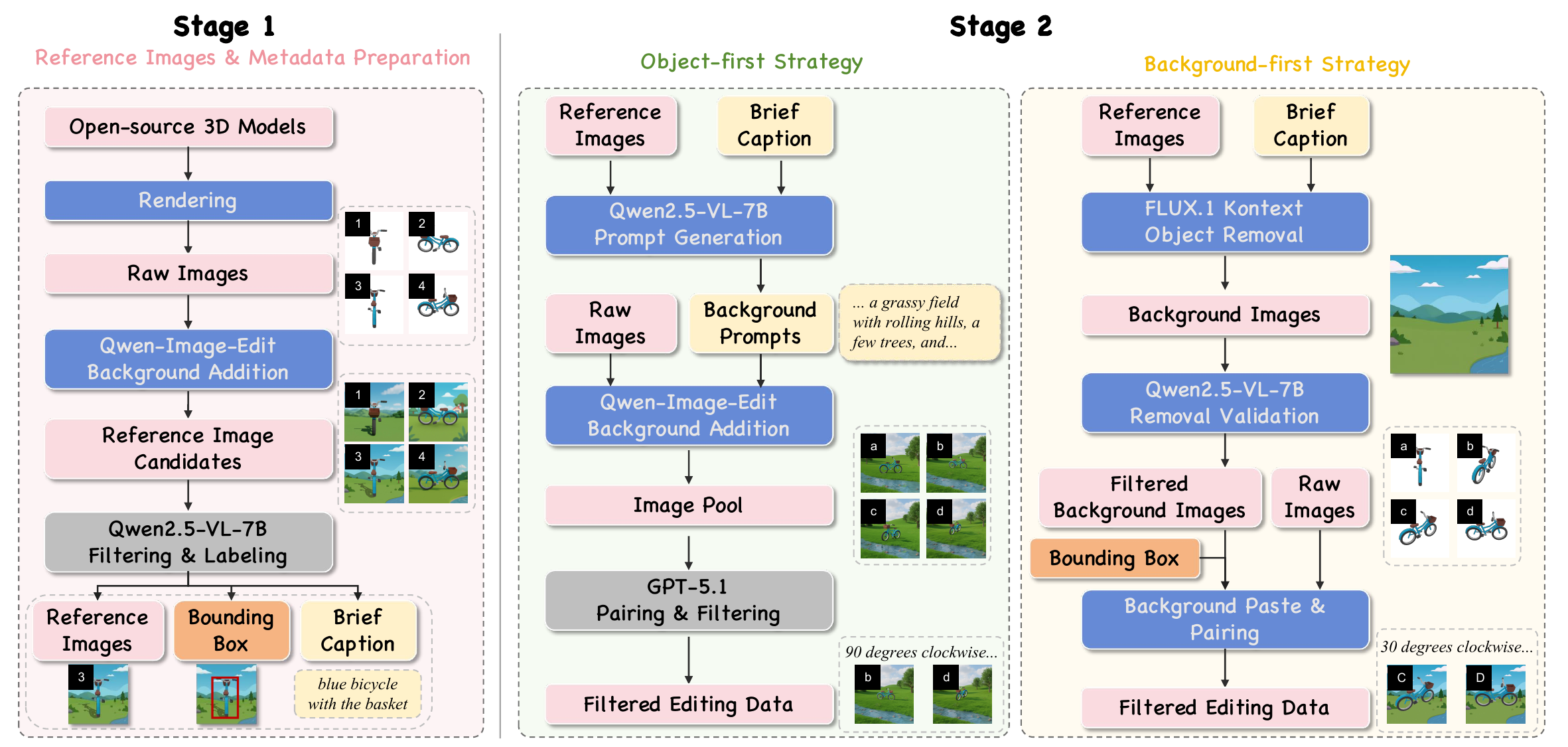} 
    \caption{
    \textbf{Overview of our spatial rotation editing data synthesis pipeline.}
    Stage 1 prepares a pool of filtered reference images. Subsequently, Stage 2 generates the final editing pairs via either the Object-First strategy for object-context integration, or the Background-First strategy for strict background preservation.
    }
    \label{fig:rotation-edit-data}
\end{figure}

We leverage Objaverse~\cite{deitke2023objaverse}, an open-source 3D model dataset, to build high-quality object rotation data. We begin by rotating the objects at uniform angles and rendering. While covering a wide range of daily objects, it typically lacks rich environmental contexts. As illustrated in \cref{fig:rotation-edit-data}, to endow these objects with suitable backgrounds, we initially generate four distinct reference images featuring backgrounds for each object. These candidates then undergo a rigorous multi-step filtering mechanism:
\begin{itemize}
    \item \textbf{Bounding Box Detection}: By detecting the object's bounding box, we assess deformation, retaining only instances where the aspect ratio falls within the range of $[0.9, 1.1]$. Additionally, to ensure the feasibility of the Background-First strategy, we exclude candidates where the object would exceed the image boundaries during rotation. 
    \item \textbf{Object Consistency}: We evaluate the consistency of essential visual attributes, with a particular emphasis on orientation, and retain only those instances that demonstrate high consistency to the original input.
    \item \textbf{Generation Quality}: We assess the plausibility of object-background integration, stylistic consistency with the scene, and overall visual quality, and select the highest-scoring candidate for each object.
    \end{itemize}
Subsequently, we design two different strategies for the final data synthesis: 
\begin{itemize}
    \item \textbf{Object-First strategy}, which prioritizes the plausible integration of the object within its context. 
    We first generate context-rich instructions based on the selected reference image. The instructions together with the reference image are then used by Qwen-Image~\cite{wu2025qwenimagetechnicalreport} to synthesize the edited image. To ensure the high quality of these editing pairs, we utilize GPT-5.1 to further filter results based on two dimensions: the accuracy of the object's new orientation and the consistency of the background.
    \item \textbf{Background-First strategy}, which prioritizes maximal background consistency across edits. Leveraging the strong consistency capabilities of Flux.1 Kontext~\cite{labs2025flux1kontextflowmatching}, we perform object removal to obtain a clean background image. The success of this removal operation is verified using Qwen2.5-VL~\cite{Qwen2.5-VL}. Next, utilizing the bounding boxes detected before, we paste objects with alternative orientations onto this clean background, thereby yielding editing pairs with perfectly consistent background scenery.
\end{itemize}

\begin{figure}[htbp]
    \centering
    \includegraphics[width=0.9\linewidth]{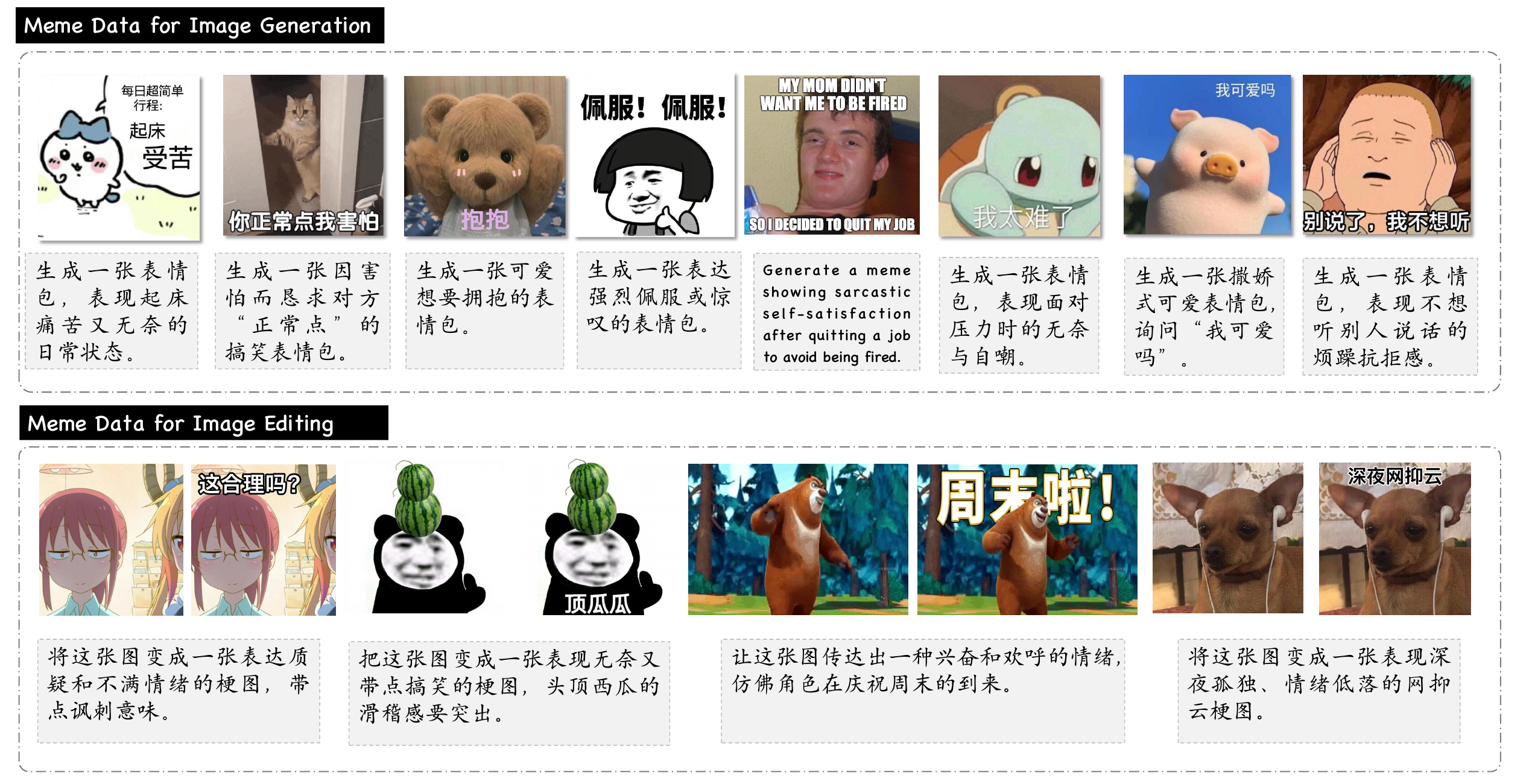} 
    \caption{\textbf{Meme data example for image generation and editing.} The meme data here captures elements of everyday human humor and expressive nuances commonly found in daily life.}
    \label{fig:meme-data}
\end{figure}

\subsection{Humor-centric Data Synthesis}

\label{meme}

Memes have become an important form of expression on the Internet, conveying humor, satire, and cultural information through the combination of visual and textual elements. 
To enhance the meme generation capability of our model, 
we synthesize two types of training data tailored to memes: text-to-image generation data and image-to-image editing data. The examples are shown in \cref{fig:meme-data}. 
The generation data targets scenarios where users provide short and abstract intent prompts, and the system produces an image that conveys concrete or implicit meanings, such as humor, sarcasm, surprise, helplessness, or other affective intentions. 
The editing data targets scenarios where a user supplies an input image and modifies it via natural-language instructions, while preserving the original content and style as much as possible. A common form is adding concise, punchy text (\eg, subtitles, labels, or dialogue) to create contrast or irony.
\begin{figure}[htbp]
    \centering
    \includegraphics[width=0.95\linewidth]{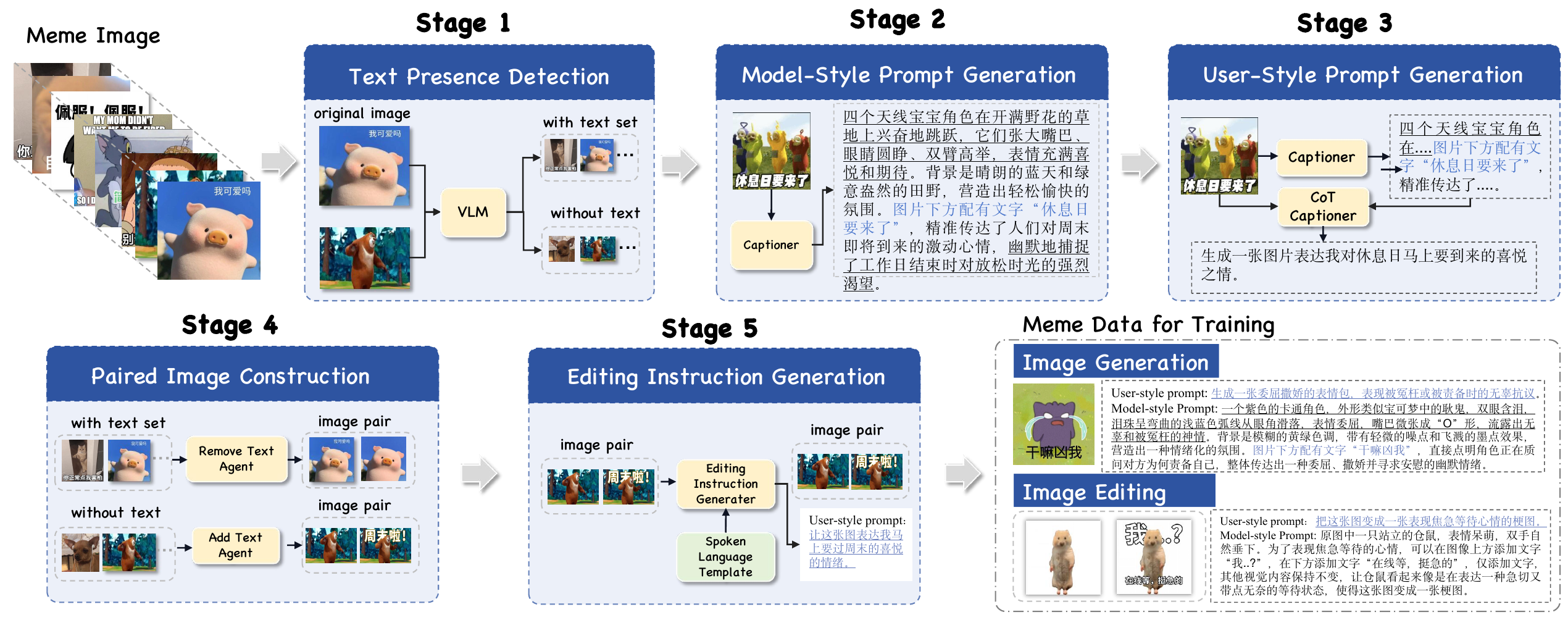} 
    \caption{\textbf{Meme data synthesis pipeline.} It comprises five stages, leveraging chain-of-thought reasoning to process internet memes and synthesize high-quality training data for meme generation and editing.
}
    \label{fig:meme-data-pipeline}
\end{figure}

To build these synthetic datasets, we first crawl and collect a large number of meme images from the Internet and open-source datasets, and then apply an automated pipeline (as illustrated in \cref{fig:meme-data-pipeline}) to generate high-quality paired images and aligned instructions:
\begin{itemize}
  \item \textbf{Stage 1 (Text Presence Detection):} Perform per-sample text detection to determine whether the image contains visible text using VLM.
  \item \textbf{Stage 2 (Model-Augmented Instruction Generation):} Use image captioner VLM to generate model-augmented instructions based on the input image and its original caption, providing fine-grained descriptions of visual details and semantic focus.
  \item \textbf{Stage 3 (User-Style Prompt Generation):} Use CoT captioner to generate shorter, user-style prompts that are closer to real user queries.
  \item \textbf{Stage 4 (Paired Image Construction):} Branch by text presence: if the image contains text, invoke ``Text Removing Agent'' 
  to erase on-image text and obtain a text-free version; otherwise, use ``Text Adding
Agent'' to add the instructed content or perform instruction-driven modifications on the original image, thereby constructing image pairs.
  \item \textbf{Stage 5 (Editing Instruction Generation):} Use editing
instruction generator to generate editing instructions that are strictly aligned with the paired images produced in Stage~3, yielding training triplets of (source image, target image, editing instruction).
\end{itemize}

\subsection{Reasoning-centric Data Synthesis}
\label{sec:data-reasoning}

In real-world applications, user-provided generation and editing instructions are often brief, high-level, and abstract, especially in complex or domain-specific scenarios. While such concise prompts are natural and convenient for users, they frequently omit critical details, such as explicit attribute specifications, spatial relationships, executable editing steps, and domain-specific constraints, making it difficult for models to accurately interpret intent and consistently produce reliable results.

To address this challenge, we propose \textbf{\emph{reasoning-centric}} data synthesis, which introduces an explicit reasoning module as an \emph{interpreter} between raw user instructions and the final multimodal supervision signals. Given a short and abstract instruction, this module automatically derives a more structured, concrete, and actionable specification, including refined objectives, decomposed sub-tasks, verifiable constraints, and ordered editing operations. Such structured interpretations provide clearer learning signals during training. The synthesized data jointly organizes abstract instructions, their explicated reasoning traces, and corresponding execution targets, enabling models to better follow underspecified instructions, improve robustness in challenging domains, and enhance controllability in both generation and editing tasks. In particular, we focus on four core application settings: 
\textbf{\textit{(1) General Images}}, \textbf{\textit{(2) Knowledge-infused Images}}, \textbf{\textit{(3) Meme Images}}, and \textbf{\textit{(4) Science Images}}.

\begin{figure}[h]
    \centering
    \includegraphics[width=0.95\linewidth]{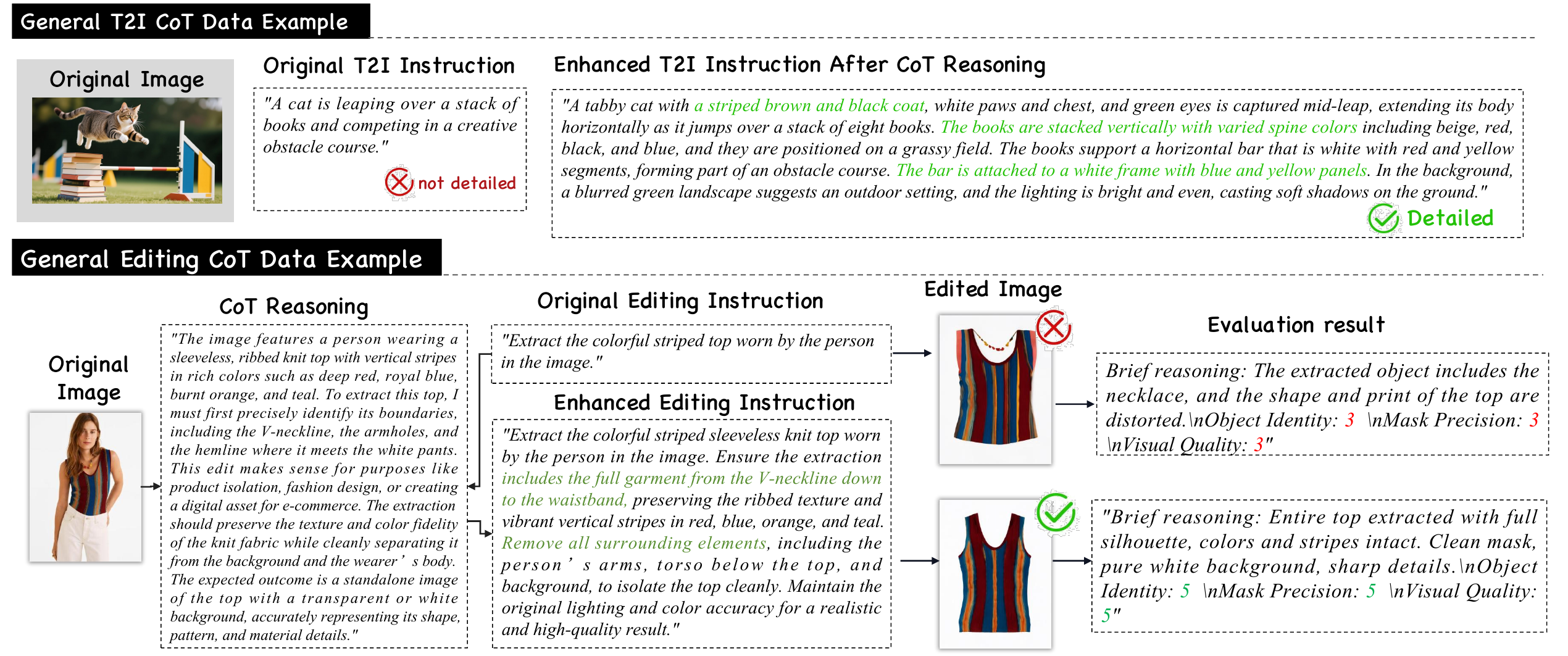} 
    \caption{\textbf{Examples of CoT reasoning and enhancement for general image generation and editing}. The original prompt is enriched through our chain-of-thought reasoning, adding finer-grained details that enable the model to perform generation and editing with greater accuracy and improved fidelity.}
    \label{fig:general-cot}
\end{figure}

\noindent\textbf{General Images.}
In general image generation and editing, 
user-provided instructions are often short and underspecified. Ambiguous descriptions of scene composition, target regions, or modification attributes can cause models to misinterpret user intent, leading to low-fidelity generation or inaccurate fine-grained edits. To mitigate this issue, reasoning-based prompt rewriting and refinement are crucial. Accordingly, we apply CoT augmentation to all instructions in our constructed dataset. For generation tasks, we expand abstract concepts into detailed visual descriptions of objects, backgrounds, and styles; for editing tasks, we enrich instructions with clearer references for localizing target regions, explicit attributes to modify or preserve, and necessary visual-consistency constraints. This process produces more grounded and informative prompts without altering the original intent, thereby improving the learnability of supervision signals and the controllability of outcomes. As shown in Figure~\ref{fig:general-cot}, using CoT-enhanced instructions at inference time yields noticeably more accurate generation and editing results that are better aligned with user requirements.

\begin{figure}[h]
    \centering
    \includegraphics[width=0.9\linewidth]{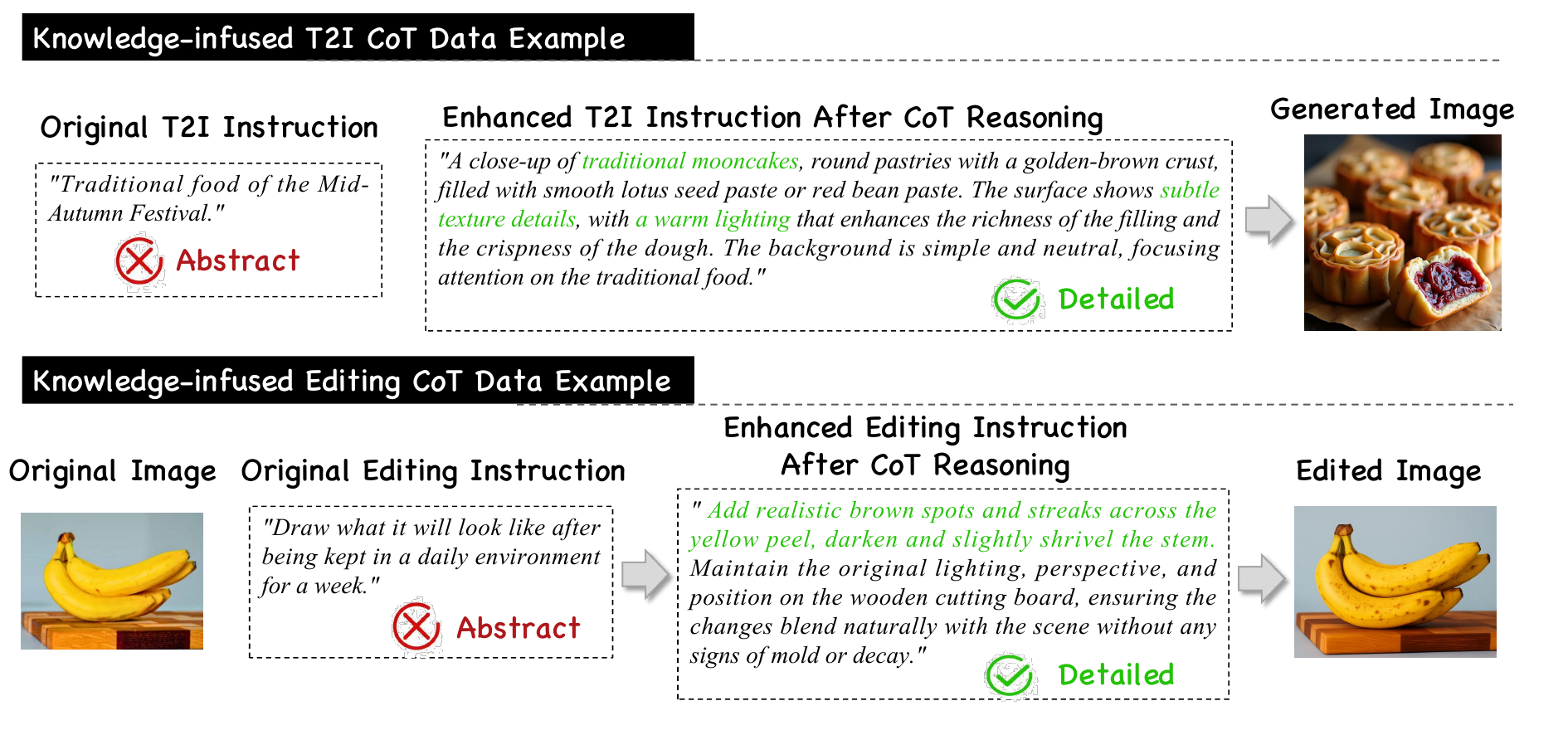} 
    \vspace{-1em}
    \caption{\textbf{Examples of CoT reasoning and enhancement for knowledge-infused T2I and editing.}
    After CoT reasoning, abstract knowledge concepts are concretized, enabling more precise generation and editing.
    }
\label{fig:Knowledge-cot}
\end{figure}

\noindent\textbf{Knowledge-infused Images.}
Knowledge-infused image generation and editing require models to not only accurately comprehend user instructions, which are often concise and abstract, but also to possess deep capabilities for knowledge-based reasoning and analysis. The core of this task lies in deconstructing the implicit background knowledge behind the instructions and translating it into concrete visual concepts or detailed scene descriptions, thereby effectively bridging the informational gap in user inputs. Subsequently, this enriched and explicit semantic information is fed into the generative model, guiding it to produce high-quality images that align with the user's underlying intent while maintaining logical consistency. As illustrated in Figure~\ref{fig:Knowledge-cot}, we employ open-source models to perform detailed-level reasoning enhancement on the instructions within our curated knowledge-based image generation and editing dataset. For instance, the model can automatically establish the cultural association between the ``Mid-Autumn Festival'' and its traditional food, ``mooncakes,'' or link the state of a ``banana after one week'' with the visual feature of ``brown spots'', thus achieving a concrete representation of abstract concepts.

\noindent\textbf{Meme Images.}
In Section~\ref{meme}, we have introduced the tasks of meme generation and editing. In meme scenarios, user instructions are often brief and highly abstract (\eg, ``generate an image that expresses my happiness''), and typically lack descriptions of concrete visual details. 
To address this, we introduce reasoning-based prompt enhancement, converting short instructions into explicit and controllable specifications. Without altering the original intent, we enrich the prompts with (1) concrete scene and visual element details, (2) the intended emotional stance and expression mechanism (\ie, the humor structure), and (3) template and typography constraints (\eg, caption placement postion). For editing tasks, we further specify localized operations and clearly state which regions and attributes should remain unchanged. The resulting pipeline is illustrated in Figure~\ref{fig:meme-data-pipeline}. This reasoning-centric instruction formulation provides stronger and more stable supervision signals, improving both controllability and alignment with user intent for meme generation and editing.

\noindent\textbf{Science Images.}
Science is a typical domain that requires detailed reasoning to generate a semantically correct image.
For science T2I tasks, especially those involving structured symbols such as mathematical formulas and chemical molecular notations, we use LLMs first parses the scientific concepts to produce intermediate reasoning steps, including conceptual analysis and layout planning, and then generates the final image description. For science image editing, the model interprets the source image and instructions, converting abstract editing intents into executable text directives that follow correct scientific logic
For physics editing data, we use the prompt before the summarization stage as the reasoning-informed prompt, which includes explicit and detailed step-by-step instructions. 
For computer science editing data, 
we first define the data structures for problem-solving based on the input and output images, and employ traditional algorithms to solve the problem. Then we record each step of the solution process in detail and map them into 1–3 predefined templates prepared by human. Qwen3-Max~\cite{Qwen3-VL} is utilized to rewrite these templates to increase the diversity of CoT data. 
\begin{figure}[H]
    \centering
    \includegraphics[width=0.95\linewidth]{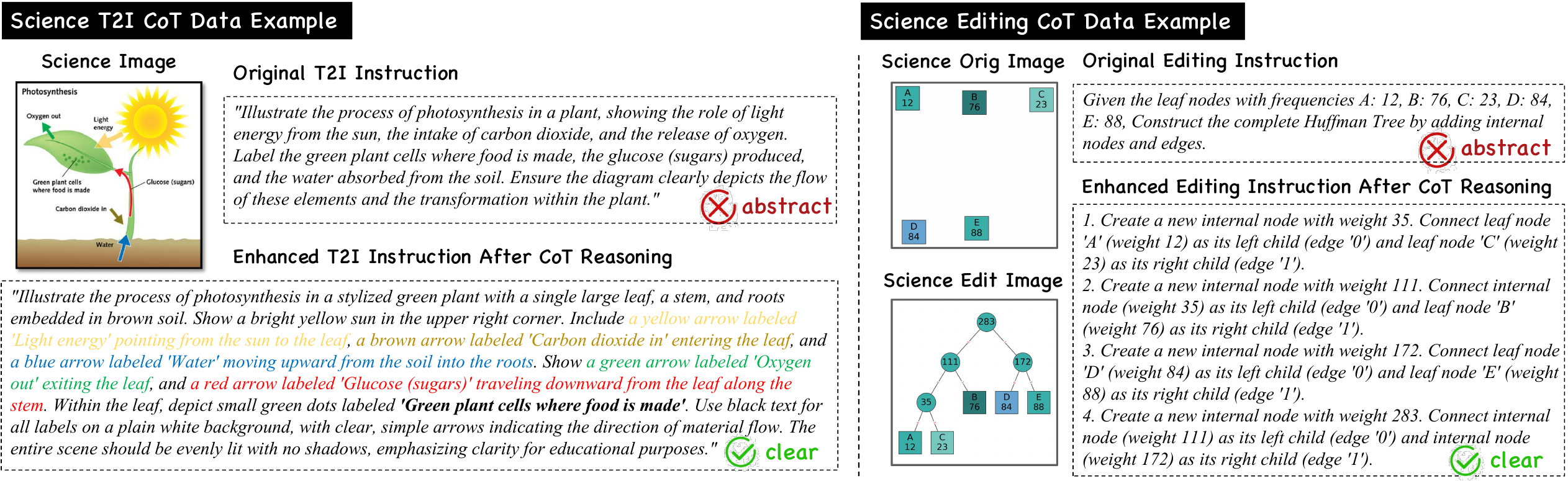} 
    \caption{\textbf{Examples of CoT reasoning and enhancement for Science T2I and Editing.} Through CoT reasoning, scientific knowledge is injected into the generation process in a more explicit and detailed manner (\eg, how elements should be depicted and how operations should be executed).
    }
    \label{fig:editing-cot}
\end{figure}

%% file: tables/data-opensource.tex
\begin{table}[!htbp]
    \centering
    \caption{\textbf{Overview of collected open-source datasets.}}
    \label{tab:tasks_datasets}
    \begin{tabular}{l p{0.7\linewidth}} 
        \toprule
        \textbf{Task} & \textbf{Datasets} \\
        \midrule
        Text-To-Image (T2I) & LAION~\cite{schuhmann2022laion}, BLIP-3o~\cite{chen2025blip3o}, ShareGPT-4o-Image~\cite{chen2025sharegpt}, OSP~\cite{lin2024open}, Echo-4o-Image~\cite{ye2025echo4o}, OpenGPT-4o~\cite{chen2025opengpt4oimagecomprehensivedatasetadvanced}, FaceCaption~\cite{dai202415m}, Flux-Reason-6M~\cite{fang2025flux}, HumanCaption~\cite{dai2024humanvlmfoundationhumanscenevisionlanguage}, POSTER-TEXT~\cite{gao2023textpainter}, AutoPoster~\cite{lin2023autoposter}, CTW~\cite{yuan2019ctw} \\
        \midrule
        Image Editing (IT2I) & InstructPix2Pix~\cite{brooks2023instructpix2pix}, AnyEdit~\cite{yu2024anyedit}, PIPE~\cite{wasserman2024paint}, ImgEdit~\cite{ye2025imgedit}, SEED-Data-Edit~\cite{ge2024seed}, OmniEdit~\cite{wei2024omniedit}, UltraEdit~\cite{zhao2024ultraeditinstructionbasedfinegrainedimage}, HQEdit~\cite{hui2024hq}, ShareGPT-4o-Image~\cite{chen2025sharegpt}, OpenGPT-4o~\cite{chen2025opengpt4oimagecomprehensivedatasetadvanced}, X2Edit~\cite{ma2025x2editrevisitingarbitraryinstructionimage}, X2I2~\cite{wu2025omnigen2}, UniWorld perception~\cite{lin2025uniworld}, NHR-Edit~\cite{Layer2025NoHumansRequired}, GPT-hqedit~\cite{wang2025gptimageedit15mmillionscalegptgeneratedimage}, GPT-omniedit~\cite{wang2025gptimageedit15mmillionscalegptgeneratedimage}, GPT-ultraedit~\cite{wang2025gptimageedit15mmillionscalegptgeneratedimage}, Nano-consistent-150k~\cite{ye2025echo4o}, Pico Banana~\cite{qian2025picobanana400klargescaledatasettextguided} \\
        \bottomrule
    \end{tabular}
\end{table}

%% file: sections/5.experiment.tex
\section{Experiments}
\label{sec:experiments}

\subsection{Experimental Setups}

\modelname is implemented upon InternVL3.5-2B~\cite{wang2025internvl3_5}, using its weight to initialize the visual understanding encoder and the multimodal context backbone. The text tokenizer and conversation format are also the same. For image generation, we use the same VAE as Qwen-Image~\cite{wu2025qwenimagetechnicalreport}.
The visual generation head is randomly initialized, containing 1.7B parameters. The total number of parameters of \modelname is 4B. Detailed configuration is shown in Table~\ref{tab:model_config}.
Following previous works~\cite{brooks2023instructpix2pix}, we also adopt the classifier-free guidance (CFG) for both image and text conditions. During training, for text-to-image generation data, the condition is dropped with a 10\% probability; for image editing data, the multimodal condition (including text and image) is dropped with a 5\% probability, and there is also a 5\% probability of dropping only the text while retaining the image input.
During inference, Flow-DPM-Solver~\cite{xie2024sana} is adopted with 20 inference steps. The CFG scales for dropping the entire condition and text condition only (for editing tasks) are set to 3.5 and 1.5, respectively.
\cref{sys_prompt} illustrates the system and user prompts used under different settings, of which the embeddings are truncated when fed into the visual generation head.
The detailed training settings of each stage are shown in \cref{tab:train_config}. We use VLMEvalkit~\cite{duan2024vlmevalkit} to evaluate the multimodal understanding and reasoning benchmarks. For the evaluation on image generation and editing tasks, we use our self-developed evaluation toolkit named GenEditEvalKit$^\dagger$, which is also open-sourced.

\blfootnote{$\dagger$ \url{https://github.com/open-compass/GenEditEvalKit}}

\input{tables/config-model}

\input{tables/config-train}

\begin{figure}[t]
\begin{AcademicBox}[\footnotesize System Prompt for T2I task]
<|im\_start|>system \\ 
你是书生·万象，英文名是InternVL，是由上海人工智能实验室、清华大学及多家合作单位联合开发的多模态大语言模型。请通过详细描述图像中物体和背景的颜色、形状、大小、纹理、数量、文字内容以及空间位置关系等来对图像进行全面描述: <|im\_end|>
\end{AcademicBox}
\begin{AcademicBox}[\footnotesize System Prompt for IT2I task]
<|im\_start|>system \\ 
你是书生·万象，英文名是InternVL，是由上海人工智能实验室、清华大学及多家合作单位联合开发的多模态大语言模型。请描述输入图像的关键特征（颜色、形状、大小、纹理、物体、背景等），然后解释用户的文本指令应该如何改变或修改图像，从而生成一个满足用户要求的新图像，并适当保持与原始输入的一致性: <|im\_end|>
\end{AcademicBox}
\begin{AcademicBox}[\footnotesize User Prompt for dropping image and text conditions]
<|im\_start|>user \\
Here is a random image <img\_uncond>:<|im\_end|>
\end{AcademicBox}
\begin{AcademicBox}[\footnotesize User Prompt for dropping text conditions only]
<|im\_start|>user \\
Generate an image based on reference images.<|im\_end|>
\end{AcademicBox}
\vspace{-1em}
\caption{\textbf{System prompts and user prompts adopted during training for different tasks}, 
where <img\_uncond> is a learnable special token for image generation.\looseness=-1}
\label{sys_prompt}
\end{figure}

\subsection{Multimodal Understanding and Reasoning}

To assess multimodal understanding and reasoning capabilities, we evaluate \modelname on 7 widely recognized MLLM benchmarks, including MME-P~\cite{fu2025mme}, SEED~\cite{li2023seed}, ChartQA~\cite{masry2022chartqa}, OCRBench~\cite{liu2023ocrbench}, MMMU~\cite{MMMU}, MathVerse~\cite{zhang2024mathverse}, and LogicVista~\cite{xiao2024logicvista}. 
As shown in \cref{tab:exp_und_reason}, \modelname demonstrates robust performance on multimodal understanding and reasoning benchmarks, significantly surpassing comparable-sized UMMs such as Janus-Pro~\cite{chen2025janus} and Ovis-U1~\cite{wang2025ovis} across key metrics like MME-P~\cite{fu2025mme} (1607.5) and OCRBench ~\cite{liu2023ocrbench}(83.9). Remarkably, despite its compact architecture (2B+1.7B), it delivers reasoning capabilities comparable to the significantly larger BAGEL~\cite{deng2025bagel}(7B+7B) model, particularly on MMMU~\cite{MMMU}(54.7 vs. 55.3). These results indicate that our unified training strategy effectively retains the strong visual–language comprehension of understanding-only baselines, showing minimal degradation while achieving a superior balance between understanding and generation.

\input{tables/final/understand_reason_v2}

\subsection{Text-to-Image Generation}

To comprehensively evaluate the text-to-image generation capabilities, we adopt GenEval~\cite{ghosh2023geneval}, DPG-Bench~\cite{hu2024ella}, TIIF~\cite{wei2025tiif}, and OneIG~\cite{chang2025oneig} for general assessment, LongText~\cite{geng2025x} and CVTG-2k~\cite{du2025textcrafter} for text rendering quality, and WISE~\cite{niu2025wise} and GenExam~\cite{GenExam} for knowledge-intensive generation. 

\subsubsection{General Image Generation}

\noindent\textbf{GenEval}. GenEval~\cite{ghosh2023geneval} is an object-focused framework for evaluating compositional image properties such as object co-occurrence, position, count, and color. In \cref{tab:exp_geneval}, \modelname achieves highest overall score (0.85) among existing unified models such as BAGEL~\cite{deng2025bagel} with only half or even less number of parameters. It also surpasses most of the specialized generation models. 
\input{tables/final/geneval}

\noindent\textbf{DPG-Bench}.
\cref{tab:exp_dpgbench} shows the results on DPG-Bench~\cite{hu2024ella}, which provides dense prompts that describe multiple objects to assess the intricate semantic alignment capabilities of text-to-image models. In \cref{tab:exp_dpgbench}, our model exhibits stronger performance than other unified models, especially on \textit{Global} and \textit{Entity} dimensions.

\input{tables/final/dpgbench}

\newpage

\noindent\textbf{TIIF}. TIIF~\cite{wei2025tiif} aims to systematically evaluate the ability to follow intricate instructions. As shown in \cref{tab:exp_tiif_short,tab:exp_tiif_long}, our \modelname achieves strong performance among unified models, especially on advanced instruction following. There are still notable differences between unified models and generation models, indicating potential future improvements in instruction following.
\input{tables/final/tiif-short}
\input{tables/final/tiif-long}
\newpage
\noindent\textbf{OneIG-Bench}. In \cref{tab:exp_oneigbench,tab:exp_oneigbench_zh}, we assess our \modelname on OneIG-Bench~\cite{chang2025oneig}, which is designed for fine-grained evaluation across subject-element alignment, text rendering precision, reasoning-generated content, stylization, and diversity. \modelname shows highest overall score among open-source unified models with a small parameter scale, demonstrating its fine-grained alignment capability with multi-lingual robustness. 

\input{tables/final/oneig}
\input{tables/final/oneig-zh}

\begin{figure}[H]
    \centering
    \includegraphics[width=\linewidth]{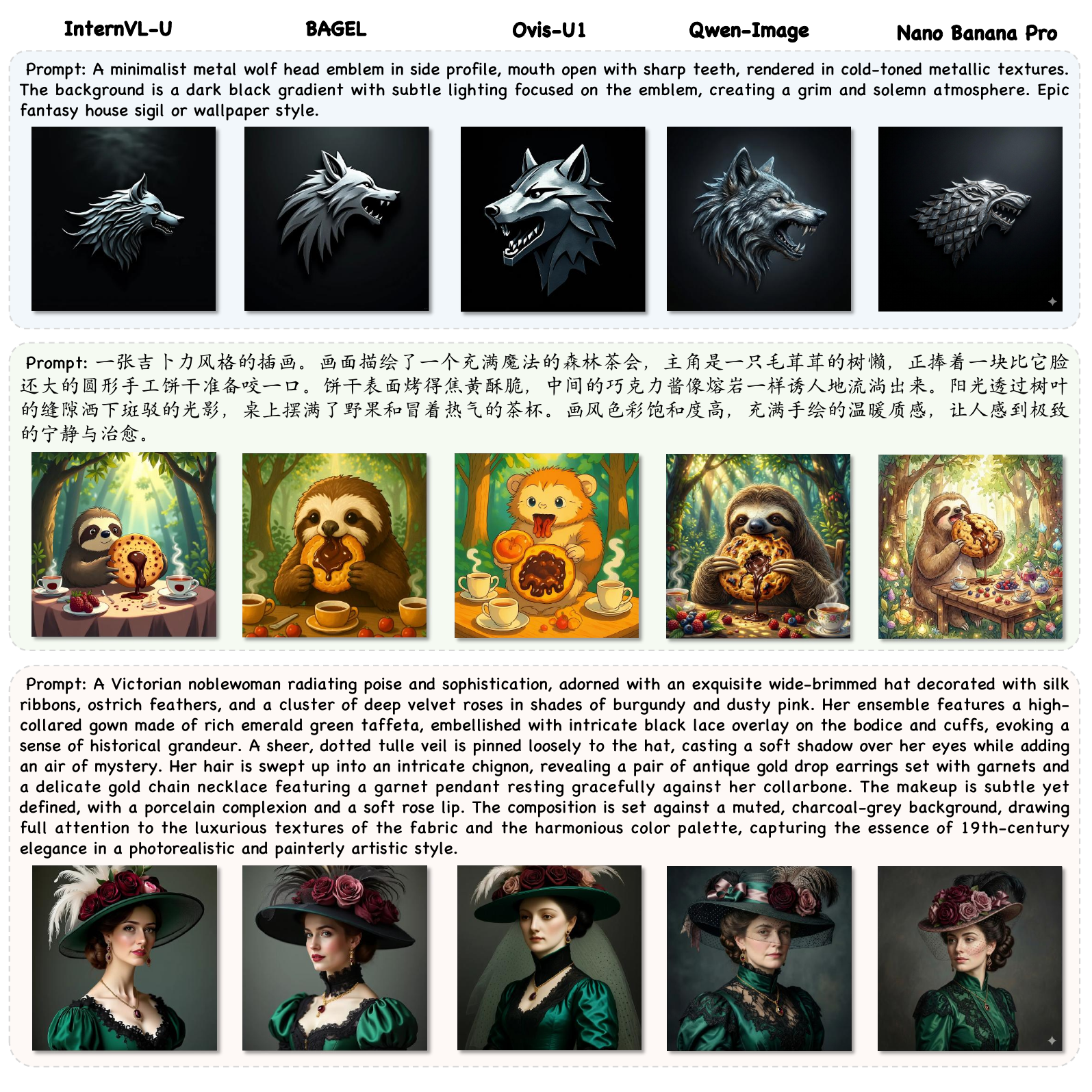} 
    \caption{\textbf{Visualizations of general image generation.} Compared to other open-source models, \modelname demonstrates exceptional fidelity in rendering intricate textures and nuanced lighting effects, capturing the precise intent of each instruction.}
    \label{fig:general-t2i-vis}
\end{figure}

\newpage
\noindent\textbf{Qualitative Results}. 
To further illustrate the practical strengths  beyond quantitative metrics, we provide additional qualitative comparisons. 
As shown in Figure~\ref{fig:general-t2i-vis}, InternVL-U demonstrates exceptional visual fidelity in general image generation, particularly in rendering intricate textures and nuanced lighting effects, while precisely capturing the intent of each instruction.

\newpage

\subsubsection{Text-centric Image Generation}

\noindent\textbf{CVTG-2k}. In \cref{tab:exp_cvtg}, we provide results on CVTG-2k~\cite{du2025textcrafter}, which is specialized designed for complex visual text generation. \modelname achieves state-of-the-art performance among unified models, with an average word accuracy of 0.623. 
\input{tables/final/cvtg}

\noindent\textbf{LongText-Bench}.  LongText-Bench~\cite{geng2025x} assesses the ability to render longer texts on images. As shown in \cref{tab:exp_longtext}, \modelname demonstrates robust multilingual text generation with scores of 0.738 in English and 0.860 in Chinese, surpassing previous unified models by large margins. These results show that our model effectively addresses the previous deficiency of unified models in rendering legible text.

\input{tables/final/longtext}

\begin{figure}[H]
    \centering
    \includegraphics[width=0.99\linewidth]{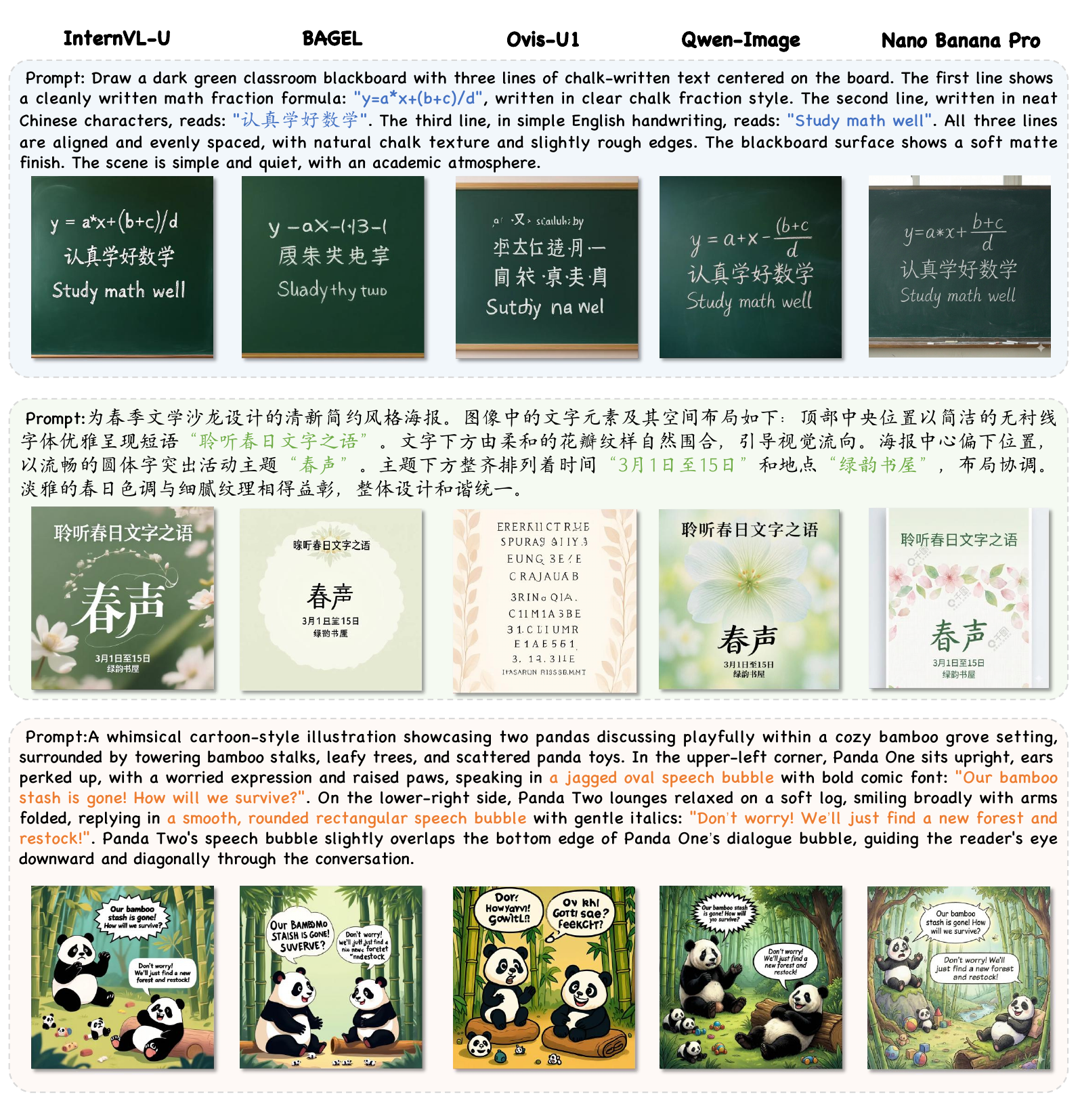} 
    \caption{\textbf{Visualizations of text-centric image generation.} The results show that our InternVL-U has excellent ability in rendering symbols of Chinese, English, and numerical formulas. Compared with the open-source unified multimodal models BAGEL and Ovis-U1, it has better rendering ability and comparable performance with Qwen Image, a large parameter and commercial closed source model Nano Banana Pro.}
    \label{fig:text-generation-vis}
\end{figure}

\noindent\textbf{Qualitative Results}. 
As shown in Figure~\ref{fig:text-generation-vis}, InternVL-U shows excellent capability in rendering Chinese and English characters as well as numerical and mathematical symbols with higher readability and fewer artifacts. Compared with open-source unified multimodal baselines such as BAGEL~\cite{deng2025bagel} and Ovis-U1~\cite{wang2025ovis}, it achieves better text rendering quality, and remains competitive with the 20B large-scale model Qwen-Image~\cite{wu2025qwenimagetechnicalreport} and the closed-source model Nano-Banana-Pro~\cite{deepmind_gemini3proimage_2025}.

\input{tables/final/wise}
\input{tables/final/genexam}
\subsubsection{Knowledge-informed Image Generation}
\noindent\textbf{WISE}. WISE~\cite{niu2025wise} evaluates whether models can integrate world knowledge into text-to-image generation. As shown in \cref{tab:exp_wise}, \modelname with CoT yields significant performance gains (from 0.46 to 0.58 overall scores) and surpasses other unified baselines like BAGEL~\cite{deng2025bagel} and UniWorld-V1~\cite{lin2025uniworld}, suggesting high capabilities in cultural commonsense, spatio-temporal reasoning and natural science. 

\noindent\textbf{GenExam}. GenExam~\cite{GenExam} assesses the ability of text-to-image models to understand reasoning with disciplinary knowledge through exam-style instructions. In \cref{tab:exp_genexam}, our model achieves the highest scores among unified models, especially on physics, chemistry, and biology. Boosted with CoT, \modelname attains an overall score of 22.9 with only 3.7B parameters. This validates \modelname's on science-centric image generation and its integrated ability of understanding, reasoning and generating.

\begin{figure}[t]
    \centering
    \includegraphics[width=0.95\linewidth]{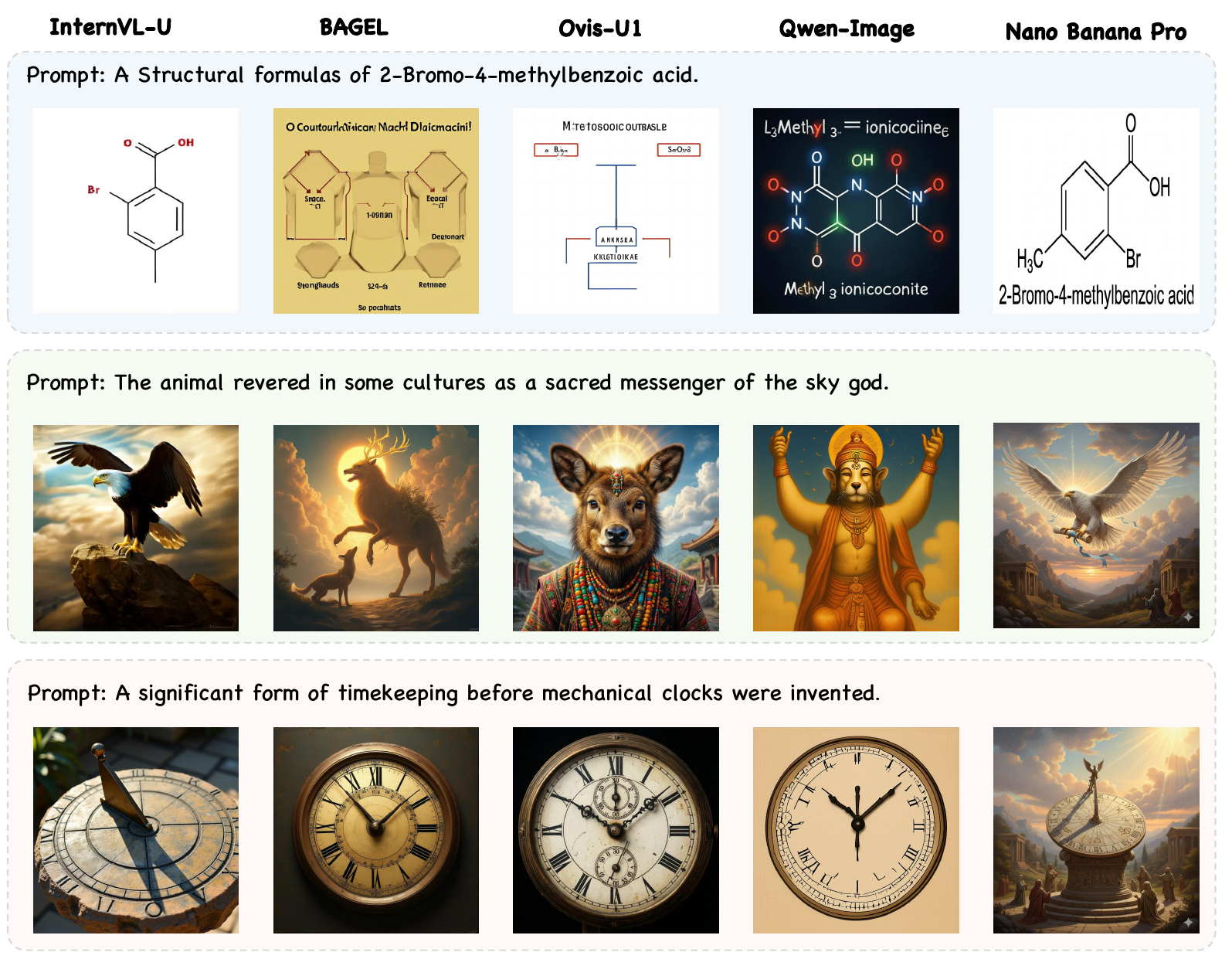} 
    \caption{\textbf{Visualizations of knowledge-informed image generation.} InternVL-U exhibits superior capability in accurate knowledge rendering. By effectively integrating domain knowledge, our model produces visually faithful results for complex prompts, significantly outperforming baselines that lack specific world knowledge.}
    \label{fig:Knowledge-informed-t2i-vis}
\end{figure}

\noindent\textbf{Qualitative Results}. 
As shown in Figure~\ref{fig:Knowledge-informed-t2i-vis}, for prompts that require the model to understand world knowledge, InternVL-U delivers stronger knowledge-grounded rendering, producing visually faithful results for complex instructions and markedly outperforming baselines without explicit knowledge integration.

\subsection{Image Editing}

For image editing, we employ existing benchmarks of ImgEdit~\cite{ye2025imgedit}, GEdit-Bench~\cite{liu2025step1x-edit}, and RISEBench~\cite{zhao2025envisioning}. Furthermore, given the extensive application scenarios for text editing, we have constructed an additional text-centric image editing benchmark, namely TextEdit, to evaluate models' accuracy in performing text editing across virtual and real-world scenarios.

\subsubsection{General Image Editing}

\input{tables/final/imgedit}
\input{tables/final/gedit}

\noindent\textbf{ImgEdit}.
ImgEdit~\cite{ye2025imgedit} covers diverse range of single- and multi-turn editing tasks. As shown in \cref{tab:exp_imgedit}, 
\modelname demonstrates competitive editing proficiency among unified models, where the CoT models achieves 3.82 overall score.

\noindent\textbf{GEdit-Bench}. GEdit-Bench~\cite{liu2025step1x-edit} contains prompts with both real-world editing requirements
and high diversity. In \cref{tab:exp_gedit}, \modelname achieves an average score of 6.66, surpassing baselines like BAGEL~\cite{deng2025bagel} (6.52) and Ovis-U1~\cite{wang2025ovis} (6.42). Notably, applying the CoT strategy further enhances performance, raising the score to 6.88. While specialized editing models like Qwen-Image-Edit~\cite{wu2025qwenimagetechnicalreport} still hold a lead in certain metrics, \modelname's performance confirms the viability of a unified architecture for diverse editing tasks, particularly when augmented with explicit reasoning steps.

\begin{figure}[H]
    \centering
    \includegraphics[width=0.95\linewidth]{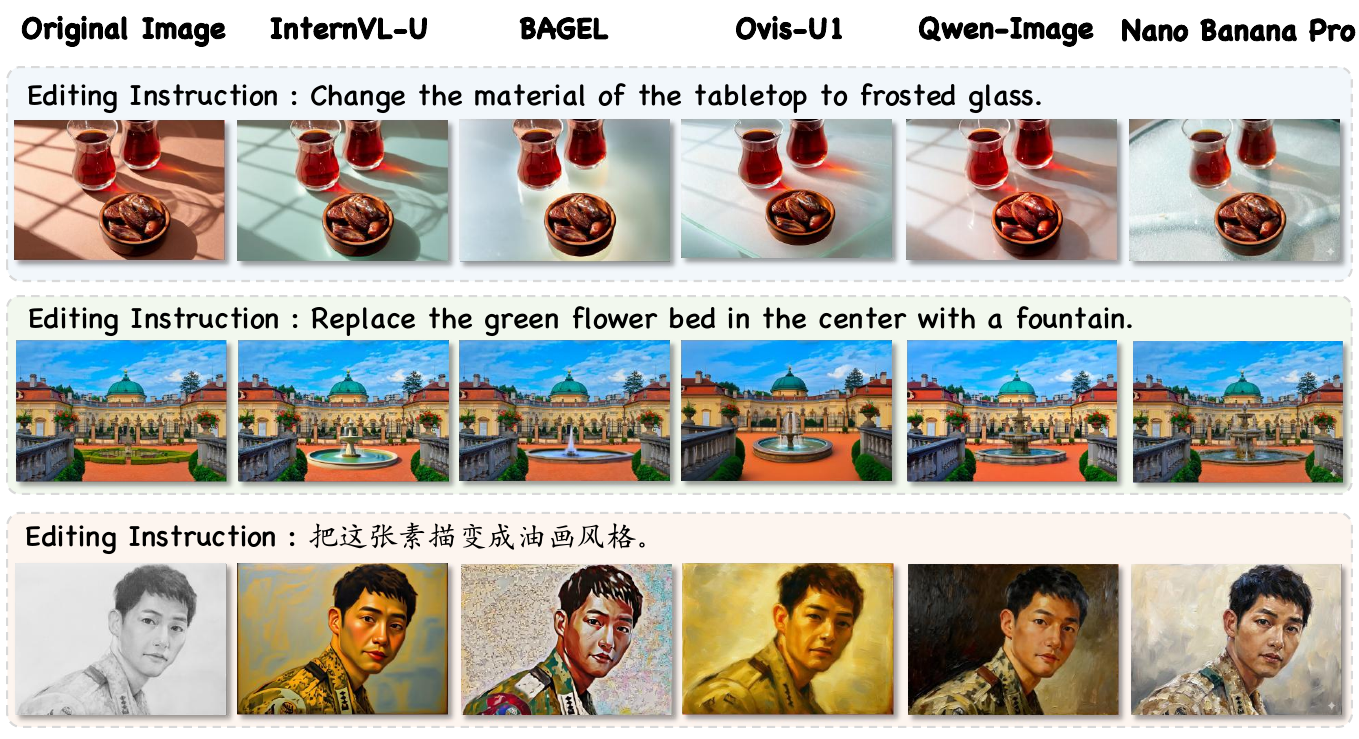} 
    \caption{\textbf{Visualizations of general image editing.} \modelname excels in generating realistic textures and styles while maintaining high fidelity to the original image's lighting and structural details, demonstrating superior performance across various tasks compared to other open-source models.}
    \label{fig:general-edit-vis}
\end{figure}

\noindent\textbf{Qualitative Results}. 
As shown in Figure~\ref{fig:general-edit-vis}, InternVL-U excels at producing realistic textures and styles while preserving the source image's lighting and structural details, resulting in more natural and coherent edits across a wide range of scenarios compared with other open-source models.

\subsubsection{Text-centric Image Editing}

\noindent\textbf{TextEdit}. To rigorously evaluate text editing capabilities, we introduce \textbf{TextEdit}$^\dagger$, a novel benchmark containing 2,148 samples featuring diverse editing scenarios and high-quality edited image ground truth (see Section~\ref{sec:textedit} for more details about the benchmark construction). As shown in \cref{tab:exp_textedit_mllm,tab:exp_textedit_rule}, \modelname demonstrates superior performance on this benchmark, achieving a F1 score of 0.71 in classic metrics, matching the Nano Banano Pro~\cite{deepmind_gemini3proimage_2025} and drastically outperforming unified counterparts like Ovis-U1~\cite{wang2025ovis} (0.35). This advantage is further corroborated by MLLM-based evaluations, where \modelname attains an average score of 0.88 on the images from real scene, significantly surpassing BAGEL~\cite{deng2025bagel} (0.53) and exhibiting competitive capability against closed-source commercial models like GPT-Image-1.5~\cite{GPT-Image-1.5}.

\input{tables/final/textedit-rule}
\input{tables/final/textedit-mllm}

\begin{figure}[h]
    \centering
    \includegraphics[width=0.9\linewidth]{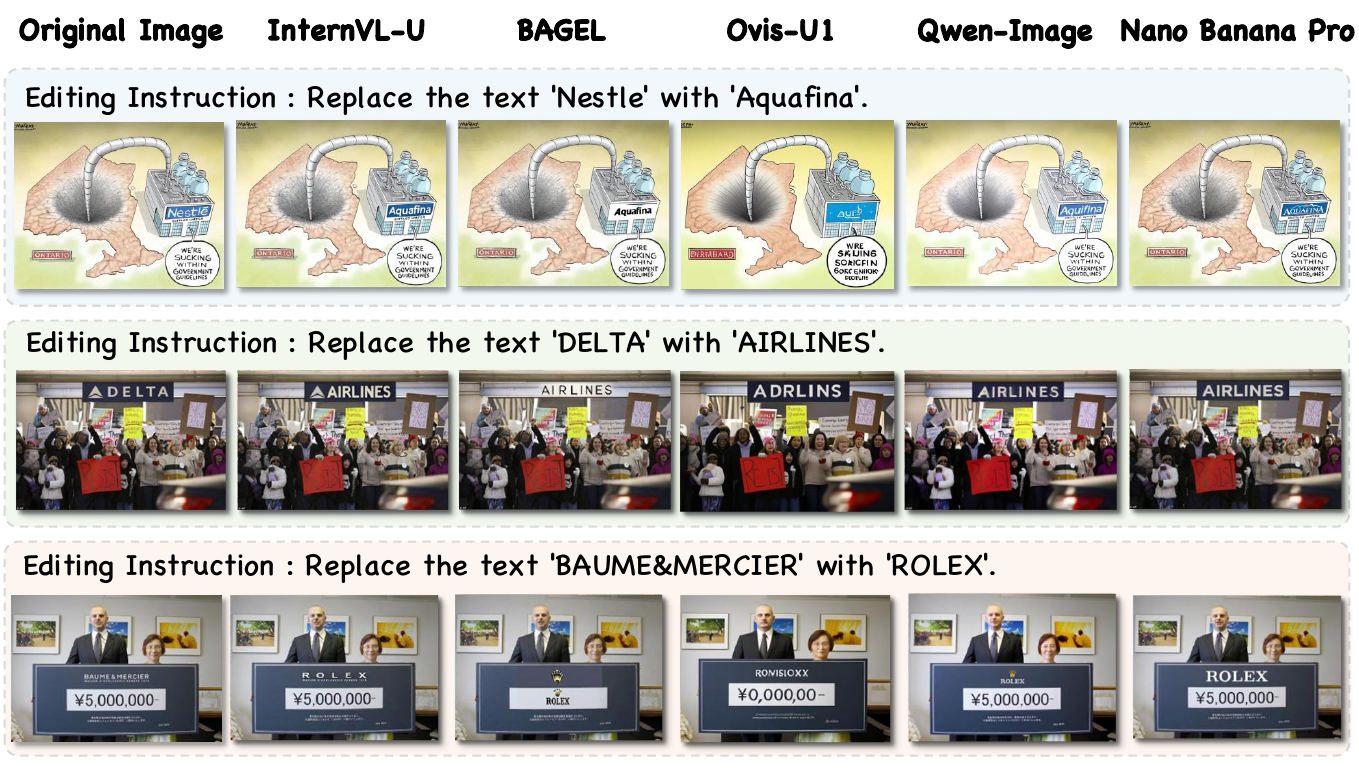} 
    \caption{\textbf{Visualizations of text-centric image editing}. InternVL-U demonstrates more accurate and faithful text editing capabilities, while maintaining strong consistency for both textual and visual content outside the target editing region.
    }
    \label{fig:text-bench-vis}
\end{figure}

\noindent\textbf{Qualitative Results}. 
As shown in \cref{fig:text-bench-vis}, we visualize the performance of representative top-tier open-source and commercial models on our proposed benchmark TextEdit, our  InternVL-U, which achieves strong results across a wide range of text-editing scenarios. In particular, InternVL-U can accurately localize the text to be replaced in the image and substitute it with the target text, while preserving both visual aesthetics and textual correctness.  These results provide a clear picture of the current state of the art and effectively mark the upper bounds of existing text editing capabilities, highlighting how our benchmark delineates the performance frontier of text-centric image editing.
\blfootnote{$\dagger$ \url{https://github.com/open-compass/TextEdit}}

\subsubsection{Reasoning-informed Image Editing}

\noindent\textbf{RISEBench}.
We further assessed the model's ability to handle complex editing instructions that require logical deduction on the RISEBench~\cite{zhao2025envisioning} benchmark. As detailed in \cref{tab:exp_rise}, the introduction of the CoT strategy yields a remarkable performance boost, elevating \modelname's overall score from 3.6 to 9.4. This enhancement allows it to surpass both open-source unified baselines (e.g., BAGEL~\cite{deng2025bagel} at 6.1) and specialized generation models like Qwen-Image-Edit~\cite{wu2025qwenimagetechnicalreport} (8.9). Notably, CoT significantly improves Instruction Reasoning (IR) and Appearance Consistency (AC), demonstrating that explicit reasoning is essential for the accurate execution of complex, logic-dependent editing tasks.

\input{tables/final/rise}

\begin{figure}[h]
    \centering
    \includegraphics[width=0.92\linewidth]{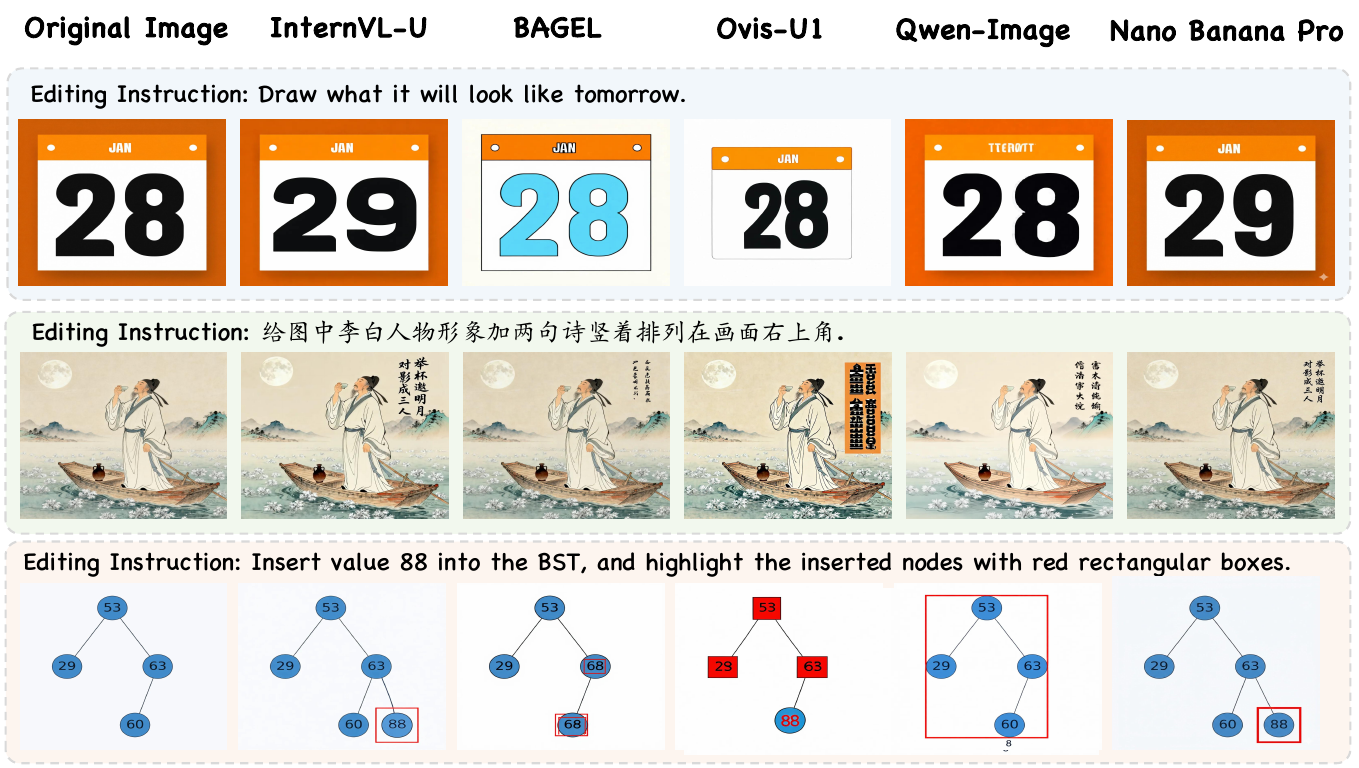} 
    \caption{\textbf{Visualizations of reasoning-informed image editing.} InternVL-U outperforms state-of-the-art models in handling complex prompts requiring multi-step reasoning. The results demonstrate our model's superior ability to accurately interpret and execute diverse logical constraints, spanning temporal calculations for updating calendar dates, spatial and cultural understanding for poem placement, and precise algorithmic rules for binary search tree insertion.}
    \label{fig:reasoning-informed-edit-vis}
\end{figure}

\noindent\textbf{Qualitative Results}. 
As shown in \cref{fig:reasoning-informed-edit-vis}, InternVL-U handles complex editing instructions that require multi-step reasoning and strict logical constraints more reliably than prior methods. It can accurately interpret and execute diverse constraints, including temporal calculations (e.g., updating calendar dates), spatial and cultural understanding (e.g., retrieve contextually appropriate poetry given an image), and precise algorithmic rules (e.g., binary search tree insertion).

\begin{figure}[H]
    \centering
    \includegraphics[width=0.98\linewidth]{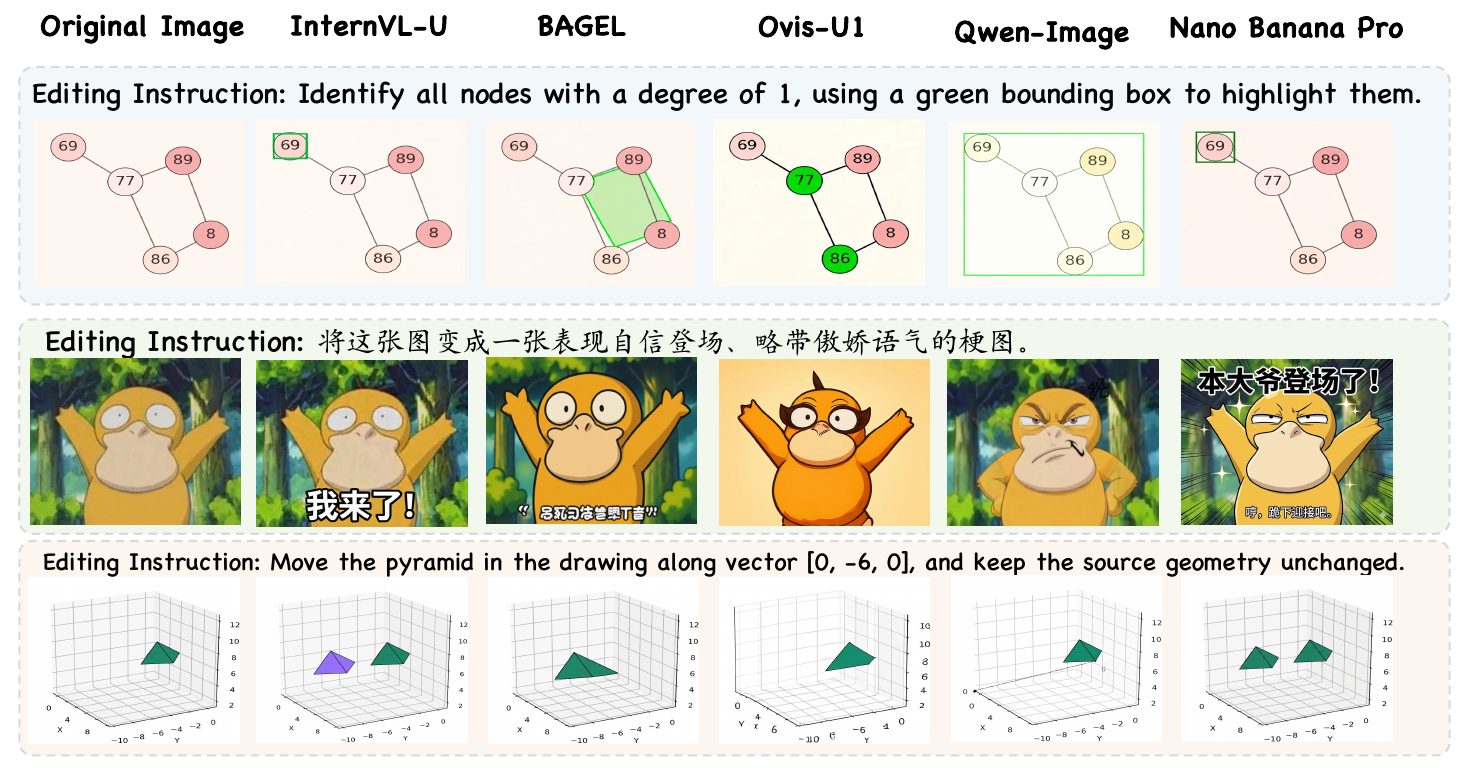} 
    \caption{\textbf{Visualizations of more special image editing examples.}  InternVL-U demonstrates its advanced spatial reasoning and precise control in complex editing tasks. The results highlight the model's superior ability to accurately identify graph properties (e.g., node degrees), generate humor-centric content with appropriate expressions, and execute precise 3D geometric transformations based on coordinate vectors, showing its broad applicability in specialized domains.}
    \label{fig:special-edit-vis}
\end{figure}

\subsection{More Qualitative Results}
As shown in Figure~\ref{fig:special-edit-vis}, we present specialized editing examples that showcase InternVL-U’s distinctive capabilities beyond standard editing, enabling it to handle uncommon and challenging requirements robustly, including computer-science knowledge, humor-centric content creation, and mathematics-related edits, demonstrating robust controllability and broad applicability beyond standard editing settings.

%% file: tables/config-model.tex
\begin{table}[htbp]
\centering
\caption{\textbf{Detailed configuration of \modelname architecture.}}
\label{tab:model_config}
\begin{tabular}{l|c|c|c}
\hline
\textbf{Configuration} & \textbf{Visual Und. Encoder} & \textbf{Context Backbone} & \textbf{Visual Gen. Head} \\
\hline
\# Layers             & 24      & 28     & 20      \\
\# Num Heads (Q / KV) & 16 / 16 & 16 / 8 & 12 / 12 \\
Head Size             & 64      & 128    & 128     \\
Intermediate Size     & 4,096   & 6,144  & 6,144   \\
Patch / Scale Factor  & 14      & -      & 2       \\
\hline
\# Parameters         & 0.3B    & 2B     & 1.7B    \\
\hline
\end{tabular}
\end{table}

%% file: tables/config-train.tex
\begin{table}[htbp]
\centering
\caption{\textbf{Hyperparameters for different training stages}.
}
\label{tab:train_config}
\begin{tabular}{l|ccc}
\hline
\textbf{Hyperparameters} & \textbf{Stage 1} & \textbf{Stage 2} & \textbf{Stage 3} \\
\hline
\multicolumn{4}{l}{\textit{Trainable modules}} \\
\quad Backbone & \textcolor{red}{$\times$} & \textcolor{red}{$\times$} & \textcolor{green}{\checkmark} \\
\quad Visual gen. head & \textcolor{green}{\checkmark} & \textcolor{green}{\checkmark} & \textcolor{green}{\checkmark} \\
\hline
Learning rate & 3e-04 & 1e-04 & 1e-05 \\
LR scheduler & Constant & Cosine & Cosine \\
Weight decay & 0.0 & 0.0 & 0.01 \\
Gradient norm clip & 0.2 & 0.2 & 1 \\
\hline
Optimizer & \multicolumn{3}{c}{AdamW ($\beta_1=0.9, \beta_2=0.999, \epsilon=10^{-8}$)} \\
Warm-up steps & \multicolumn{3}{c}{1000} \\
\hline
Training steps & 250,000 & 60,000 & 20,000 \\
Batch size & 2048 & 1024 & 1024 \\
Gen. resolution (min, max) & (512, 512) & (512, 1024) & (512, 1024) \\
Und. resolution (min, max) & \multicolumn{3}{c}{(448, 448)} \\
Diffusion timestep shift & \multicolumn{3}{c}{3.0} \\
\hline
Data / tasks & T2I + IT2I & T2I + IT2I & T2I + IT2I + Und \\
Data ratio & 4:1 & 3:4 & 1:1:2 \\
Loss weight (NTP:VP) & 0:1 & 0:1 & 1:20 \\
\hline
\end{tabular}
\end{table}

%% file: tables/final/understand_reason_v2.tex
\begin{table}[htbp]
\centering
\caption{\textbf{Comparisons between \modelname and baseline models on multimodal understanding and reasoning benchmarks.} * indicates the results based on our evaluation scripts. Sizes of unified models in ``A + B'' indicate separate understanding (A) and generation (B) parameters.
}
\resizebox{\textwidth}{!}{
\begin{tabular}{l|c|cccc|ccc}
\hline
\multirow{2}{*}{\textbf{Model}} & \multirow{2}{*}{\textbf{\#Params}} & \multicolumn{4}{c|}{\textbf{Understanding}} & \multicolumn{3}{c}{\textbf{Reasoning}} \\
\cline{3-6} \cline{7-9}
& & \textbf{MME-P~\cite{fu2025mme}} & \textbf{SEED~\cite{li2023seed}} & \textbf{ChartQA~\cite{masry2022chartqa}} & \textbf{OCRBench~\cite{liu2023ocrbench}} & \textbf{MMMU~\cite{MMMU}} & \textbf{MathVerse~\cite{zhang2024mathverse}} & \textbf{LogicVista~\cite{xiao2024logicvista}} \\
\hline
\rowcolor{gray!15}
\multicolumn{9}{l}{\textit{MLLMs w/o Generator}} \\
LLaVA-1.5V~\cite{liu2023improvedllava} & 7B & 1510.7 & 65.8 & 17.8 & 31.8 & 35.7 & 7.6 & -- \\
Qwen2.5-VL~\cite{Qwen2.5-VL} & 3B & 1574.9 & 73.7 & 84.0 & 79.7 & 53.1 & 31.2 & 40.3 \\
InternVL3.5~\cite{wang2025internvl3_5} & 2B & 1552.1 & 75.3 & 80.7 & 83.6 & 59.0 & 53.4 & 47.7 \\
\hline
\rowcolor{gray!15}
\multicolumn{9}{l}{\textit{UMMs w/ Generator}} \\
JanusFlow~\cite{ma2025janusflow} & 1.3B & 1333.1 & 70.5 & 42.4 & 53.2 & 29.3 & -- & -- \\
Janus-Pro~\cite{chen2025janus} & 1.5B & 1444.0 & 68.3 & 23.4 & 48.7 & 36.3 & -- & -- \\
Show-o2~\cite{xie2025show} & 1.5B & 1450.9 & 65.6 & 40.0 & 24.5 & 37.1 & -- & -- \\
TUNA~\cite{liu2025tuna} & 1.5B & 1461.5 & 69.3 & 82.1 & 71.9 & 39.1 & -- & -- \\
MetaQuery-L~\cite{pan2025transfer} & 3B+1B & 1574.3 & 73.8 & -- & -- & 53.1 & -- & -- \\
Ovis-U1~\cite{wang2025ovis} & 2.4B+1.2B & 1508.0* & 75.5* & 76.4* & 88.3 & 51.1 & 30.6* & 32.4* \\
Emu3~\cite{cui2025emu3} & 8B & -- & 68.2 & -- & 68.7 & 31.6 & -- & -- \\
BAGEL~\cite{deng2025bagel} & 7B+7B & 1687.0 & 78.5 & 78.5 & 73.3 & 55.3 & 48.1* & 44.3* \\
\modelname & 2B+1.7B & 1607.5 & 75.2 & 76.6 & 83.9 & 54.7 & 45.6 & 40.3 \\
\hline
\end{tabular}
}
\label{tab:exp_und_reason}
\end{table}

%% file: tables/final/geneval.tex
\begin{table}[htbp]
\centering
\caption{\textbf{Evaluation of general text-to-image generation ability on GenEval~\cite{ghosh2023geneval}.} Sizes of unified models in ``A + B'' indicate separate understanding (A) and generation (B) parameters.
}
\resizebox{\textwidth}{!}{
\begin{tabular}{l|c|ccccccc}
\hline
\textbf{Model} & \textbf{\#Params} & \textbf{Single Object} & \textbf{Two Object} & \textbf{Counting} & \textbf{Colors} & \textbf{Position} & \textbf{Color Attribution} & \textbf{Overall} \\
\hline
\rowcolor{gray!15}
\multicolumn{9}{l}{\textit{Generation Models}} \\
FLUX.1 [dev]~\cite{flux2024} & 12B & 0.98 & 0.81 & 0.74 & 0.79 & 0.22 & 0.45 & 0.66 \\
SD3-Medium~\cite{mmdit} & 2B & 0.99 & 0.94 & 0.72 & 0.89 & 0.33 & 0.60 & 0.74 \\
Seedream 3.0~\cite{gao2025seedream} & - & 0.99 & 0.96 & 0.91 & 0.93 & 0.47 & 0.80 & 0.84 \\
GPT Image 1 [High]~\cite{GPT-Image-1} & - & 0.99 & 0.92 & 0.85 & 0.92 & 0.75 & 0.61 & 0.84 \\
Z-Image~\cite{cai2025z} & 6B & 1.00 & 0.94 & 0.78 & 0.93 & 0.62 & 0.77 & 0.84 \\
Qwen-Image & 20B & 0.99 & 0.92 & 0.89 & 0.88 & 0.76 & 0.77 & 0.87 \\
\hline
\rowcolor{gray!15}
\multicolumn{9}{l}{\textit{Unified Models}} \\
Show-o2~\cite{xie2025show} & 7B & 1.00 & 0.87 & 0.58 & 0.92 & 0.52 & 0.62 & 0.76 \\
Janus-Pro~\cite{chen2025janus} & 7B & 0.99 & 0.89 & 0.59 & 0.90 & 0.79 & 0.66 & 0.80 \\
UniWorld-V1~\cite{lin2025uniworld} & 7B+13B & 0.99 & 0.93 & 0.79 & 0.89 & 0.49 & 0.70 & 0.80 \\
OmniGen2~\cite{wu2025omnigen2} & 3B+4B & 1.00 & 0.95 & 0.64 & 0.88 & 0.55 & 0.76 & 0.80 \\
BAGEL~\cite{deng2025bagel} & 7B+7B & 0.99 & 0.94 & 0.81 & 0.88 & 0.64 & 0.63 & 0.82 \\
\modelname & 2B+1.7B & 0.99 & 0.94 & 0.74 & 0.91 & 0.77 & 0.74 & 0.85 \\
\hline
\end{tabular}
}

\label{tab:exp_geneval}
\end{table}

%% file: tables/final/dpgbench.tex
\begin{table}[htbp]
\centering
\caption{\textbf{Evaluation of general text-to-image generation ability on DPG-Bench~\cite{hu2024ella}.} Sizes of unified models in ``A + B'' indicate separate understanding (A) and generation (B) parameters.
}
\resizebox{0.85\textwidth}{!}{
\begin{tabular}{l|c|cccccc}
\hline
\textbf{Model} & \textbf{\#Params} & \textbf{Global} & \textbf{Entity} & \textbf{Attribute} & \textbf{Relation} & \textbf{Other} & \textbf{Overall} \\
\hline
\rowcolor{gray!15}
\multicolumn{8}{l}{\textit{Generation Models}} \\
FLUX.1 [dev]~\cite{flux2024} & 12B & 82.10 & 89.50 & 88.80 & 91.10 & 89.40 & 84.00 \\
SD3-Medium~\cite{mmdit} & 2B & 87.90 & 91.01 & 88.83 & 80.70 & 88.68 & 84.08 \\
GPT Image 1 [High]~\cite{GPT-Image-1} & - & 88.89 & 88.94 & 89.84 & 92.63 & 90.96 & 85.15 \\
Nano Banana Pro~\cite{deepmind_gemini3proimage_2025} & - & 91.00 & 92.85 & 91.56 & 92.39 & 89.93 & 87.16 \\
Z-Image~\cite{cai2025z} & 6B & 93.39 & 91.22 & 93.16 & 92.22 & 91.52 & 88.14 \\
Qwen-Image~\cite{wu2025qwenimagetechnicalreport} & 20B & 91.32 & 91.56 & 92.02 & 94.31 & 92.73 & 88.32 \\
Seedream 4.5~\cite{seedream45} & - & 89.24 & 94.30 & 92.14 & 92.23 & 93.83 & 88.63 \\
\hline
\rowcolor{gray!15}
\multicolumn{8}{l}{\textit{Unified Models}} \\
UniWorld-V1~\cite{lin2025uniworld} & 7B+13B & 83.64 & 88.39 & 88.44 & 89.27 & 87.22 & 81.38 \\
OmniGen2~\cite{wu2025omnigen2} & 3B+4B & 88.81 & 88.83 & 90.18 & 89.37 & 90.27 & 83.57 \\
Ovis-U1~\cite{wang2025ovis} & 2.4B+1.2B & 82.37 & 90.08 & 88.68 & 93.35 & 85.20 & 83.72 \\
Janus-Pro~\cite{chen2025janus} & 7B & 86.90 & 88.90 & 89.40 & 89.32 & 89.48 & 84.19 \\
BAGEL~\cite{deng2025bagel} & 7B+7B & 88.94 & 90.37 & 91.29 & 90.82 & 88.67 & 85.07 \\
\modelname & 2B+1.7B & 90.39 & 90.78 & 90.68 & 90.29 & 88.77 & 85.18 \\
\hline
\end{tabular}
}

\label{tab:exp_dpgbench}
\end{table}

%% file: tables/final/tiif-short.tex
\begin{table}[htbp]
\centering
\caption{\textbf{Evaluation of general text-to-image generation ability on TIIF~\cite{wei2025tiif} (Short Prompts).} Sizes of unified models in ``A + B'' indicate separate understanding (A) and generation (B) parameters. Abbreviations: \textbf{Attr}=Attribute, \textbf{Rel}=Relation, \textbf{Reas}=Reasoning, \textbf{A+R}=Attribute+Relation, \textbf{A+Re}=Attribute+Reasoning, \textbf{R+Re}=Relation+Reasoning, \textbf{RW}=Real World.}
\resizebox{\textwidth}{!}{%
\begin{tabular}{l|c|cccc|cccccc|c|c}
\hline
\multirow{2}{*}{\textbf{Model}} & \multirow{2}{*}{\textbf{\#Params}}
& \multicolumn{4}{c|}{\textbf{Basic Following}} & \multicolumn{6}{c|}{\textbf{Advanced Following}} & \multicolumn{1}{c|}{\textbf{Designer}} & \multirow{2}{*}{\textbf{Overall}} \\
\cline{3-13}
& & \textbf{Avg} & \textbf{Attr} & \textbf{Rel} & \textbf{Reas} & \textbf{Avg} & \textbf{A+R} & \textbf{A+Re} & \textbf{R+Re} & \textbf{Style} & \textbf{Text} & \textbf{RW} & \\
\hline

\rowcolor{gray!15}
\multicolumn{14}{l}{\textit{Generation Models}} \\
FLUX.1 [dev]~\cite{flux2024} & 12B & 83.1 & 87.1 & 87.3 & 75.0 & 65.8 & 67.1 & 73.8 & 69.1 & 66.7 & 43.8 & 70.7 & 71.1 \\
Z-Image~\cite{cai2025z} & 6B & 78.4 & 79.5 & 80.5 & 75.1 & 72.9 & 72.9 & 67.0 & 73.9 & 90.0 & 94.8 & 88.1 & 80.2 \\
Seedream 3.0~\cite{gao2025seedream} & - & 87.1 & 90.5 & 89.6 & 80.9 & 79.2 & 79.8 & 77.2 & 75.6 & 100.0 & 97.2 & 83.2 & 86.0 \\
Qwen-Image~\cite{wu2025qwenimagetechnicalreport} & 20B & 86.2 & 90.5 & 88.2 & 79.8 & 79.3 & 79.2 & 78.9 & 75.6 & 100.0 & 92.8 & 90.3 & 86.1 \\

\hline

\rowcolor{gray!15}
\multicolumn{14}{l}{\textit{Unified Models}} \\
Show-o~\cite{xie2024show} & 1.3B & 73.1 & 74.8 & 78.8 & 65.6 & 53.7 & 61.0 & 68.6 & 66.5 & 63.3 & 3.8 & 55.0 & 59.7 \\
Janus-Pro~\cite{chen2025janus} & 7B & 79.3 & 79.3 & 78.3 & 80.3 & 59.7 & 66.1 & 70.5 & 67.2 & 60.0 & 28.8 & 65.8 & 65.5 \\
Ovis-U1~\cite{wang2025ovis} & 2.4B+1.2B & 77.8 & 83.5 & 80.1 & 69.9 & 67.4 & 71.8 & 66.8 & 69.0 & 83.3 & 8.1 & 67.2 & 66.7 \\
BAGEL~\cite{deng2025bagel} & 7B+7B & 81.8 & 82.5 & 83.0 & 79.9 & 70.2 & 74.4 & 67.4 & 72.0 & 86.7 & 29.4 & 68.3 & 71.5 \\
Lumina-DiMOO~\cite{xin2025lumina} & 8B & 84.9 & 87.0 & 87.6 & 79.8 & 72.8 & 74.8 & 76.8 & 69.8 & 70.0 & 51.1 & 75.0 & 74.7 \\
\modelname & 2B+1.7B & 82.3 & 86.0 & 84.1 & 76.7 & 73.5 & 75.3 & 70.4 & 75.5 & 93.3 & 47.5 & 65.3 & 74.9 \\
\hline
\end{tabular}
}
\label{tab:exp_tiif_short}
\end{table}

%% file: tables/final/tiif-long.tex
\begin{table}[htbp]
\centering
\caption{\textbf{Evaluation of general text-to-image generation ability on TIIF~\cite{wei2025tiif} (Long Prompts).} Sizes of unified models in ``A + B'' indicate separate understanding (A) and generation (B) parameters. Abbreviations: \textbf{Attr}=Attribute, \textbf{Rel}=Relation, \textbf{Reas}=Reasoning, \textbf{A+R}=Attribute+Relation, \textbf{A+Re}=Attribute+Reasoning, \textbf{R+Re}=Relation+Reasoning, \textbf{RW}=Real World.}
\resizebox{\textwidth}{!}{%
\begin{tabular}{l|c|cccc|cccccc|c|c}
\hline
\multirow{2}{*}{\textbf{Model}} & \multirow{2}{*}{\textbf{\#Params}} 
& \multicolumn{4}{c|}{\textbf{Basic Following}} & \multicolumn{6}{c|}{\textbf{Advanced Following}} & \multicolumn{1}{c|}{\textbf{Designer}} & \multirow{2}{*}{\textbf{Overall}} \\
\cline{3-13}
& & \textbf{Avg} & \textbf{Attr} & \textbf{Rel} & \textbf{Reas} & \textbf{Avg} & \textbf{A+R} & \textbf{A+Re} & \textbf{R+Re} & \textbf{Style} & \textbf{Text} & \textbf{RW} & \\
\hline

\rowcolor{gray!15}
\multicolumn{14}{l}{\textit{Generation Models}} \\
FLUX.1 [dev]~\cite{flux2024} & 12B & 78.7 & 83.2 & 80.4 & 72.4 & 68.5 & 73.7 & 73.3 & 71.6 & 66.7 & 52.8 & 71.5 & 71.8 \\
Z-Image~\cite{cai2025z} & 6B & 82.8 & 86.5 & 79.9 & 81.9 & 77.0 & 77.6 & 73.8 & 75.6 & 93.3 & 93.2 & 85.5 & 83.0 \\
Seedream 3.0~\cite{gao2025seedream} & - & 84.9 & 90.1 & 85.9 & 78.9 & 80.6 & 81.8 & 78.9 & 78.6 & 93.3 & 87.8 & 83.6 & 84.3 \\
Qwen-Image~\cite{wu2025qwenimagetechnicalreport} & 20B & 87.2 & 91.5 & 90.8 & 79.4 & 80.9 & 79.8 & 81.7 & 78.6 & 100.0 & 89.1 & 91.4 & 86.8 \\
\hline

\rowcolor{gray!15}
\multicolumn{14}{l}{\textit{Unified Models}} \\
Show-o~\cite{xie2024show} & 1.3B & 75.8 & 79.8 & 78.3 & 69.3 & 50.4 & 56.8 & 69.0 & 56.2 & 66.7 & 2.8 & 50.9 & 58.9 \\
Janus-Pro~\cite{chen2025janus} & 7B & 78.3 & 82.3 & 73.3 & 79.1 & 58.8 & 56.2 & 70.8 & 60.0 & 70.0 & 33.8 & 60.3 & 65.0 \\
Ovis-U1~\cite{wang2025ovis} & 2.4B+1.2B & 79.4 & 81.5 & 81.4 & 75.2 & 67.8 & 68.3 & 73.8 & 65.9 & 86.7 & 12.7 & 68.7 & 68.2 \\
Lumina-DiMOO~\cite{xin2025lumina} & 8B & 78.0 & 81.5 & 79.8 & 72.6 & 68.5 & 74.1 & 69.1 & 66.4 & 63.3 & 40.7 & 72.0 & 68.8 \\
BAGEL~\cite{deng2025bagel} & 7B+7B & 80.1 & 83.5 & 79.9 & 76.8 & 72.2 & 75.0 & 70.1 & 74.9 & 83.3 & 33.9 & 67.9 & 71.7 \\
\modelname & 2B+1.7B & 81.5 & 81.5 & 82.2 & 80.9 & 72.7 & 76.2 & 67.6 & 75.8 & 83.3 & 50.7 & 66.8 & 73.9 \\
\hline
\end{tabular}
}
\label{tab:exp_tiif_long}
\end{table}

%% file: tables/final/oneig.tex
\begin{table}[H]
\centering
\caption{\textbf{Evaluation of general text-to-image generation ability on OneIG-EN~\cite{chang2025oneig}.} Sizes of unified models in ``A + B'' indicate separate understanding (A) and generation (B) parameters.}
\resizebox{0.9\textwidth}{!}{
\begin{tabular}{l|c|cccccc}
\hline
\textbf{Model} & \textbf{\#Params} & \textbf{Alignment} & \textbf{Text} & \textbf{Reasoning} & \textbf{Style} & \textbf{Diversity} & \textbf{Overall} \\
\hline
\rowcolor{gray!15}
\multicolumn{8}{l}{\textit{Generation Models}} \\
SDXL~\cite{mmdit} & 2.6B & 0.69 & 0.03 & 0.24 & 0.33 & 0.30 & 0.32 \\
FLUX.1 [dev]~\cite{flux2024} & 12B & 0.79 & 0.52 & 0.25 & 0.37 & 0.24 & 0.43 \\
Qwen-Image~\cite{wu2025qwenimagetechnicalreport} & 20B & 0.88 & 0.89 & 0.31 & 0.42 & 0.18 & 0.54 \\
Z-Image~\cite{cai2025z} & 6B & 0.88 & 0.99 & 0.28 & 0.39 & 0.19 & 0.55 \\
Seedream 4.5~\cite{seedream45} & - & 0.89 & 1.00 & 0.35 & 0.43 & 0.21 & 0.58 \\
Nano Banana Pro~\cite{deepmind_gemini3proimage_2025} & - & 0.89 & 0.94 & 0.33 & 0.48 & 0.25 & 0.58 \\
\hline
\rowcolor{gray!15}
\multicolumn{8}{l}{\textit{Unified Models}} \\
Janus-Pro~\cite{chen2025janus} & 7B & 0.55 & 0.00 & 0.14 & 0.28 & 0.37 & 0.27 \\
Show-o2~\cite{xie2025show} & 7B & 0.82 & 0.00 & 0.23 & 0.32 & 0.18 & 0.31 \\
Ovis-U1~\cite{wang2025ovis} & 2.4B+1.2B & 0.81 & 0.03 & 0.22 & 0.45 & 0.18 & 0.34 \\
BAGEL~\cite{deng2025bagel} & 7B+7B & 0.77 & 0.24 & 0.17 & 0.37 & 0.25 & 0.36 \\
Lumina-DiMOO~\cite{xin2025lumina} & 8B & 0.82 & 0.55 & 0.28 & 0.40 & 0.23 & 0.46 \\
OmniGen2~\cite{wu2025omnigen2} & 3B+4B & 0.80 & 0.68 & 0.27 & 0.38 & 0.24 & 0.47 \\
\modelname & 2B+1.7B & 0.82 & 0.74 & 0.27 & 0.40 & 0.25 & 0.50 \\
\hline
\end{tabular}
}

\label{tab:exp_oneigbench}
\end{table}

%% file: tables/final/oneig-zh.tex
\begin{table}[htbp]
\centering
\caption{\textbf{Evaluation of general text-to-image generation ability on OneIG-ZH~\cite{chang2025oneig}.} Sizes of unified models in ``A + B'' indicate separate understanding (A) and generation (B) parameters.}
\resizebox{0.9\textwidth}{!}{
\begin{tabular}{l|c|cccccc}
\hline
\textbf{Model} & \textbf{\#Params} & \textbf{Alignment} & \textbf{Text} & \textbf{Reasoning} & \textbf{Style} & \textbf{Diversity} & \textbf{Overall} \\
\hline

\rowcolor{gray!15}
\multicolumn{8}{l}{\textit{Generation Models}} \\
Qwen-Image~\cite{wu2025qwenimagetechnicalreport} & 20B & 0.83 & 0.96 & 0.27 & 0.41 & 0.21 & 0.53 \\
Z-Image~\cite{cai2025z} & 6B & 0.79 & 0.99 & 0.27 & 0.39 & 0.24 & 0.54 \\
Seedream 4.5~\cite{seedream45} & - & 0.83 & 0.99 & 0.30 & 0.43 & 0.21 & 0.55 \\
Nano Banana Pro~\cite{deepmind_gemini3proimage_2025} & - & 0.84 & 0.98 & 0.31 & 0.46 & 0.24 & 0.57 \\
\hline

\rowcolor{gray!15}
\multicolumn{8}{l}{\textit{Unified Models}} \\
Janus-Pro~\cite{chen2025janus} & 7B & 0.32 & 0.15 & 0.10 & 0.26 & 0.36 & 0.24 \\
BLIP-3o~\cite{chen2025blip3} & 8B & 0.61 & 0.09 & 0.21 & 0.37 & 0.23 & 0.30 \\
Lumina-DiMOO~\cite{xin2025lumina} & 8B & 0.68 & 0.15 & 0.23 & 0.37 & 0.24 & 0.33 \\ 
Ovis-U1~\cite{wang2025ovis} & 2.4B+1.2B & 0.72 & 0.15 & 0.21 & 0.43 & 0.20 & 0.34 \\ 
BAGEL~\cite{deng2025bagel} & 7B+7B & 0.67 & 0.37 & 0.19 & 0.36 & 0.27 & 0.37 \\
\modelname & 2B+1.7B & 0.75 & 0.90 & 0.23 & 0.37 & 0.26 & 0.50 \\
\hline
\end{tabular}
}

\label{tab:exp_oneigbench_zh}
\end{table}

%% file: tables/final/cvtg.tex
\begin{table}[htbp]
\centering
\caption{\textbf{Evaluation of text-centric text-to-image generation ability on CVTG-2k~\cite{du2025textcrafter}.} Sizes of unified models in ``A + B'' indicate separate understanding (A) and generation (B) parameters.}
\resizebox{0.95\textwidth}{!}{
\begin{tabular}{l|c|c|c|ccccc}
\hline
\multirow{2}{*}{\textbf{Model}} & \multirow{2}{*}{\textbf{\#Params}} & \multirow{2}{*}{\textbf{NED}} & \multirow{2}{*}{\textbf{CLIPScore}} & \multicolumn{5}{c}{\textbf{Word Accuracy}} \\
\cline{5-9}
& & & & \textbf{2 regions} & \textbf{3 regions} & \textbf{4 regions} & \textbf{5 regions} & \textbf{average} \\
\hline

\rowcolor{gray!15}
\multicolumn{9}{l}{\textit{Generation Models}} \\
FLUX.1 [dev]~\cite{flux2024} & 12B & 0.688 & 0.740 & 0.609 & 0.553 & 0.466 & 0.432 & 0.497 \\
Nano Banana Pro~\cite{deepmind_gemini3proimage_2025} & - & 0.875 & 0.737 & 0.737 & 0.775 & 0.786 & 0.793 & 0.779 \\
Qwen-Image~\cite{wu2025qwenimagetechnicalreport} & 20B & 0.912 & 0.802 & 0.837 & 0.836 & 0.831 & 0.816 & 0.829 \\
Z-Image~\cite{cai2025z} & 6B & 0.937 & 0.797 & 0.901 & 0.872 & 0.865 & 0.851 & 0.867 \\
Seedream 4.5~\cite{seedream45} & - & 0.948 & 0.807 & 0.878 & 0.895 & 0.908 & 0.901 & 0.899 \\
\hline
\rowcolor{gray!15}
\multicolumn{9}{l}{\textit{Unified Models}} \\
Ovis-U1~\cite{wang2025ovis} & 2.4B+1.2B & 0.477 & 0.725 & 0.133 & 0.109 & 0.091 & 0.065 & 0.093  \\ 
BAGEL~\cite{deng2025bagel} & 7B+7B & 0.657 & 0.779 & 0.498 & 0.391 & 0.332 & 0.291 & 0.356 \\
Lumina-DiMOO~\cite{xin2025lumina} & 8B & 0.805 & 0.831 & 0.723 & 0.646 & 0.571 & 0.505 & 0.590 \\
\modelname & 2B+1.7B & 0.804 & 0.816 & 0.729 & 0.660 & 0.618 & 0.549 & 0.623 \\
\hline
\end{tabular}
}

\label{tab:exp_cvtg}
\end{table}

%% file: tables/final/longtext.tex
\begin{table}[htbp]
\centering
\caption{\textbf{Evaluation of text-centric text-to-image generation ability on LongText-Bench~\cite{geng2025x}.} Sizes of unified models in ``A + B'' indicate separate understanding (A) and generation (B) parameters.}
\resizebox{0.75\textwidth}{!}{
\begin{tabular}{l|c|cc}
\hline
\textbf{Model} & \textbf{\#Params} & \textbf{LongText-Bench-EN} & \textbf{LongText-Bench-ZH} \\
\hline
\rowcolor{gray!15}
\multicolumn{4}{l}{\textit{Generation Models}} \\
FLUX.1 [dev]~\cite{flux2024} & 12B & 0.607 & 0.005 \\
Z-Image~\cite{cai2025z} & 6B & 0.943 & 0.946 \\
Qwen-Image~\cite{wu2025qwenimagetechnicalreport} & 20B & 0.943 & 0.946 \\
Nano Banana Pro~\cite{deepmind_gemini3proimage_2025} & - & 0.981 & 0.949 \\
Seedream 4.5~\cite{seedream45} & - & 0.989 & 0.987 \\
\hline

\rowcolor{gray!15}
\multicolumn{4}{l}{\textit{Unified Models}} \\
Janus-Pro~\cite{chen2025janus} & 7B & 0.019 & 0.006 \\
BLIP-3o~\cite{chen2025blip3} & 8B & 0.021 & 0.018 \\
Ovis-U1~\cite{wang2025ovis} & 2.4B+1.2B & 0.030 & 0.051 \\
BAGEL~\cite{deng2025bagel} & 7B+7B & 0.373 & 0.310 \\
Lumina-DiMOO~\cite{xin2025lumina} & 8B & 0.437 & 0.047 \\
OmniGen2~\cite{xiao2024omnigen} & 3B+4B & 0.561 & 0.059 \\
\modelname & 2B+1.7B & 0.738 & 0.860 \\
\hline
\end{tabular}
}

\label{tab:exp_longtext}
\end{table}

%% file: tables/final/wise.tex
\begin{table}[H]
\centering
\caption{\textbf{Evaluation of knowledge-informed text-to-image generation ability on WISE~\cite{niu2025wise}.} Sizes of unified models in ``A + B'' indicate separate understanding (A) and generation (B) parameters.}
\resizebox{\textwidth}{!}{
\begin{tabular}{l|c|ccccccc}
\hline
\textbf{Model} & \textbf{\#Params} & \textbf{Cultural} & \textbf{Time} & \textbf{Space} & \textbf{Biology} & \textbf{Physics} & \textbf{Chemistry} & \textbf{Overall} \\
\hline
\rowcolor{gray!15} 
\multicolumn{9}{l}{\textit{Generation Models}} \\
SD3-Medium~\cite{mmdit} & 2B & 0.43 & 0.50 & 0.52 & 0.41 & 0.53 & 0.33 & 0.45 \\
FLUX.1 [dev]~\cite{flux2024} & 12B & 0.48 & 0.58 & 0.62 & 0.42 & 0.51 & 0.35 & 0.50 \\
Qwen-Image~\cite{wu2025qwenimagetechnicalreport} & 20B & 0.63 & 0.62 & 0.76 & 0.60 & 0.72 & 0.39 & 0.63 \\
\hline
\rowcolor{gray!15} 
\multicolumn{9}{l}{\textit{Unified Models}} \\
Janus-Pro~\cite{chen2025janus} & 7B & 0.30 & 0.37 & 0.49 & 0.36 & 0.42 & 0.26 & 0.35 \\
Lumina-DiMOO~\cite{xin2025lumina} & 8B & 0.35 & 0.43 & 0.59 & 0.31 & 0.49 & 0.34 & 0.40 \\
Ovis-U1~\cite{wang2025ovis} & 2.4B+1.2B & 0.36 & 0.46 & 0.64 & 0.35 & 0.52 & 0.28 & 0.42 \\
BAGEL~\cite{deng2025bagel} & 7B+7B & 0.44 & 0.52 & 0.65 & 0.42 & 0.62 & 0.41 & 0.49 \\
UniWorld-V1~\cite{lin2025uniworld} & 7B+13B & 0.53 & 0.55 & 0.73 & 0.45 & 0.59 & 0.41 & 0.55 \\
\modelname & 2B+1.7B & 0.37 & 0.51 & 0.68 & 0.39 & 0.62 & 0.39 & 0.46 \\
\modelname (w/ CoT) & 2B+1.7B & 0.55 & 0.57 & 0.74 & 0.51 & 0.72 & 0.46 & 0.58 \\
\hline
\end{tabular}
}

\label{tab:exp_wise}
\end{table}

%% file: tables/final/genexam.tex
\begin{table}[htbp]
\centering
\caption{\textbf{Evaluation of knowledge-informed text-to-image generation ability on GenExam~\cite{GenExam} (Relaxed Scores).} Sizes of unified models in ``A + B'' indicate separate understanding (A) and generation (B) parameters. Abbreviations: \textbf{Math}=Mathematics, \textbf{Phy}=Physics, \textbf{Chem}=Chemistry, \textbf{Bio}=Biology, \textbf{Geo}=Geography, \textbf{Comp}=Computer Science, \textbf{Eng}=Engineering, \textbf{Econ}=Economics, \textbf{Hist}=History.
}

\resizebox{\textwidth}{!}{
\begin{tabular}{l|c|ccccccccccc}
\hline
\textbf{Model} & \textbf{\#Params} & \textbf{Math} & \textbf{Phy} & \textbf{Chem} & \textbf{Bio} & \textbf{Geo} & \textbf{Comp} & \textbf{Eng} & \textbf{Econ} & \textbf{Music} & \textbf{Hist} & \textbf{Overall} \\
\hline

\rowcolor{gray!15}
\multicolumn{13}{l}{\textit{Generation Models}} \\
FLUX.1 [dev]~\cite{flux2024} & 12B & 12.2 & 14.4 & 12.5 & 22.8 & 36.4 & 11.0 & 14.0 & 9.2 & 21.3 & 21.7 & 17.6 \\
HunyuanImage-3.0~\cite{cao2025hunyuanimage} & - & 17.0 & 17.2 & 18.8 & 18.7 & 30.4 & 15.5 & 16.9 & 11.7 & 23.9 & 20.4 & 19.1 \\
Qwen-Image~\cite{wu2025qwenimagetechnicalreport} & 20B & 18.9 & 26.3 & 15.3 & 32.1 & 49.6 & 18.9 & 32.0 & 20.3 & 23.4 & 38.6 & 27.5 \\
Seedream 4.5~\cite{seedream45} & - & 44.7 & 63.4 & 48.9 & 75.8 & 67.6 & 57.9 & 69.7 & 67.3 & 38.0 & 55.0 & 58.8 \\
GPT-Image-1.5~\cite{wu2025qwenimagetechnicalreport} & - & 65.8 & 85.4 & 78.1 & 91.9 & 92.5 & 75.8 & 86.4 & 85.5 & 70.8 & 90.9 & 82.3 \\
Nano Banana Pro~\cite{deepmind_gemini3proimage_2025} & - & 86.3 & 95.1 & 88.7 & 95.9 & 96.5 & 91.7 & 95.1 & 97.2 & 91.0 & 99.9 & 93.7 \\
\hline

\rowcolor{gray!15}
\multicolumn{13}{l}{\textit{Unified Models}} \\
BLIP-3o~\cite{chen2025blip3} & 8B & 6.4 & 5.5 & 4.7 & 7.0 & 16.7 & 3.6 & 8.4 & 2.5 & 6.0 & 11.2 & 7.2 \\
Janus-Pro~\cite{chen2025janus} & 7B & 13.7 & 8.8 & 8.2 & 7.2 & 18.8 & 3.9 & 10.5 & 4.2 & 14.5 & 6.6 & 9.6 \\
Ovis-U1~\cite{wang2025ovis} & 2.4B+1.2B & 12.2 & 10.8 & 6.6 & 10.0 & 25.4 & 6.1 & 8.8 & 5.4 & 13.2 & 15.1 & 11.4 \\
BAGEL~\cite{deng2025bagel} & 7B+7B & 14.7 & 10.6 & 7.9 & 10.8 & 24.5 & 6.8 & 10.2 & 5.3 & 13.7 & 14.4 & 11.9 \\
Show-o2~\cite{xie2025show} & 7B & 10.8 & 11.9 & 4.8 & 12.8 & 33.3 & 4.7 & 11.8 & 7.0 & 8.8 & 14.5 & 12.0 \\
\modelname & 2B+1.7B & 21.5 & 22.2 & 19.3 & 20.0 & 31.2 & 9.9 & 19.6 & 21.5 & 17.8 & 24.9 & 20.8 \\
\modelname (w/ CoT) & 2B+1.7B & 25.6 & 24.2 & 23.5 & 23.6 & 35.6 & 12.0 & 21.4 & 24.4 & 18.4 & 20.3 & 22.9 \\
\hline
\end{tabular}
}

\label{tab:exp_genexam}
\end{table}

%% file: tables/final/imgedit.tex
\begin{table}[h]
\centering
\caption{\textbf{Evaluation of general image editing ability on ImgEdit~\cite{ye2025imgedit}.} Sizes of unified models in ``A + B'' indicate separate understanding (A) and generation (B) parameters.}
\resizebox{\textwidth}{!}{
\begin{tabular}{l|c|cccccccccc}
\hline
\textbf{Model} & \textbf{\#Params} & \textbf{Add} & \textbf{Adjust} & \textbf{Extract} & \textbf{Replace} & \textbf{Remove} & \textbf{Background} & \textbf{Style} & \textbf{Hybrid} & \textbf{Action} & \textbf{Overall} \\
\hline

\rowcolor{gray!15}
\multicolumn{12}{l}{\textit{Generation Models}} \\
FLUX.1 Kontext~\cite{labs2025flux1kontextflowmatching} & 12B & 4.25 & 4.15 & 2.35 & 4.56 & 3.57 & 4.26 & 4.57 & 3.68 & 4.63 & 4.00 \\
GPT-Image-1 [High]~\cite{GPT-Image-1} & - & 4.61 & 4.33 & 2.90 & 4.35 & 3.66 & 4.57 & 4.93 & 3.96 & 4.89 & 4.20 \\
Qwen-Image-Edit~\cite{wu2025qwenimagetechnicalreport} & 20B & 4.38 & 4.16 & 3.43 & 4.66 & 4.14 & 4.38 & 4.81 & 3.82 & 4.69 & 4.27 \\
Z-Image-Edit~\cite{cai2025z} & 6B & 4.40 & 4.14 & 4.30 & 4.57 & 4.13 & 4.14 & 4.85 & 3.63 & 4.50 & 4.30 \\
\hline

\rowcolor{gray!15}
\multicolumn{12}{l}{\textit{Unified Models}} \\
Lumina-DiMOO~\cite{xin2025lumina} & 8B & 3.41 & 2.38 & 1.90 & 3.26 & 2.21 & 2.11 & 4.19 & 2.26 & 3.17 & 2.77 \\
BAGEL~\cite{deng2025bagel} & 7B+7B & 3.56 & 3.31 & 1.70 & 3.30 & 2.62 & 3.24 & 4.49 & 2.38 & 4.17 & 3.20 \\
UniWorld-V1~\cite{lin2025uniworld} & 20B & 3.82 & 3.64 & 2.27 & 3.47 & 3.24 & 2.99 & 4.21 & 2.96 & 2.74 & 3.26 \\
OmniGen2~\cite{wu2025omnigen2} & 3B+4B & 3.57 & 3.06 & 1.77 & 3.74 & 3.20 & 3.57 & 4.81 & 2.52 & 4.68 & 3.44 \\
Ovis-U1~\cite{wang2025ovis} & 2.4B+1.2B & 3.99 & 3.73 & 2.66 & 4.38 & 4.15 & 4.05 & 4.86 & 3.43 & 4.68 & 3.97 \\
\modelname & 2B+1.7B & 4.13 & 3.40 & 2.27 & 4.13 & 3.39 & 3.84 & 4.77 & 3.03 & 4.05 & 3.67 \\
\modelname (w/ CoT) & 2B+1.7B & 4.24 & 3.80 & 2.58 & 4.36 & 3.51 & 3.92 & 4.69 & 3.00 & 4.31 & 3.82 \\
\hline
\end{tabular}
}

\label{tab:exp_imgedit}
\end{table}

%% file: tables/final/gedit.tex
\begin{table}[htbp]
\centering
\caption{\textbf{Evaluation of general image editing ability on GEdit-Bench~\cite{liu2025step1x-edit}.} Sizes of unified models in ``A + B'' indicate separate understanding (A) and generation (B) parameters. Abbreviations: \textbf{BC}=Background Change, \textbf{CA}=Color Alteration, \textbf{MM}=Material Modification, \textbf{MC}=Motion Change, \textbf{PB}=Portrait Beautification, \textbf{ST}=Style Transfer, \textbf{SA}=Subject Addition, \textbf{SR}=Subject Removal, \textbf{SRp}=Subject Replacement, \textbf{TM}=Text Modification, \textbf{TT}=Tone Transfer.
}
\resizebox{\textwidth}{!}{
\begin{tabular}{l|c|ccccccccccc|c}
\hline
\textbf{Models} & \textbf{\#Params} & \textbf{BC} & \textbf{CA} & \textbf{MM} & \textbf{MC} & \textbf{PB} & \textbf{ST} & \textbf{SA} & \textbf{SR} & \textbf{SRp} & \textbf{TM} & \textbf{TT} & \textbf{Avg/G\_O} \\
\hline

\rowcolor{gray!15}
\multicolumn{14}{l}{\textit{Generation Models}} \\
GPT Image 1~\cite{GPT-Image-1} & - & 6.96 & 6.85 & 7.10 & 5.41 & 6.74 & 7.44 & 7.51 & 8.73 & 8.55 & 8.45 & 8.69 & 7.49 \\
Qwen-Image-Edit~\cite{wu2025qwenimagetechnicalreport} & 20B & 8.23 & 8.30 & 7.33 & 8.05 & 7.49 & 6.74 & 8.57 & 8.09 & 8.29 & 8.48 & 8.50 & 8.01 \\
\hline

\rowcolor{gray!15}
\multicolumn{14}{l}{\textit{Unified Models}} \\
Lumina-DiMOO~\cite{xin2025lumina} & 8B & 3.43 & 4.27 & 3.08 & 2.77 & 4.74 & 5.19 & 4.44 & 3.80 & 4.38 & 2.68 & 4.20 & 3.91 \\
Ovis-U1~\cite{wang2025ovis} & 2.4B+1.2B & 7.49 & 6.88 & 6.21 & 4.79 & 5.98 & 6.46 & 7.49 & 7.25 & 7.27 & 4.48 & 6.31 & 6.42 \\
BAGEL~\cite{deng2025bagel} & 7B+7B & 7.32 & 6.91 & 6.38 & 4.75 & 4.57 & 6.15 & 7.90 & 7.16 & 7.02 & 7.32 & 6.22 & 6.52 \\
\modelname & 2B+1.7B & 7.08 & 7.05 & 6.38 & 7.02 & 6.03 & 6.27 & 7.13 & 6.55 & 6.33 & 6.59 & 6.85 & 6.66 \\
\modelname (w/ CoT) & 2B+1.7B & 7.05 & 7.87 & 6.50 & 6.99 & 5.77 & 6.10 & 7.33 & 7.16 & 7.12 & 7.36 & 6.46 & 6.88 \\
\hline
\end{tabular}
}

\label{tab:exp_gedit}
\end{table}

%% file: tables/final/textedit-rule.tex
\begin{table}[htbp]
\centering
\setlength{\tabcolsep}{3pt}
\caption{\textbf{Evaluation of text-centric image editing on TextEdit (Classic Metrics).} Sizes of unified models in ``A + B'' indicate separate understanding (A) and generation (B) parameters. ``Real'' refers to source images from real-world scene, while ``Virtual'' refers to images from virtual scene. Abbreviations: \textbf{OA}=OCR Accuracy, \textbf{OP}=OCR Precision, \textbf{OR}=OCR Recall, \textbf{F1}=OCR F1-Score, \textbf{NED}=ROI-Aware NED, \textbf{CLIP}=CLIPScore, \textbf{AES}=Aesthetic Score. For detailed evaluation metric, please refer to Appendix Section~\ref{sec:textedit}.}
\label{tab:exp_textedit_rule}
\resizebox{\textwidth}{!}{
\begin{tabular}{l|c|ccccccc|ccccccc}
\hline
\multirow{2}{*}{\textbf{Models}} & \multirow{2}{*}{\textbf{\# Params}} & \multicolumn{7}{c|}{\textbf{Real}} & \multicolumn{7}{c}{\textbf{Virtual}} \\
\cline{3-9}
\cline{10-16}
& & \textbf{OA} & \textbf{OP} & \textbf{OR} & \textbf{F1} & \textbf{NED} & \textbf{CLIP} & \textbf{AES}
  & \textbf{OA} & \textbf{OP} & \textbf{OR} & \textbf{F1} & \textbf{NED} & \textbf{CLIP} & \textbf{AES} \\
\hline

\multicolumn{16}{l}{\cellcolor{gray!15}\textit{Generation Models}} \\
Qwen-Image-Edit~\cite{wu2025qwenimagetechnicalreport} & 20B
& 0.75 & 0.68 & 0.66 & 0.67 & 0.71 & 0.75 & 5.72
& 0.78 & 0.75 & 0.73 & 0.74 & 0.75 & 0.81 & 5.21 \\
GPT-Image-1.5~\cite{GPT-Image-1.5} & -
& 0.74 & 0.69 & 0.67 & 0.68 & 0.68 & 0.75 & 5.78
& 0.73 & 0.72 & 0.71 & 0.71 & 0.70 & 0.80 & 5.28 \\
Nano Banana Pro~\cite{deepmind_gemini3proimage_2025} & -
& 0.77 & 0.72 & 0.70 & 0.71 & 0.72 & 0.75 & 5.79
& 0.80 & 0.78 & 0.77 & 0.78 & 0.78 & 0.81 & 5.28 \\
\hline

\multicolumn{16}{l}{\cellcolor{gray!15}\textit{Unified Models}} \\
Lumina-DiMOO~\cite{xin2025lumina} & 8B
& 0.22 & 0.23 & 0.19 & 0.20 & 0.19 & 0.69 & 5.53
& 0.22 & 0.25 & 0.21 & 0.22 & 0.20 & 0.72 & 4.76 \\
Ovis-U1~\cite{wang2025ovis} & 2.4B+1.2B
& 0.40 & 0.37 & 0.34 & 0.35 & 0.35 & 0.72 & 5.32
& 0.37 & 0.40 & 0.38 & 0.39 & 0.33 & 0.75 & 4.66 \\
BAGEL~\cite{deng2025bagel} & 7B+7B
& 0.60 & 0.59 & 0.53 & 0.55 & 0.55 & 0.74 & 5.71
& 0.57 & 0.60 & 0.56 & 0.57 & 0.54 & 0.78 & 5.19 \\
\modelname & 2B+1.7B
& 0.77 & 0.73 & 0.70 & 0.71 & 0.72 & 0.75 & 5.70
& 0.79 & 0.77 & 0.75 & 0.75 & 0.77 & 0.80 & 5.12 \\
\hline
\end{tabular}
}
\end{table} 

%% file: tables/final/textedit-mllm.tex
\begin{table}[htbp]
\centering
\caption{\textbf{Evaluation of text-centric image editing on TextEdit (MLLM-based Metrics).} Sizes of unified models in ``A + B'' indicate separate understanding (A) and generation (B) parameters. ``Real'' refers to source images from real-world scene, while ``Virtual'' refers to images from virtual scene. Abbreviations: \textbf{TA}: Text Accuracy, \textbf{TP}: Text Preservation, \textbf{SI}: Scene Integrity,  \textbf{LR}: Local Realism, \textbf{VC}: Visual Coherence, \textbf{Avg}: MLLM Overall Average. For detailed evaluation metric, please refer to Appendix Section~\ref{sec:textedit}.}

\resizebox{\textwidth}{!}{
\begin{tabular}{l|c|cccccc|cccccc}
\hline
\multirow{2}{*}{\textbf{Models}} & \multirow{2}{*}{\textbf{\# Params}} & \multicolumn{6}{c|}{\textbf{Real}} & \multicolumn{6}{c}{\textbf{Virtual}} \\
\cline{3-8}
\cline{9-14}
& & \textbf{TA} & \textbf{TP} & \textbf{SI} & \textbf{LR} & \textbf{VC} & \textbf{Avg} & \textbf{TA} & \textbf{TP} & \textbf{SI} & \textbf{LR} & \textbf{VC} & \textbf{Avg} \\
\hline

\rowcolor{gray!15}
\multicolumn{14}{l}{\textit{Generation Models}} \\
Qwen-Image-Edit~\cite{wu2025qwenimagetechnicalreport} & 20B & 0.92 & 0.82 & 0.75 & 0.57 & 0.80 & 0.77 & 0.57 & 0.79 & 0.92 & 0.80 & 0.77 & 0.77 \\
GPT-Image-1.5~\cite{GPT-Image-1.5} & - & 0.96 & 0.94 & 0.86 & 0.80 & 0.93 & 0.90 & 0.82 & 0.93 & 0.96 & 0.91 & 0.87 & 0.90 \\
Nano Banana Pro~\cite{deepmind_gemini3proimage_2025} & - & 0.96 & 0.95 & 0.85 & 0.88 & 0.93 & 0.91 & 0.87 & 0.92 & 0.96 & 0.94 & 0.89 & 0.92 \\
\hline

\rowcolor{gray!15}
\multicolumn{14}{l}{\textit{Unified Models}} \\
Lumina-DiMOO~\cite{xin2025lumina} & 8B & 0.17 & 0.06 & 0.04 & 0.02 & 0.05 & 0.09 & 0.02 & 0.06 & 0.16 & 0.05 & 0.03 & 0.08 \\
Ovis-U1~\cite{wang2025ovis} & 2.4B+1.2B & 0.31 & 0.12 & 0.12 & 0.07 & 0.18 & 0.18 & 0.06 & 0.16 & 0.31 & 0.14 & 0.13 & 0.19 \\
BAGEL~\cite{deng2025bagel} & 7B+7B & 0.68 & 0.60 & 0.38 & 0.35 & 0.56 & 0.53 & 0.38 & 0.51 & 0.68 & 0.62 & 0.42 & 0.54 \\
\modelname & 2B+1.7B & 0.94 & 0.90 & 0.71 & 0.80 & 0.80 & 0.88 & 0.87 & 0.86 & 0.91 & 0.82 & 0.62 & 0.83 \\
\hline
\end{tabular}
}

\label{tab:exp_textedit_mllm}
\end{table}

%% file: tables/final/rise.tex
\begin{table}[htbp]
\centering
\caption{\textbf{Evaluation of reasoning-informed image editing ability on RISEBench~\cite{zhao2025envisioning}.} Sizes of unified models in ``A + B'' indicate separate understanding (A) and generation (B) parameters. Abbreviations: \textbf{IR}=Instruction Reasoning, \textbf{AC}=Appearance Consistency, \textbf{VP}=Visual Plausibility.}
\resizebox{0.9\textwidth}{!}{
\begin{tabular}{l|c|ccccc|ccc}
\hline
\textbf{Models} & \textbf{\#Params} & \textbf{Temporal} & \textbf{Causal} & \textbf{Spatial} & \textbf{Logical} & \textbf{Overall} & \textbf{IR} & \textbf{AC} & \textbf{VP} \\
\hline

\rowcolor{gray!15}
\multicolumn{10}{l}{\textit{Generation Models}} \\
FLUX.1 Kontext~\cite{labs2025flux1kontextflowmatching} & 12B & 2.3 & 5.5 & 13.0 & 1.2 & 5.8 & 26.0 & 71.6 & 85.2 \\
Qwen-Image-Edit~\cite{wu2025qwenimagetechnicalreport} & 20B & 4.7 & 10.0 & 17.0 & 2.4 & 8.9 & 37.2 & 66.4 & 86.9 \\
Seedream 4.0~\cite{seedream2025seedream} & - & 12.9 & 12.2 & 11.0 & 7.1 & 10.8 & 58.9 & 67.4 & 91.2 \\
Nano Banana Pro~\cite{deepmind_gemini3proimage_2025} & - & 41.2 & 61.1 & 48.0 & 37.6 & 47.2 & 77.0 & 85.5 & 94.4 \\
GPT-Image-1.5~\cite{GPT-Image-1.5} & - & 54.1 & 60.0 & 62.0 & 21.2 & 50.0 & 69.7 & 92.5 & 94.9 \\
\hline

\rowcolor{gray!15}
\multicolumn{10}{l}{\textit{Unified Models}} \\
Lumina-DiMOO~\cite{xin2025lumina} & 8B & 2.4 & 1.1 & 4.0 & 1.2 & 2.2 & 34.0 & 50.7 & 72.3 \\
Ovis-U1~\cite{wang2025ovis} & 2.4B+1.2B & 1.2 & 3.3 & 4.0 & 2.4 & 2.8 & 33.9 & 52.7 & 72.9 \\
BAGEL~\cite{deng2025bagel} & 7B+7B & 2.4 & 5.6 & 14.0 & 1.2 & 6.1 & 36.5 & 53.5 & 73.0 \\
\modelname & 2B+1.7B & 3.5 & 2.2 & 5.0 & 3.5 & 3.6 & 35.6 & 52.7 & 75.9 \\
\modelname (w/ CoT) & 2B+1.7B & 4.7 & 7.8 & 1.8 & 5.9 & 9.4 & 43.9 & 64.4 & 79.7 \\
\hline
\end{tabular}
}

\label{tab:exp_rise}
\end{table}

%% file: sections/6.conclusion.tex
\section{Conclusion}

In this work, we presented \modelname, a unified multimodal model that effectively democratizes the capabilities of understanding, reasoning, generation, and editing. By adhering to the principles of unified context modeling with modality-specific modularity and decoupled visual representations, our architecture seamlessly integrates generative capabilities into a strong understanding backbone. To further bridge the gap between high-level intelligence and visual generation, we introduced a comprehensive data synthesis pipeline with the Chain-of-Thought (CoT) paradigm, enabling the model to align abstract user intent with precise visual execution. Empirical results confirm that \modelname not only excels in knowledge-intensive generation and editing but also retains competitive performance in multimodal understanding and reasoning benchmarks. We hope \modelname serves as a robust baseline and accelerates the community's progress toward developing comprehensive, omni-capable AGI-oriented UMMs.

%% file: sections/appendix.tex
\section{TextEdit Benchmark}
\label{sec:textedit}

\subsection{Design Motivation}
With the increasing adoption of text-to-image and image editing models in real-world applications, \textbf{text-centric image editing} has become a frequent requirement in advertising design, poster revision, UI localization, and commercial asset updates. However, existing general-purpose image editing models remain unreliable when processing text content. On the one hand, the generated text often suffers from misspellings, distorted glyphs, broken multi-line layouts, or unnatural blending with the background. On the other hand, while replacing text, models frequently inadvertently alter non-target regions (e.g., material textures, facial details, or background structures), effectively turning ``edit the text'' into ``edit the whole image''. Moreover, as is shown in Table~\ref{tab:benchmarks-class}, existing text-centric benchmarks (e.g., AnyText~\cite{tuo2023anytext}, LongText~\cite{geng2025x}, and CVTG-2K~\cite{du2025textcrafter}) primarily focus on the accuracy of text \textit{generation}, lacking adequate modeling for specific text \textit{editing} scenarios. Although text editing benchmarks like MARIO-Eval-edit~\cite{lan2025flux} exist, their image sources cover only a limited set of text scenarios, failing to fully reflect the diverse text carriers and layout forms encountered in both real-world and synthetic design materials. Furthermore, these benchmarks lack a systematic evaluation of edit faithfulness and visual preservation.
\input{tables/text-bench-compare}

To comprehensively evaluate the capabilities of image editing models, we introduce \textbf{TextEdit}, a novel and meticulously structured benchmark. TextEdit addresses the limitations of existing text-centric editing benchmarks by providing a more systematic, fine-grained, and human-curated evaluation framework. It is distinguished by the following key advantages:
\begin{itemize}
    \item \textbf{Human-curated Data Pipeline:} Unlike prior benchmarks that rely heavily on synthetic or automatically collected data, TextEdit adopts a human-filtered pipeline to ensure high-quality, realistic editing scenarios and reduce noisy or ambiguous samples.

    \item \textbf{Manually Annotated Ground Truth:} We provide \textbf{manually edited Ground Truth (GT) images}, enabling precise quantitative evaluation. This supports reliable computation of pixel-level fidelity metrics and accurate assessment of background preservation.

    \item \textbf{Fine-grained Scenario Taxonomy:} TextEdit covers \textbf{18 diverse sub-classes} of text editing scenarios, offering a more systematic and detailed evaluation compared to existing benchmarks with limited or coarse categorizations.

    \item \textbf{Alignment with LLM-based Instructions:} The benchmark follows a pure text-instruction paradigm aligned with modern LLM-based interaction. It removes the need for auxiliary inputs such as glyph maps or segmentation masks, enabling a more natural evaluation of instruction-following ability.

    \item \textbf{Hybrid Evaluation Protocol:} We combine classic OCR, image-fidelity metrics and modern multimodal LLM-based evaluation across target accuracy, text preservation, scene integrity, local realism and visual coherence.  This dual-track protocol enables comprehensive assessment.
\end{itemize}

\subsection{Design Details}
Grounded in practical user scenarios, TextEdit bridges the gap between diverse real-world needs and model assessment through a systematic \textbf{Scenario Taxonomy} and a robust \textbf{Evaluation Protocol}. This integrated design enables a comprehensive evaluation of both technical text-editing performance and practical usability.
\subsubsection{Scenario Taxonomy}

\begin{figure}[h]
    \centering
    \includegraphics[width=0.85\linewidth]{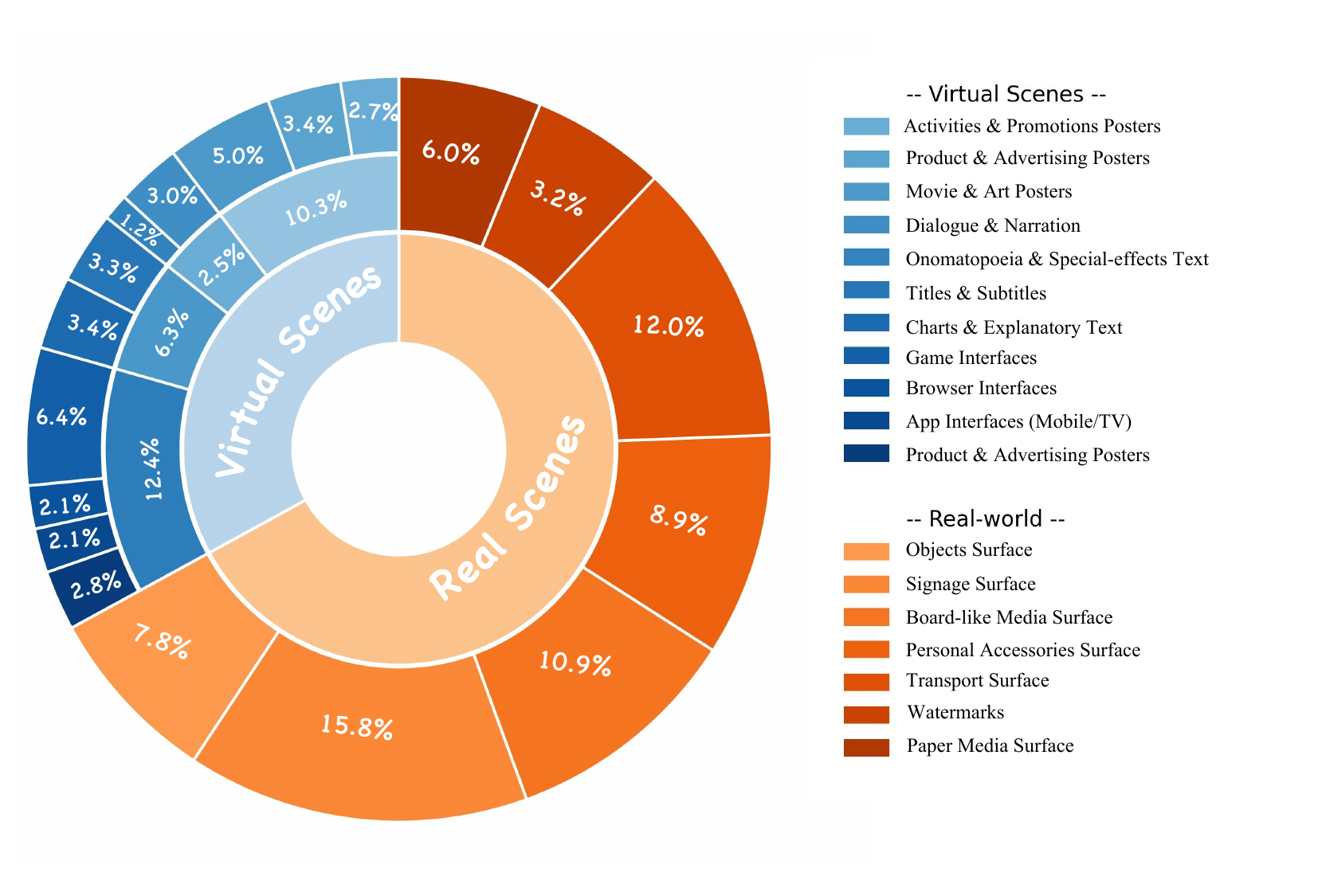} 
    \vspace{-1em}
    \caption{\textbf{Data Distribution of TextEdit benchmark.}}
    \label{fig:text-bench}
\end{figure}

We organize text-centric editing scenarios into two overarching domains: \textbf{Virtual Scenes} and \textbf{Real-world Scenes}. This fine-grained taxonomy, detailed with definitions and statistics in \cref{tab:benchmarks-class}, captures the nuances of diverse text carriers, ranging from digital formats like \textit{posters},~\textit{comic},~\textit{slide} and \textit{GUIs} to real-world environments such as \textit{products},~\textit{buliding},~\textit{board-like media},~\textit{personal accessorie},~\textit{transport},~\textit{watermarks} and \textit{paper media}. \cref{fig:text-bench} further illustrates the distribution of these subcategories. To construct this dataset, we curated high-quality source images aligning with these scenarios and synthesized editing instructions via a rule-based mechanism. The resulting Ground Truth images underwent rigorous manual verification to ensure visual naturalness and instruction faithfulness, establishing a reliable gold standard for quantitative evaluation.

\vspace{-1em}
\input{tables/text-bench-classification}

\input{tables/text-bench-evluation}

\subsubsection{Evaluation Metrics}
\label{sec:metrics}

Quantitative evaluation of text editing is challenging due to the dual requirement of manipulating specific text content while strictly preserving the background. To provide a holistic assessment, we employ a hybrid evaluation strategy combining \textbf{Classic Metrics} and \textbf{MLLM-based Metrics}.

\textbf{Classic Metrics} focus on the existence and correctness of text. We utilize standard OCR tools to measure edit distance and detection rates. Specifically, we decouple the evaluation into \textit{Target Region} (measuring editing success) and \textit{Background Region} (measuring preservation capabilities). Additionally, we use CLIPScore~\cite{hessel2021clipscore} to assess general image quality and semantic alignment, along with aesthetic quality evaluation. 

\textbf{MLLM-based Metrics} are further introduced to better capture visual nuances such as ``ghosting'' artifacts, lighting inconsistencies, or partial erasure. By simulating an expert forensic analysis, we use the most powerful multimodal understanding model Gemini-3-Pro~\cite{Gemini-3-Pro} as judge to provide fine-grained scoring on dimensions like local realism and scene integrity, offering an evaluation that aligns closer with human preference.

Below are the detailed implementations of each evaluation metric, organized following the structure in Table~\ref{tab:benchmark_metrics}.

\vspace{1em}
\noindent\textbf{(a) Classic Metrics (Text-Centric)}

\vspace{0.5em}
We first define a normalized similarity function $S(s_1, s_2)$ based on the Levenshtein distance $D_{lev}(\cdot, \cdot)$, which forms the foundation for our text-centric metrics:
\begin{equation}
    S(s_1, s_2) = 1 - \frac{D_{lev}(s_1, s_2)}{\max(|s_1|, |s_2|, 1)}
\end{equation}
where $s_1$ and $s_2$ denote two comparison strings, and $|s_1|, |s_2|$ represent their respective lengths. The term $\max(|s_1|, |s_2|, 1)$ serves as a normalization factor to ensure the denominator is non-zero, bounding the similarity score $S$ within the range $[0, 1]$, where $1$ indicates a perfect match.

\noindent\textit{(i) OCR Accuracy.} 
This metric evaluates whether the target text is correctly rendered in the editing region. Let $\mathcal{T}_{gen}$ be the set of text strings detected in the generated image that significantly overlap with the target editing region (Intersection over Union, $\text{IoU} > 0.5$). The accuracy is defined as the maximum similarity between the detected texts and the ground truth target text $t_{tgt}$, adjusted by a penalty factor:
\begin{equation}
    \text{Acc} = \max_{t \in \mathcal{T}_{gen}} S(t, t_{tgt}) \times \mathbb{P}_{fail}
\end{equation}
Here, $t$ represents a candidate string within the set $\mathcal{T}_{gen}$. The term $\mathbb{P}_{fail}$ is a penalty coefficient designed to punish failed editing attempts. Specifically, $\mathbb{P}_{fail}$ is set to $0.2$ if the original source text $t_{src}$ is still detected in the region while the target text $t_{tgt}$ is absent; otherwise, $\mathbb{P}_{fail} = 1.0$.

\noindent\textit{(ii) OCR Precision.} 
This metric measures the accuracy of background text preservation, penalizing hallucinated or incorrect background text. For each detected text item $t_i$ in the background region (those with $\text{IoU} < 0.5$ with the target region), we find its best match among the original background texts $\mathcal{T}_{bg}^{orig}$ from the source image:
\begin{equation}
    \text{Precision} = \frac{1}{|\mathcal{T}_{bg}^{gen}|} \sum_{t \in \mathcal{T}_{bg}^{gen}} \max_{t' \in \mathcal{T}_{bg}^{orig}} S(t, t')
\end{equation}
where $\mathcal{T}_{bg}^{gen}$ denotes the set of background text detected in the generated image, and $|\mathcal{T}_{bg}^{gen}|$ is its cardinality. This metric achieves a high score when all detected background texts closely match the original ones, thus avoiding spurious text generation.

\noindent\textit{(iii) OCR Recall.} 
This metric assesses the completeness of background text preservation, penalizing missing background text. For each original background text $t'$ in $\mathcal{T}_{bg}^{orig}$, we measure how well it is preserved in the generated image:
\begin{equation}
    \text{Recall} = \frac{1}{|\mathcal{T}_{bg}^{orig}|} \sum_{t' \in \mathcal{T}_{bg}^{orig}} \max_{t \in \mathcal{T}_{bg}^{gen}} S(t, t')
\end{equation}
A high recall indicates that most of the original background texts are successfully retained in the edited image without being accidentally removed or altered.

\noindent\textit{(iv) OCR F1-Score.} 
The F1-score provides a balanced metric by computing the harmonic mean of OCR Precision and OCR Recall:
\begin{equation}
    \text{F1} = 2 \times \frac{\text{Precision} \times \text{Recall}}{\text{Precision} + \text{Recall}}
\end{equation}
This unified metric captures both the accuracy and completeness of background text preservation, offering a comprehensive assessment of text-level editing quality.

\noindent\textit{(v) ROI-Aware NED.}
This metric strictly evaluates the editing quality within the specific Region of Interest (ROI) defined by the original source text bounding box. We calculate the similarity between the predicted text string $t_{pred}$ extracted from the ROI and the target string $t_{tgt}$. A \textit{failed erasure penalty} is applied if the residual similarity to the source text remains high:
\begin{equation}
    \text{NED} = S(t_{pred}, t_{tgt}) \times \mathbb{I}_{resid}
\end{equation}
where $t_{pred}$ denotes the OCR recognition result directly cropped from the ROI. $\mathbb{I}_{resid}$ is a residual indicator function acting as a penalty term: $\mathbb{I}_{resid} = 0.2$ if $S(t_{pred}, t_{src}) > 0.9$ (implying the source text $t_{src}$ was not effectively erased), and $\mathbb{I}_{resid} = 1.0$ otherwise. This metric ensures that the target region contains the correct new text while completely removing the original text.

\vspace{1em}
\noindent\textbf{(b) Classic Metrics (General)}

\vspace{0.5em}
\noindent\textit{(i) CLIPScore.} 
We use CLIPScore~\cite{hessel2021clipscore} to measure the semantic alignment between the predicted edited image and the caption text. First, we employ Qwen3-VL~\cite{Qwen3-VL} to generate a concise caption  text for the labeled GT image, with a length that satisfies the input constraints of the CLIP text encoder. Let $\mathbf{v}_{img}$ denote the CLIP visual embedding of the predicted edited image, and $\mathbf{v}_{text}$ denote the CLIP text embedding of the generated caption. The CLIPScore is computed as the cosine similarity:
\begin{equation}
    \text{CLIPScore} = \frac{\mathbf{v}_{img} \cdot \mathbf{v}_{text}}{|\mathbf{v}_{img}| \cdot |\mathbf{v}_{text}|}
\end{equation}
A higher CLIPScore indicates better semantic consistency between the visual content and the textual description, reflecting successful integration of the edited text into the scene context.

\noindent\textit{(ii) Aesthetic Score.} 
We use a CLIP-based aesthetic predictor to evaluate the overall visual appeal of the generated image. Let $f_{\text{aes}}(\cdot)$ denote the aesthetic prediction model that maps a CLIP image embedding to an aesthetic quality score:
\begin{equation}
    \text{AesScore} = f_{\text{aes}}(\mathbf{v}_{img})
\end{equation}
The aesthetic score typically ranges from 1 to 5 (or 0 to 1 after normalization), where higher values indicate better visual quality, composition, and perceptual appeal. This metric helps ensure that text editing does not degrade the overall image quality.

\vspace{1em}
\noindent\textbf{(c) MLLM-based Metrics}

\vspace{0.5em}
We utilize a strong commercial MLLM (\ie Gemini-3-Pro-Preview~\cite{Gemini-3-Pro}) to simulate expert forensic analysis across $N=5$ dimensions ($D_1$ to $D_5$). Each dimension is evaluated on a Likert scale from 1 to 5, where higher scores indicate better quality. The raw scores are then normalized and aggregated as follows.

\vspace{0.3em}
\noindent\textit{(i) Target Accuracy ($D_1$).}
This dimension evaluates the spelling correctness and erasure quality of the target text in the edited region. The MLLM assigns a score $s_1 \in [1, 5]$ based on:
\begin{itemize}
    \item Whether the target text is rendered with correct spelling
    \item Whether the original source text has been completely erased
    \item The visual clarity and legibility of the new text
\end{itemize}
A score of 5 indicates perfect text replacement with no residual artifacts from the source text.

\noindent\textit{(ii) Text Preservation ($D_2$).}
This dimension assesses whether non-target background text remains intact and unaffected by the editing operation. The MLLM evaluates:
\begin{itemize}
    \item Completeness of background text retention
    \item Absence of unintended modifications to surrounding text
    \item Preservation of text layout and positioning
\end{itemize}
The score $s_2 \in [1, 5]$ reflects the degree to which the model successfully preserves the original background text elements.

\noindent\textit{(iii) Scene Integrity ($D_3$).}
This dimension measures the stability of background geometry and objects, checking for structural distortions or artifacts introduced by the editing process. The evaluation criteria include:
\begin{itemize}
    \item Preservation of architectural elements, object boundaries, and spatial relationships
    \item Absence of geometric distortions or warping effects
    \item Maintenance of scene perspective and depth cues
\end{itemize}
The score $s_3 \in [1, 5]$ indicates how well the overall scene structure is maintained.

\noindent\textit{(iv) Local Realism ($D_4$).}
This dimension evaluates the quality of inpainting in the edited region, focusing on:
\begin{itemize}
    \item Edge cleanness and seamlessness between edited and original regions
    \item Absence of visible artifacts such as blurring, ghosting, or seams
    \item Natural integration of the new text with its immediate surroundings
\end{itemize}
The score $s_4 \in [1, 5]$ reflects the photorealistic quality of the local editing region.

\noindent\textit{(v) Visual Coherence ($D_5$).}
This dimension assesses the harmony of font style, lighting, and texture with the original scene context. The MLLM evaluates:
\begin{itemize}
    \item Font style consistency with surrounding text or scene aesthetics
    \item Lighting direction, intensity, and color temperature matching
    \item Shadow casting and texture patterns that blend naturally with the background
\end{itemize}
The score $s_5 \in [1, 5]$ measures how well the edited text integrates into the overall visual style of the image.

\vspace{0.3em}
\noindent\textit{(vi) Score Normalization and Cutoff Mechanism.}
The raw Likert scores $s_i \in [1, 5]$ for each dimension $i$ are first normalized to $s'_i \in [0, 1]$ using the mapping: $s'_i = (s_i - 1) / 4$. To ensure evaluation validity, we implement a \textbf{Cutoff Mechanism}: if the primary editing task (Target Text Accuracy, $D_1$) fails significantly (i.e., $s_1 < 4$), the scores for secondary dimensions are penalized to zero, as a failed text edit renders other quality assessments meaningless. The final weighted score $V_{score}$ is formulated as:
\begin{equation}
    V_{score} = w_1 s'_1 + \mathbb{I}_{(s_1 \ge 4)} \cdot \sum_{i=2}^{5} w_i s'_i
\end{equation}
In this formulation:
\begin{itemize}
    \item $s'_i$ denotes the normalized score for the $i$-th dimension.
    \item $w_i$ represents the weight assigned to the $i$-th dimension, satisfying $\sum_{i=1}^{5} w_i = 1$. By default, we use $w_1 = 0.4$, $w_2 = 0.3$, $w_3 = 0.1$, $w_4 = 0.1$, $w_5 = 0.1$.
    \item $s_1$ is the raw score for the primary text accuracy dimension.
    \item $\mathbb{I}_{(s_1 \ge 4)}$ is an indicator function that equals $1$ if the condition $s_1 \ge 4$ is met, and $0$ otherwise. This ensures that if the textual content is incorrect (score $<4$), the visual quality of the background (dimensions $D_2$ to $D_5$) does not contribute to the final score.
\end{itemize}

\vspace{0.3em}
\noindent\textit{(vii) MLLM Overall Average.}
The overall MLLM-based metric is defined as the weighted average across all five dimensions with the cutoff mechanism applied:
\begin{equation}
    \text{MLLM Overall Avg} = V_{score} = w_1 s'_1 + \mathbb{I}_{(s_1 \ge 4)} \cdot \sum_{i=2}^{5} w_i s'_i
\end{equation}
This comprehensive metric provides a single scalar value that captures both the primary text editing success and the secondary visual quality factors, with appropriate penalties for fundamental editing failures. The metric is computed separately for \textbf{Virtual} (synthetic scenes, category 1.x.x) and \textbf{Real} (real-world scenes, category 2.x) subsets of the benchmark to enable fine-grained performance analysis across different scene types.

\begin{figure}[H]
\begin{AcademicBox}[\footnotesize System Prompt for T2I task]
    {
    
    You are an expert Forensic Image Analyst and Design QA Specialist. 
    
    Your task is to evaluate the quality of an AI-edited image by comparing three images. 
    
    \textbf{Images Provided (in order):}
    \begin{enumerate}[label=\arabic*., nosep, leftmargin=1.5em]
        \item \textbf{Original Image}: The unedited source image containing the text ``\{raw\_text\}''. 
        \item \textbf{Ground Truth Image}: A human-created reference showing the ideal result with text ``\{target\_text\}''. 
        \item \textbf{Edited Image}: The AI-generated result to be evaluated. 
    \end{enumerate}
    
    \textbf{Editing Task Information:}
    \begin{itemize}[nosep, leftmargin=1.5em]
        \item \textbf{Text to Remove}: ``\{raw\_text\}''
        \item \textbf{Text to Add}: ``\{target\_text\}''
    \end{itemize}
    
    \textbf{EVALUATION RUBRIC (1-5 SCORING SYSTEM)}

    Please evaluate the \textbf{Edited Image} based on the following 5 dimensions. Use the strict criteria below to assign a score from 1 to 5.
    } 
\end{AcademicBox}

\begin{AcademicBox}[\footnotesize Text Accuracy]
    \textbf{Q1. [Target Text Accuracy]}
    \par
    \emph{Focus: Spelling, erasure correctness, and legibility of ``\{target\_text\}''.}
    \begin{itemize}[nosep, leftmargin=1.5em, label=\textbullet]
        \item \textbf{5 (Perfect)}: Exact spelling match (case-sensitive). Old text completely erased. No ghosting.
        \item \textbf{4 (Minor Flaw)}: Text is correct but has 1 character error/typo, OR slight casing issue, OR extremely faint ghosting visible only on close inspection.
        \item \textbf{3 (Readable but Flawed)}: 2--3 character errors but word is recognizable. OR visible ghosting/remnants of old text that affect cleanness.
        \item \textbf{2 (Major Error)}: $>$3 character errors (misspelled heavily). OR old text is still clearly readable (failed erasure).
        \item \textbf{1 (Failed)}: Text is missing, gibberish, or completely wrong word. Old text remains fully intact.
    \end{itemize}
\end{AcademicBox}

\begin{AcademicBox}[\footnotesize Text Preservation]
    \textbf{Q2. [Non-Target Text Preservation]}
    \par
    \emph{Focus: Preservation/legibility of background text other than the edited target.}
    \begin{itemize}[nosep, leftmargin=1.5em, label=\textbullet]
        \item \textbf{5 (Perfect)}: All non-target text is 100\% preserved and legible, identical to Original/GT.
        \item \textbf{4 (Good)}: Main background text is preserved. Minor distant text is slightly softened/blurred but still readable.
        \item \textbf{3 (Fair)}: One or two secondary text elements are blurred, damaged, or missing.
        \item \textbf{2 (Poor)}: Critical nearby text (directly adjacent to target) is damaged, erased, or hallucinated.
        \item \textbf{1 (Destructive)}: Widespread destruction or hallucination of background text.
    \end{itemize}
\end{AcademicBox}

\begin{AcademicBox}[\footnotesize Scene Integrity]
    \textbf{Q3. [Global Scene Integrity]}
    \par
    \emph{Focus: Geometric stability of non-edited areas (background, objects, people).}
    \begin{itemize}[nosep, leftmargin=1.5em, label=\textbullet]
        \item \textbf{5 (Perfect)}: Pixel-perfect preservation of background geometry. No distortions.
        \item \textbf{4 (Good)}: Almost perfect, but very minor shift ($<$1\%) in background lines or perspective.
        \item \textbf{3 (Noticeable)}: Visible distortion in straight lines (wavy), or slight warping of objects/faces.
        \item \textbf{2 (Severe)}: Major structural damage (e.g., a person's face is melted, a building collapsed).
        \item \textbf{1 (Chaos)}: The scene structure is completely changed or nonsensical compared to Original.
    \end{itemize}
\end{AcademicBox}

\begin{AcademicBox}[\footnotesize Local Realism]
    \textbf{Q4. [Local Realism \& Artifacts]}
    \par
    \emph{Focus: Inpainting quality, edge cleanliness, and seamlessness around the edited area.}
    \begin{itemize}[nosep, leftmargin=1.5em, label=\textbullet]
        \item \textbf{5 (Excellent)}: Invisible edit. Clean edges, no halos, no smudges. Professional quality.
        \item \textbf{4 (Good)}: Very minor artifacts (e.g., slight pixelation on zoom-in), but looks natural at a glance.
        \item \textbf{3 (Fair)}: Visible seams, blurry rectangular patch, or ``smudged'' look around the text.
        \item \textbf{2 (Poor)}: Obvious artifacts, messy edges, or white/black box artifacts.
        \item \textbf{1 (Garbage)}: The edited area looks like a corrupted file or pure noise.
    \end{itemize}
\end{AcademicBox}
\caption{\textbf{The system prompt template used for MLLM-based automated evaluation}, covering the analyst persona, task definition, and the first four scoring dimensions (Q1--Q4).}
\label{fig:eval_prompt_part1}
\end{figure}

\begin{figure}[H]
\begin{AcademicBox}[\footnotesize Visual Coherence]
    \textbf{Q5. [Aesthetic \& Lighting Harmony]}
    \par
    \emph{Focus: Style matching (font), lighting, shadow, and texture harmony.}
    \begin{itemize}[nosep, leftmargin=1.5em, label=\textbullet]
        \item \textbf{5 (Seamless)}: Font style matches the GT/Context perfectly. Lighting/shadows are physically correct. Texture (grain) matches the photo.
        \item \textbf{4 (Integrated)}: Good style match. Lighting is mostly correct. Texture is slightly too smooth but acceptable.
        \item \textbf{3 (Artificial)}: Text looks ``pasted on'' (digital sticker look). Font style is generic (e.g., Arial) and clashes with the scene.
        \item \textbf{2 (Disjointed)}: Wrong color, wrong perspective, or no shading where needed.
        \item \textbf{1 (Mismatch)}: Text floats awkwardly, completely ignoring the scene's physics and style.
    \end{itemize}
\end{AcademicBox}

\begin{AcademicBox}[\footnotesize Final Output Format]
    \textbf{\large FINAL OUTPUT FORMAT (JSON ONLY)}
    \par
    You must output a valid JSON object containing two dictionaries:score (integers) and reason (strings).
    \par
    \textbf{Example Output:}
    \par
    \vspace{0.2em}
    \noindent
    \textbf{Instruction:} \textit{"Replace the text \textquotesingle MUSIC\textquotesingle\ with \textquotesingle PARTY\textquotesingle ."}
    
    \noindent
    \begin{minipage}[t]{0.60\textwidth}
        \vspace{0pt} 
        \begin{tcolorbox}[colback=gray!10, boxrule=0pt, sharp corners, left=5pt, top=5pt, bottom=5pt]
        \{ \\
        \hspace*{1em} "score": \{ \\
        \hspace*{2em} "Q1": 5, \\
        \hspace*{2em} "Q2": 1, \\
        \hspace*{2em} "Q3": 2, \\
        \hspace*{2em} "Q4": 5, \\
        \hspace*{2em} "Q5": 4 \\
        \hspace*{1em} \}, \\
        \hspace*{1em} "reason": \{ \\
        \hspace*{2em} "Q1": "The target text \textquotesingle PARTY\textquotesingle\ is spelled correctly and is clearly legible. The specific target text to remove ( \textquotesingle MUSIC\textquotesingle\ ) is completely gone with no ghosting.", \\
        \hspace*{2em} "Q2": "The model caused widespread destruction of non-target text. \textquotesingle NIGHT CLUB\textquotesingle\ , \textquotesingle 31 OCT\textquotesingle\ , \textquotesingle FREE DRINKS\textquotesingle\ , \textquotesingle LIVE\textquotesingle\ , and \textquotesingle PRICE\textquotesingle\ were all erroneously removed, and \textquotesingle 10\$\textquotesingle\ was corrupted into the hallucinated text \textquotesingle 1TY\textquotesingle\ .", \\
        \hspace*{2em} "Q3": "Global scene integrity is severely compromised. The skeleton\textquotesingle s arm holding the maraca was erased, leaving the maraca floating in mid-air, which breaks the physical logic of the illustration.", \\
        \hspace*{2em} "Q4": "Despite the semantic failures, the technical quality of the image is excellent. The edges are sharp, the background inpainting is smooth, and there are no visible pixel artifacts, blur, or noise.", \\
        \hspace*{2em} "Q5": "The font style selected for \textquotesingle PARTY\textquotesingle\ integrates well with the hand-drawn vector aesthetic of the poster, although the color is a darker maroon compared to the bright red of the original text." \\
        \hspace*{1em} \} \\
        \}
        \end{tcolorbox}
    \end{minipage}%
    \hfill
    \begin{minipage}[t]{0.38\textwidth}
        \vspace{0pt} 
        \centering
        \includegraphics[width=0.8\linewidth]{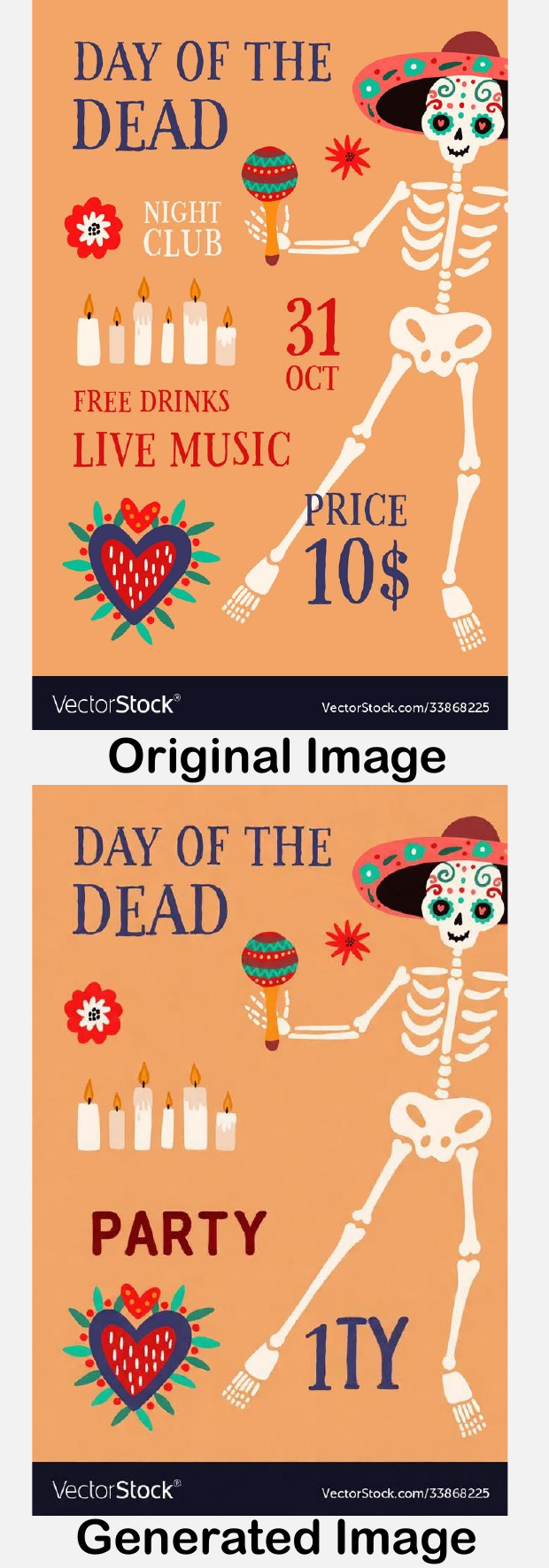}
        \par
        \footnotesize \textit{}
    \end{minipage}
    
    \par
    Do not output any markdown or conversational text outside the JSON block.
\end{AcademicBox}

\caption{\textbf{The continuation of the evaluation prompt}, detailing the final scoring dimension (Q5) and the strict JSON output schema required for parsing results.}
\label{fig:eval_prompt_part2}
\end{figure}

\subsection{MiniSet-500 Results}
To provide a lightweight and standardized evaluation subset for the open-source community, we construct a \textbf{MiniSet-500} from the full TextEdit benchmark. It is built by randomly sampling instances from each of the 18 subcategories to ensure a balanced distribution across scenario types. The MiniSet-500 contains a total of \textbf{500 image editing pairs}, preserving the diversity of tasks while significantly reducing evaluation cost. It serves as an efficient protocol for rapid benchmarking and ablation studies, while the full benchmark remains the standard for comprehensive evaluation. In Table~\ref{tab:exp_textedit_rule_miniset}  and Table~\ref{miniset-textedit-mllm}, we report the performance of various models on the MiniSet-500 TextEdit benchmark.
\input{tables/final/textedit-rule-miniset}
\input{tables/final/textedit-mllm-miniset}

\clearpage
\section{Data Construction Details}

In this section, we provide details of our data construction process.

\subsection{Filtering details for General Science Image Generation}
\label{sec:appendix-general-science}

The second stage of filtering is based on the following dimensions:

\begin{enumerate}[itemsep=5pt, topsep=0pt, parsep=0pt]
\item Image Type: images are categorized into one of the 30 image types defined in MMMU~\cite{MMMU}, \eg \texttt{Posters, Diagrams, Screenshots}. Certain image types are removed depending on the subject, like \texttt{Microscopic\_Images} for biology or \texttt{Tables} for all subjects. 

\item Subject: images are classified into a list of give subjects and removed if they belong to subjects like Literature, Art, Design, \etc, since those images are usually natural images like paintings or photographs.

\item Text Length: all text in the images is extracted. Images with text longer than 200 characters are removed.

\item Image Complexity: images are rated in terms of their complexity from 1 (very complex) to 10 (very simple), \ie whether the image is too complex to draw, considering the number and complexity of components, objects, and text. We set different ranges (\eg 5-7) for each subject to filter images of moderate complexity.

\item Subject Knowledge Density: similar to Image Complexity, images are rated by whether the model needs subject knowledge and reasoning to generate them, from 1 (least knowledge) to 10 (very dense knowledge). Different ranges are also applied for each subject, \eg 7-8.
\end{enumerate}

\subsection{Chemistry Text-to-Image Data Synthesis}
In addition to the filtered general science text-to-image data from existing datasets and the Internet, we also devised an automated pipeline to acquire large-scale complex organic compound data for chemistry images. In the initial phase, we employed automated acquisition protocols to harvest 800,000 raw entries from PubChem, a publicly accessible and authoritative chemical repository. This raw corpus encompasses essential physicochemical descriptors, including unique Compound IDs (CIDs), chemical nomenclature, molecular formulas, SMILES representations, and 2D structural diagrams. Subsequently, we implemented a rigorous filtration mechanism to sanitize the dataset by eliminating erroneous or incomplete entries, yielding a curated set of 600,000 high-fidelity chemical instances. In the second phase, we focused on the construction of instruction-following data. By establishing diverse QA templates that integrate chemical names, SMILES strings, and their corresponding 2D visualizations, we successfully synthesized 600,000 visual-textual QA pairs specifically optimized for chemical image generation tasks.

\subsection{Computer Science Editing}
\label{sec:appendix-computer-science}

The definitions of each task are given below:

\begin{enumerate}[itemsep=5pt, topsep=0pt, parsep=0pt]

\item Tree Topology Editing \& Node Manipulation: perform completion (complete into a full binary tree), insertion (insert a new node at a specific position) or pruning (delete a node and its subtree or nodes within a specific region) on a tree.

\item Tree Traversal Visualization: visualize the traversal path or label the traversal order of a binary tree, including pre-order, in-order and post-order.

\item Binary Search Tree (BST) Operations: insert a node into a BST, or validate a BST and fix the swapped nodes.

\item Heap Operations \& Dual Views: perform 1-3 steps of insert or extract-root operations on heaps and visualize the memory array. Sift-up and sift-down processes should be included.

\item Huffman Coding Tree: given the frequency of characters, construct a Huffman coding tree or perform node merging.

\item Tree Lowest Common Ancestor (LCA) \& Path Highlighting: given two nodes, highlight their LCA or draw the connected path between them.

\item Graph $k$-hop Neighborhood: given a central node, visually label all nodes with an exact distance of $k$.

\item Graph Degree Identification: identify and box all nodes whose degree equals a specific value.

\item Graph Cycle Detection: depict the unique simple cycle in the graph.

\item Bipartite Graph Coloring: color an uncolored bipartite graph with two colors with the constraint that adjacent nodes must have different colors.

\item Graph Shortest Path: given a start node and an end node, draw the shortest connecting path between the two nodes.

\item Directed Graph Reachability: in a directed graph, box the complete set of downstream nodes reachable in the forward direction from the source node.

\item FSM String Trace: given an input string, draw the complete state transition path in the FSM.

\item FSM State Role Identification: identify and color the start state and accepting states in the FSM.

\item FSM Transition Logic Completion: complete the missing arrowed edge and its input label in the incomplete FSM.

\end{enumerate}

\subsection{Solid Geometry}

Implementations of each task are as follows:

\begin{enumerate}[itemsep=5pt, topsep=0pt, parsep=0pt]

\item Solid of Revolution: We first construct a family of 3D solids of revolution using GeoGebra, parameterized by three possible rotation axes (x, y, or z). Taking the z-axis as an example, we generate a planar polygon in the xoz plane whose one edge lies on the rotation axis. Specifically, we fix the y-coordinate to 0 and sample a sequence of vertices with monotonically increasing integer z-coordinates. The x-coordinates are sampled as random integers that are either strictly positive (polygon lies on the positive x half-axis) or strictly negative (polygon lies on the negative x half-axis). The first and last vertices are constrained to lie exactly on the z-axis (x = 0), ensuring that one side of the polygon coincides with the rotation axis. The number of vertices in the polygon is randomly chosen in the range $[3, 6]$. This polygon then serves as the generator curve (meridian), which defines the rotational surface when swept around the specified axis.
\item Plane Symmetry: Plane symmetry is also implemented with GeoGebra. To maintain visually pleasing layouts, we restrict the symmetry plane to be perpendicular to the xoy plane and ensure that the original solid lies entirely on one side of this plane prior to reflection. To enrich the diversity of the dataset, we support multiple configurable geometric primitives, including regular prisms, regular pyramids, cylinders, cones, and spheres. For prisms and pyramids, the number of edges of the regular base polygon is sampled in the range $[3, 6]$, and the edge length is configurable. In addition, both the color of the solid and the color of the symmetry plane are parameterized, allowing us to generate samples with varied appearance while preserving geometric clarity and visual aesthetics.

\item Point Symmetry: For point symmetry, we implement two parallel sample generation pipelines based on GeoGebra and matplotlib, respectively. In both pipelines, the type of geometric primitive (\eg prisms, pyramids, cylinders, cones, spheres) and its color are randomly configurable, enabling a wide variety of instance appearances. To further increase the diversity of the dataset, we apply a random initial rotation to each base solid and allow the camera viewpoint to be varied across samples.

\item Solid Translation: The translation task is implemented using a matplotlib-based rendering pipeline, in which the type and color of the 3D solids are fully configurable. To keep the visual effect both aesthetically pleasing and controllable, we restrict the translation vector to three canonical configurations: along the x-axis, y-axis, or z-axis. For each sample, the translation magnitude is randomly chosen as an integer in the range 
$[4,10]$, which introduces sufficient variation in the relative positions of the original and translated solids while ensuring that both remain clearly visible within the same view.

\item Solid Projection: The projection task is implemented in matplotlib, with the type and color of 3D solids randomly sampled. Each solid is first rotated by a configurable angle to avoid axis-aligned degeneracies, and then orthographically projected onto the xoy plane.

\end{enumerate}

%% file: tables/text-bench-compare.tex
\begin{table}[H]
\centering
\caption{\textbf{Key attributes comparison of open-source text generation or editing benchmarks.}}
\label{tab:benchmarks-class}
\small
\renewcommand{\arraystretch}{1.2}
\setlength{\tabcolsep}{3pt}

\begin{tabular*}
{\linewidth}{l@{\extracolsep{\fill}}ccccccc}
\toprule
\textbf{Benchmarks} & \textbf{Type} & \textbf{Size} & \textbf{Human Filter} & \textbf{GT Ann.} & \textbf{Sub-class} & \textbf{Traditial Eval.} & \textbf{LLM Eval.} \\
\midrule
AnyText~\cite{tuo2023anytext} & Text Generation & 2,000 & {\color{red}\ding{55}} & {\color{red}\ding{55}} & - & {\color{green!60!black}\checkmark} & {\color{red}\ding{55}} \\
LongText~\cite{geng2025x} & Text Generation & 320 & {\color{green!60!black}\checkmark} & {\color{red}\ding{55}} & 8 & {\color{red}\ding{55}} & {\color{green!60!black}\checkmark} \\
CVTG-2K~\cite{du2025textcrafter} & Text Generation & 2,000 & {\color{green!60!black}\checkmark} & {\color{red}\ding{55}} & 2 & {\color{green!60!black}\checkmark} & {\color{red}\ding{55}} \\
MARIO-Eval-edit~\cite{lan2025flux} & Text Edit & 4,000 & {\color{red}\ding{55}} & {\color{red}\ding{55}} & - & {\color{green!60!black}\checkmark} & {\color{red}\ding{55}} \\

\midrule
\textbf{TextEdit (Ours)} & Text Edit& 2,148 & {\color{green!60!black}\checkmark} & {\color{green!60!black}\checkmark} & \textbf{18} & {\color{green!60!black}\checkmark} & {\color{green!60!black}\checkmark} \\
\bottomrule
\end{tabular*}
\end{table}

%% file: tables/text-bench-classification.tex
\begin{table}[H]
\centering
\small
\renewcommand{\arraystretch}{0.8} 
\setlength{\tabcolsep}{3pt}

\caption{\textbf{Unified benchmark taxonomy with data statistics.}}
\label{tab:taxonomy_counts}

\begin{tabularx}{\linewidth}{c c l c X c}
\toprule
\textbf{Major} & \textbf{ID} & \textbf{Category (Mid)} & \textbf{ID} & \textbf{Specific Scene (Sub)} & \textbf{Count} \\
\midrule

\multirow{15}{*}{\rotatebox{90}{\textbf{Virtual Scenes}}} 
& \multirow{3}{*}{1.1} & \multirow{3}{*}{Poster Scenes} 
    & 1.1.1 & Activities \& Promotions Posters & 57 \\ 
& & & 1.1.2 & Product \& Advertising Posters & 74 \\
& & & 1.1.3 & Movie \& Art Posters & 107 \\
\cmidrule(l){2-6} 
& \multirow{2}{*}{1.2} & \multirow{2}{*}{Comic Scenes}
    & 1.2.1 & Dialogue \& Narration & 65 \\
& & & 1.2.2 & Onomatopoeia \& Special-effects Text & 26 \\
\cmidrule(l){2-6}
& \multirow{2}{*}{1.3} & \multirow{2}{*}{Slide / Presentation}
    & 1.3.1 & Titles \& Subtitles & 71 \\
& & & 1.3.2 & Charts \& Explanatory Text & 73 \\
\cmidrule(l){2-6}
& \multirow{4}{*}{1.4} & \multirow{4}{*}{GUI Scenes}
    & 1.4.1 & Game Interfaces & 138 \\
& & & 1.4.2 & Browser Interfaces & 44 \\
& & & 1.4.3 & App Interfaces (Mobile/TV) & 45 \\
& & & 1.4.4 & Operating-System Desktops & 61 \\

\midrule 

\multirow{7}{*}{\rotatebox{90}{\textbf{Real-world}}} 
& 2.1 & Objects Surface & - &  e.g., Packages, Bottles, Boxes, Coins & 168 \\
& 2.2 & Signage Surface &-&e.g., Building Signs, Storefronts, Billboards & 339 \\
& 2.3 & Board-like Media Surface&-&e.g., Blackboards, Whiteboards & 235 \\
& 2.4 & Personal Accessories Surface &-&e.g., Clothing Prints, Badges
& 192 \\
& 2.5 & Transport Surface  &  -& e.g., Cars, Buses, Trains, Ships
 & 257 \\
& 2.6 & Watermarks &  - & e.g., Photo watermarks, Brand Marks, Corner Stamps & 69 \\
& 2.7 & Paper Media Surface&-&e.g., Papers, Books, Newspapers, Menus  & 127 \\

\midrule 
\textbf{Total} & \multicolumn{4}{c}{} & \textbf{2148} \\ 

\bottomrule
\end{tabularx}

\end{table}

%% file: tables/text-bench-evluation.tex
\begin{table}[h]
\centering
\small
\setlength{\tabcolsep}{6pt}
\renewcommand{\arraystretch}{1.25}
\caption{\textbf{Overview of evaluation metrics.} We categorize metrics into objective (Text-Centric, General) and perceptual (MLLM-based) dimensions. Detailed formulations are provided in Section~\ref{sec:metrics}.}
\label{tab:benchmark_metrics}
\begin{tabular}{l l p{10cm}} 
\toprule
\textbf{Category} & \textbf{Metric} & \textbf{What it Measures} \\
\midrule
\multirow{8}{*}{\makecell[l]{\textbf{Classic}\\(Text-Centric)}}
& \textbf{OCR Accuracy} & Maximum similarity between the generated text in the target region and the ground-truth string. \\
& \textbf{OCR Precision} & Accuracy of background text preservation (penalizes hallucinated or incorrect background text). \\
& \textbf{OCR Recall} & Completeness of background text preservation (penalizes missing background text). \\
& \textbf{OCR F1-Score} & Harmonic mean of OCR Precision and OCR Recall. \\
& \textbf{ROI-Aware NED} & Normalized Edit Distance specifically within the source text bounding box. \\
\hdashline
\multirow{2}{*}{\makecell[l]{\textbf{Classic}\\(General)}}
& \textbf{CLIPScore} & Semantic alignment between the edited image and the target caption. \\
& \textbf{Aesthetic Score} & Visual appeal score predicted by a CLIP-based aesthetic predictor. \\
\midrule
\multirow{5}{*}{\textbf{MLLM-based}} 
& \textbf{Target Accuracy} & Evaluation of spelling correctness and erasure quality of the target text. \\
& \textbf{Text Preservation} & Assessment of whether non-target background text remains intact. \\
& \textbf{Scene Integrity} & Stability of background geometry and objects (checking for distortions). \\
& \textbf{Local Realism} & Quality of inpainting edges, checking for artifacts like blurring or seams. \\
& \textbf{Visual Coherence} & Harmony of font style, lighting, and texture with the original scene. \\
& \textbf{MLLM Overall Avg} & Weighted average of the MLLM-based sub-scores (40/30/10/10/10). \\
\bottomrule
\end{tabular}
\end{table}

%% file: tables/final/textedit-rule-miniset.tex
\begin{table}[htbp]
\centering
\setlength{\tabcolsep}{3pt}
\caption{\textbf{Evaluation of text-centric image editing on TextEdit MiniSet-500 (Classic Metrics).} Sizes of unified models in ``A + B'' indicate separate understanding (A) and generation (B) parameters. ``Real'' refers to source images from real-world scenes, while ``Virtual'' refers to images from synthetic virtual. Abbreviations: \textbf{OA}=OCR Accuracy, \textbf{OP}=OCR Precision, \textbf{OR}=OCR Recall, \textbf{F1}=OCR F1-Score, \textbf{NED}=ROI-Aware NED, \textbf{CLIP}=CLIPScore, \textbf{AES}=Aesthetic Score.}
\resizebox{\textwidth}{!}{
\begin{tabular}{l|c|ccccccc|ccccccc}
\noalign{\vspace{1pt}} 
\hline
\noalign{\vspace{1pt}} 
\multirow{2}{*}{\textbf{Models}} & 
\multirow{2}{*}{\textbf{\# Params}} & 
\multicolumn{7}{c}{\textbf{Real}} & 
\multicolumn{7}{c}{\textbf{Virtual}} \\
\cline{3-9}
\cline{10-16}
& &
\textbf{OA} & \textbf{OP} & \textbf{OR} & \textbf{F1} & \textbf{NED} & \textbf{CLIP} & \textbf{AES} &
\textbf{OA} & \textbf{OP} & \textbf{OR} & \textbf{F1} & \textbf{NED} & \textbf{CLIP} & \textbf{AES} \\
\hline

\multicolumn{16}{l}{\cellcolor{gray!15}\textit{Generation Models}} \\
Qwen-Image-Edit~\cite{wu2025qwenimagetechnicalreport} & 20B
& 0.76 & 0.69 & 0.67 & 0.67 & 0.70 & 0.75 & 5.81
& 0.74 & 0.71 & 0.70 & 0.70 & 0.70 & 0.80 & 5.27 \\
GPT-Image-1.5~\cite{GPT-Image-1.5} & -
& 0.72 & 0.68 & 0.66 & 0.67 & 0.67 & 0.75 & 5.85
& 0.68 & 0.69 & 0.68 & 0.68 & 0.65 & 0.80 & 5.32 \\
Nano Banana Pro~\cite{deepmind_gemini3proimage_2025} & -
& 0.76 & 0.71 & 0.69 & 0.70 & 0.70 & 0.75 & 5.86
& 0.77 & 0.76 & 0.75 & 0.75 & 0.76 & 0.81 & 5.32 \\
\hline

\multicolumn{16}{l}{\cellcolor{gray!15}\textit{Unified Models}} \\
Lumina-DiMOO~\cite{xin2025lumina} & 8B
& 0.20 & 0.22 & 0.18 & 0.19 & 0.19 & 0.70 & 5.58
& 0.22 & 0.25 & 0.21 & 0.22 & 0.19 & 0.73 & 4.87 \\
Ovis-U1~\cite{wang2025ovis} & 2.4B+1.2B
& 0.37 & 0.34 & 0.32 & 0.32 & 0.33 & 0.72 & 5.39
& 0.39 & 0.41 & 0.38 & 0.39 & 0.33 & 0.74 & 4.75 \\
BAGEL~\cite{deng2025bagel} & 7B+7B
& 0.61 & 0.59 & 0.52 & 0.54 & 0.54 & 0.74 & 5.79
& 0.53 & 0.58 & 0.53 & 0.55 & 0.51 & 0.78 & 5.25 \\
\modelname & 2B+1.7B
& 0.77 & 0.74 & 0.70 & 0.71 & 0.71 & 0.76 & 5.79
& 0.74 & 0.72 & 0.69 & 0.70 & 0.72 & 0.79 & 5.14 \\
\hline
\end{tabular}
}
\label{tab:exp_textedit_rule_miniset}
\end{table}

%% file: tables/final/textedit-mllm-miniset.tex
\begin{table}[htbp]
\centering
\caption{\textbf{Evaluation of text-centric image editing on TextEdit MiniSet-500(MLLM-based Metrics).} Sizes of unified models in  ``A + B'' indicate separate understanding (A) and generation (B) parameters. ``Real'' refers to source images from real-world scenes, while  ``Virtual'' refers to images from virtual scene. Abbreviations: \textbf{TA}: Text Accuracy, \textbf{TP}: Text Preservation, \textbf{SI}: Scene Integrity, \textbf{LR}: Local Realism, \textbf{VC}: Visual Coherence, \textbf{Avg}: MLLM Overall Average.}
\resizebox{\textwidth}{!}{
\begin{tabular}{l|c|cccccc|cccccc}
\noalign{\vspace{1pt}}
\hline
\noalign{\vspace{1pt}}
\multirow{2}{*}{\textbf{Models}} & 
\multirow{2}{*}{\textbf{\# Params}} & 
\multicolumn{6}{c}{\textbf{Real}} & 
\multicolumn{6}{c}{\textbf{Virtual}} \\
\cline{3-8}
\cline{9-14}
& &
\textbf{TA} & \textbf{TP} & \textbf{SI} & \textbf{LR} & \textbf{VC} & \textbf{Avg} &
\textbf{TA} & \textbf{TP} & \textbf{SI} & \textbf{LR} & \textbf{VC} & \textbf{Avg} \\
\hline

\multicolumn{14}{l}{\cellcolor{gray!15}\textit{Generation Models}} \\
Qwen-Image-Edit~\cite{wu2025qwenimagetechnicalreport} & 20B & 0.93 & 0.85 & 0.77 & 0.55 & 0.78 & 0.80 & 0.60 & 0.82 & 0.91 & 0.81 & 0.74 & 0.76 \\
GPT-Image-1.5~\cite{GPT-Image-1.5} & - & 0.97 & 0.94 & 0.86 & 0.79 & 0.92 & 0.91 & 0.85 & 0.93 & 0.95 & 0.92 & 0.83 & 0.88 \\
Nano Banana Pro~\cite{deepmind_gemini3proimage_2025} & - & 0.96 & 0.95 & 0.85 & 0.86 & 0.92 & 0.91 & 0.87 & 0.92 & 0.96 & 0.93 & 0.87 & 0.92 \\
\hline

\multicolumn{14}{l}{\cellcolor{gray!15}\textit{Unified Models}} \\
Lumina-DiMOO~\cite{xin2025lumina} & 8B & 0.16 & 0.04 & 0.04 & 0.02 & 0.06 & 0.08 & 0.02 & 0.05 & 0.19 & 0.07 & 0.03 & 0.10 \\
Ovis-U1~\cite{wang2025ovis} & 2.4B+1.2B & 0.29 & 0.11 & 0.11 & 0.08 & 0.20 & 0.17 & 0.04 & 0.16 & 0.35 & 0.18 & 0.15 & 0.22 \\
BAGEL~\cite{deng2025bagel} & 7B+7B & 0.68 & 0.61 & 0.38 & 0.34 & 0.59 & 0.53 & 0.36 & 0.52 & 0.69 & 0.64 & 0.40 & 0.54 \\
\modelname & 2B+1.7B & 0.94 & 0.91 & 0.72 & 0.73 & 0.75 & 0.89 & 0.88 & 0.87 & 0.90 & 0.78 & 0.57 & 0.79 \\
\hline
\end{tabular}
}
\label{miniset-textedit-mllm}
\end{table}